\documentclass[12pt,a12paper,openany]{book}

\usepackage[ruled, lined, commentsnumbered, longend]{algorithm2e}
\usepackage{xcolor}
\usepackage{adjustbox}
\usepackage{pifont}
\newcommand{\cmark}{\ding{51}}%
\newcommand{\xmark}{\ding{55}}%

\usepackage{cite}

\usepackage{float}

\usepackage{textcomp}

\usepackage{algorithmic}

\SetKw{KwBy}{by}

\usepackage{url}
\usepackage{multirow}

\usepackage{pslatex}
\usepackage{amsthm}
\usepackage{amssymb}
\usepackage{amsfonts}
\usepackage{mathrsfs}

\usepackage{amsmath,amsfonts,amssymb,geometry,times}
\usepackage[T1]{fontenc}
\usepackage{dsfont}
\usepackage{ae} 
\usepackage{xcolor}
\usepackage[english]{babel}
\usepackage{graphicx}
\usepackage{hyperref}
\usepackage [Lenny]{fncychap}
\begin{document}

\begin{titlepage}
\begin{center}
\topskip0pt
\vspace*{\fill}
\centerline{\LARGE   \textbf{\textbf{\bf Mining Customers' Opinions for Online Reputation}}}
\centerline{\LARGE   \textbf{\textbf{\bf Generation and Visualization in e-Commerce Platforms}}}
\bigskip
\bigskip
\bigskip
\bigskip
\bigskip
\centerline{\LARGE   \textbf{\textbf{\bf A DISSERTATION}}}
\bigskip
\bigskip
\bigskip
\bigskip
\bigskip
\centerline{\LARGE   \textbf{\textbf{\bf SIDI MOHAMED BEN ABDELLAH UNIVERSITY}}}
\centerline{\LARGE   \textbf{\textbf{\bf (USMBA)}}}
\centerline{\LARGE   \textbf{\textbf{\bf FACULTY OF SCIENCE DHAR EL MAHRAZ}}}
\centerline{\LARGE   \textbf{\textbf{\bf (FSDM)}}}
\bigskip
\bigskip
\bigskip
\bigskip
\bigskip
\centerline{\LARGE   \textbf{\textbf{\bf IN PARTIAL FULFILLMENT OF THE}}}
\centerline{\LARGE   \textbf{\textbf{\bf REQUIREMENTS FOR THE DEGREE OF}}}
\centerline{\LARGE   \textbf{\textbf{\bf DOCTOR OF COMPUTER SCIENCE}}}
\bigskip
\bigskip
\bigskip
\bigskip
\bigskip
\centerline{{\normalsize   \textbf{Abdessamad Benlahbib}}}
\bigskip
\bigskip
\bigskip
\bigskip
\bigskip
\centerline{{\normalsize   \textbf{March 2021}}}
\vspace*{\fill}
\end{center}

\end{titlepage}
\newtheorem{theorem}{Theorem}[section]
\newtheorem{definition}{Definition}[section]
\newtheorem{remark}{Remark}
\newtheorem{nott}{Notation}
\newtheorem{lemma}{Lemma}[section]
\newtheorem{proposition}{Proposition}[section]
\newtheorem{corollary}{Corollary}[section]
\newtheorem{pro}{Properties}[section]
\newtheorem{example}{Example}
\newtheorem{q}{Question}
\newcommand{\pr} {{\bf   Proof: \hspace{0.3cm}}}


\tableofcontents
\listoffigures
\listoftables


\newpage
\chapter*{Acknowledgments}
\addcontentsline{toc}{chapter}{Acknowledgments}

I would like to thank my esteemed supervisor Dr. El Habib Nfaoui for his invaluable supervision, support, and tutelage during the course of my PhD degree. I would also like to thank my friend Mohammed El Moutaouakkil for helping me during the writing and proofreading of this thesis. My appreciation also goes out to my family and friends for their encouragement and support all through my studies.




\newpage
\centerline{{\normalsize   \textbf{Abstract} }}
\addcontentsline{toc}{chapter}{Abstract}
\noindent
Customer reviews represent a very rich data source from which we can extract very valuable information about different online shopping experiences. The amount of the collected data may be very large especially for trendy items (products, movies, TV shows, hotels, services \dots), where the number of available customers’ opinions could easily surpass thousands. In fact, while a good number of reviews could indeed give a hint about the quality of an item, a potential customer may not have time or effort to read all reviews for the purpose of making an informed decision (buying, renting, booking \dots). Thus, the need for the right tools and technologies to help in such a task becomes a necessity for the buyer as for the seller. My research goal in this thesis is to develop reputation systems that can automatically provide E-commerce customers with valuable information to support them during their online decision-making process by mining online reviews expressed in natural language.\par
The first chapter describes and examines previous research work done in the area of natural language processing (NLP) techniques for decision making in E-commerce, document-level sentiment analysis, and fine-grained sentiment analysis. The chapter also covers the necessary background for understanding Bidirectional Encoder Representations from Transformers (BERT) model since we employed it to determine the sentiment orientation of customer and user reviews.\par
Chapter 2 presents a reputation system that incorporates sentiment analysis, semantic analysis, and opinion fusion to generate accurate reputation values toward online items.\par
The next chapter describes MTVRep, a movie and TV show reputation system that exploits fine-grained sentiment analysis and semantic analysis for the purpose of generating and visualizing reputation toward movies and TV shows.\par
Chapter 4 presents a reputation system that incorporates four review attributes: review helpfulness, review time, review sentiment polarity, and review rating in order to generate reputation toward various online items (products, movies, TV shows, hotels, restaurants, services).\\
The system also provides a comprehensive reputation visualization form to help potential customers make an informed decision by depicting the numerical reputation value, opinion group categories, and top-k positive reviews as top-k negative reviews.\par
Chapter 5 describes AmazonRep, a reputation system that extends the system proposed in chapter 4 by exploiting review rating, review helpfulness votes, review time, review sentiment orientation, and user credibility to support Amazon's customer decision making process.\par
\bigskip
\noindent
\textbf{Key words:}\\
Reputation Generation, Reputation Visualization, E-commerce Decision Making, Opinion Mining, Natural Language Processing (NLP), Bidirectional Encoder Representations from Transformers (BERT), Embeddings from Language Models (ELMo), Machine Learning, Deep Learning, Semantic Analysis.
\iftrue
\newpage
\centerline{{\normalsize   \textbf{Abstract français } }}
\addcontentsline{toc}{chapter}{Abstract français}
Un avis client ou avis consommateur est un élément apporté par un client ou consommateur pour évaluer sa satisfaction vis-à-vis  d’une expérience, d’un produit ou d’un service.  Aujourd’hui 88\% des consommateurs consultent les avis clients en ligne avant un achat et 90\% déclarent que ces avis ont influencé leur décision d’achat. Les avis clients représentent donc un canal incontournable pour accéder à des retours d’expérience client, des suggestions et idées de consommateurs. Cela permet de collecter des insights précieux et très diversifiés sur les produits et services qui permettent au client de se forger une première impression. Or, le nombre d'avis clients pourrait facilement dépasser les milliers envers les produits tendances peu importe leur catégorie (électroménager, films, séries TV, hôtels, services \dots). Ainsi, bien qu’un nombre élevé d’avis pourrait mieux informer sur la qualité d’un article, le client de son côté peut se trouver perdu entre une multitude d'opinions sur le produit désiré. D'où émerge un nouveau défi à relever: accompagner la clientèle au cours de son processus de prise de décision en ligne. \par
Pour remédier à cette problématique, nous proposons à travers ce travail inscrit dans le cadre de thèse de doctorat, de développer des systèmes en faveur des clients et qui, à partir des commentaires exprimés en langage naturel, peuvent générer des réputations capables d’épargner temps et effort, et donc assurer le support digital nécessaire durant la phase de prise de décision en ligne facilitant ainsi les expériences d’achat en ligne.  \par
Le premier chapitre introduit une étude bibliographique (related work) démontrant les techniques préexistantes relatives au domaine du traitement automatique du langage naturel appliquées pour assister le consommateur durant le processus de décision d'achat dans le social E-Commerce, l’analyse des sentiments au niveau du document (document-level sentiment analysis) ainsi que l'analyse fine des sentiments (fine-grained sentiment analysis). \par
Le deuxième chapitre présente un système de réputation qui combine l’analyse des sentiments (sentiment analysis), l'analyse sémantique (semantic analysis) ainsi que la fusion des opinions (opinion fusion) afin de générer une valeur de réputation envers différentes entités (produits, films, hôtels, services). \par
Le troisième chapitre introduit MTVRep, un système de réputation dédié aux films et aux séries TV. Le système proposé combine l'analyse sémantique (semantic analysis) avec l'analyse fine des sentiments (fine-grained sentiment analysis) pour la génération et la visualisation de la réputation. \par
Le quatrième chapitre présente un système de réputation qui prend en considération l’utilité du commentaire (review helpfulness), la date de publication du commentaire (review time), le sentiment exprimé dans le commentaire (review sentiment polarity) ainsi que le nombre d’étoiles associées au commentaire (review rating) dans le but de générer la réputation envers différentes entités (produits, films, séries TV, hôtels, services). \par
Le dernier chapitre décrit AmazonRep, une extension du système de réputation présenté en chapitre 4 qui considère l’utilité du commentaire (review helpfulness), la date de publication du commentaire (review time), le sentiment exprimé dans le commentaire (review sentiment polarity), le nombre d’étoiles associées au commentaire (review rating) ainsi que la crédibilité du commentateur (user credibility) afin d’assister les clients d'Amazon durant le processus: \textbf{"prise de décision d'achat"}.\\
\\
\\
\textbf{Mots Clés:}\\
Génération de la réputation, Visualisation de la réputation, Processus de prise de décision dans le social E-Commerce, Fouille d'opinions, Traitement automatique des langues, BERT, ELMo, Apprentissage automatique, Apprentissage profond, Analyse sémantique.

\fi


\newpage
\chapter*{Introduction}
\addcontentsline{toc}{chapter}{Introduction }
\section*{Overview}
Nowadays, the amount of unstructured text data that online users produce on E-commerce websites has grown dramatically. The International Data Corporation\footnote{\url{https://www.idc.com/}} (IDC) estimates that unstructured data already accounts for a staggering ninety percent of all digital data. The shocking fact is that in most cases, 90\% of that data is never analyzed! Therefore, the need to automatically process it and extract different types of knowledge from it becomes a necessity. My research goal in this thesis is to develop reputation systems that can automatically provide E-commerce customers with valuable information to support them during their online decision-making process (buying, booking, renting \dots) by mining online reviews expressed in natural language. \par
Previous studies on reputation generation of online products have primarily focused on exploiting user ratings \cite{paul2002trust, Schneider:2000:DTI:593556.593569, garcin2009aggregating, resnick2000reputation, 10.1007/978-3-642-29873-8_8, CHO20093751} to generate a score toward online products. Surprisingly, textual data (reviews), which constitute a significant part of customer information, have been somewhat ignored until 2012 when Abdel-Hafez et al. \cite{quteprints58118} designed a product reputation model that uses text reviews rather than users’ ratings for the purpose of generating a more realistic reputation value for every feature of the product as for the product itself. This was achieved by incorporating opinion orientation and opinion strength (sentiment analysis), though, there was no evidence to support the efficiency of the proposed reputation system since the authors \textbf{assumed} that the product features and the opinion orientation and strength to the features in each product review have been determined by using existing opinion mining techniques without actually applying them. In 2017, Yan et al. \cite{yan2017fusing} proposed a reputation system that exploits textual reviews and their ratings to generate numerical reputation values toward Amazon's products. The system fuses and groups reviews into several principal opinion sets based on review relevance (Latent semantic analysis and cosine similarity measure), citations, votes, and ratings, and then aggregates the fused and grouped opinions to generate a reputation value by considering the popularity and other statistics of principal opinion sets, though, previous studies on reputation generation and visualization have two shortcomings.
\begin{enumerate}
  \item\textbf{Exploited features}: Besides numerical rating and textual content, online reviews contain other useful information that could be exploited for reputation generation and visualization, such as review helpfulness, which implies that reviews that receive higher votes from other users typically provide more valuable information, review time, which means that more recent reviews generally provide users with more up-to-date information, and user credibility, which implies that reviews written by trusted users have more impact on the popularity of the target item (product, movie, hotel, service \dots). Nevertheless, previous studies on reputation generation have disregarded incorporating these features during the reputation generation phase.
  \item\textbf{Reputation visualization}: It is very important to provide potential customers with valuable information toward the target entity (product, movie, hotel, service \dots) in order to support them during the decision-making process in E-Commerce (buying, renting, booking \dots). However, previous studies on reputation generation have provided customers with insufficient information during the reputation visualization phase, for instance, the numerical reputation value toward the target entity and the opinion categories, neglecting other helpful information such as the top-k positive reviews, the top-k negative reviews, the distribution of sentiment over reviews \dots 
\end{enumerate}
The reputation systems presented in this thesis address these two shortcomings. They incorporate various features (review time, review helpfulness, numerical rating, user credibility, review sentiment orientation) and various opinion mining techniques (binary sentiment analysis, fine-grained sentiment analysis, semantic analysis) during the reputation generation phase. Furthermore, they provide customers with a comprehensive form of reputation visualization that contains valuable information toward the target entity (product, movie, hotel, service \dots) including the numerical reputation value, the distribution of sentiment over reviews, the top-k positive reviews, and the top-k negative reviews. \par

\section*{Contributions and Outline of This Thesis}

This thesis is structured as follows: 

\subsection*{Summary Chapter 1: Background and Literature Review}

The first chapter describes and examines previous research work done in the area of natural language processing (NLP) techniques for decision making in E-commerce, document-level sentiment analysis, and fine-grained sentiment analysis. Besides, the chapter covers the necessary background for understanding BERT model \cite{devlin2018bert} because we employed it to determine the polarity of customer and user reviews.

\subsection*{Summary Chapter 2: Sentiment Analysis and Opinion Fusion for Reputation Generation}

In the second chapter, we have an interest in improving Yan et al. \cite{yan2017fusing} approach by generating accurate reputation values toward online entities (movies, products, hotels, services \dots). The fact that the majority of reviews hold a positive or negative sentiment (point of view) toward a target entity led us to the idea of using a classification step in order to separate positive and negative reviews before grouping them into different sets based on semantic relations, then computing reputation toward the target entity using Weighted Arithmetic Mean.

\subsection*{Summary Chapter 3: Fine-grained Opinion Mining and Semantic Analysis for Reputation Generation}

In the third chapter, we propose MTVRep, a movie and TV show reputation system that applies fine-grained opinion mining to separate reviews into five opinion groups: strongly negative, weakly negative, neutral, weakly positive, and strongly positive. Then, it computes a custom score for each group based on acquired statistics. These statistics include the number of reviews, the sum of their ratings, and the sum of their semantic similarity (ELMo and cosine metric). Finally, a numerical reputation value is produced toward the target movie or TV show using the weighted arithmetic mean.

\subsection*{Summary Chapter 4: Aggregating Customer Review Attributes for Online Reputation Generation}

In the fourth chapter, we propose a reputation system that generates reputation toward various items (products, movies, TV shows, hotels, restaurants, services) by mining customer and user reviews expressed in natural language. The system incorporates four review attributes: review helpfulness, review time, review sentiment polarity, and review rating. The system also provides a comprehensive reputation visualization form by depicting the numerical reputation value, opinion group categories, top-k positive reviews, and top-k negative reviews.

\subsection*{Summary Chapter 5: Reputation Generation and Visualization for Amazon’s Products}

In the fifth chapter, we propose AmazonRep, a reputation system that extends the system proposed in chapter 4 by exploiting review rating, review helpfulness votes, review time, review sentiment orientation, and user credibility for the purpose of supporting Amazon's customer decision making process.\par


\chapter{Background and Literature Review}
\section{Introduction}
This chapter describes and examines previous research work done in the area of natural language processing (NLP) techniques for decision making in E-commerce, document-level sentiment analysis, and fine-grained sentiment analysis. Additionally, the chapter covers the necessary background for understanding BERT model \cite{devlin2018bert} since we employed it to determine the sentiment orientation of customer and user reviews.

\section{NLP techniques for online decision making in E-commerce}

The BusinessDictionary\footnote{\url{http://www.businessdictionary.com/definition/decision-making.html}} defines decision making as: \textit{"The thought process of selecting a logical choice from the available options"}. During the last twenty years, few systems have been proposed to help potential customers in making decisions in E-commerce websites using mainly two approaches: feature-based summarization of customer reviews and reputation generation. \par
\subsection{Feature-based opinion summarization}
Feature-based opinion summarization is :\textit{"one of the opinion summarization techniques which provide brief yet most important information containing summary about different aspects related to the target product. Since it focuses on different features instead of giving the general details about a product, it has become more significant and demanded form of summarization. This technique is also known as Aspect-based Opinion Summarization. It is actually a way of generating summaries for a set of aspects or features of a specific product"} \cite{makadia2016feature}. Hu \& Liu (2004) \cite{Hu:2004:MSC:1014052.1014073} were the first to design and build a system that produces a feature-based summary from customer reviews. The proposed system performs three tasks: (1) association rule mining \cite{Agrawal:1994:FAM:645920.672836} is used to extract product features from customer reviews, (2) WordNet \cite{Miller:1995:WLD:219717.219748} is utilized to predict the semantic orientations of opinion words, (3) a structured feature-based summary is produced. Over the last two decades, few systems have been proposed to perform feature-based summarization. The summarizers are applied on various domains: product reviews \cite{Hu:2004:MSC:1014052.1014073, Popescu:2005:EPF:1220575.1220618, garcia2013retrieving, abbasi2010selecting, chi2017adaptive}, movie reviews \cite{Zhuang:2006:MRM:1183614.1183625, 6046951}, local services reviews \cite{34368} and hotel reviews \cite{10.1016/j.ipm.2016.12.002, 8672719}, etc. Figure 1.1 depicts a sample of feature-based summary proposed by Kangale et al. \cite{doi:10.1080/00207721.2015.1116640}. \par

\begin{figure}[H]
\centering
\includegraphics[scale=0.9]{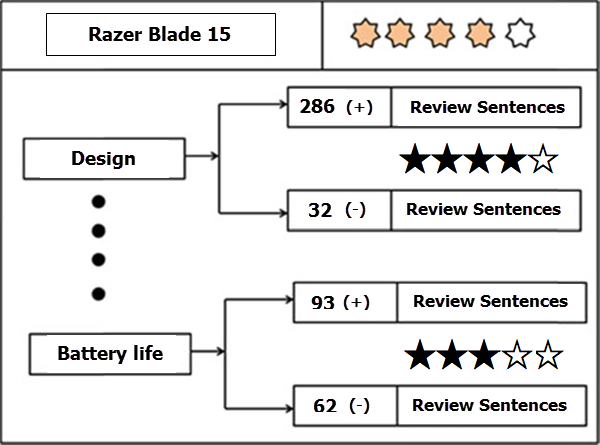}
\caption{Kangale et al. \cite{doi:10.1080/00207721.2015.1116640} feature-based summary}
\end{figure}

\subsection{Reputation generation}
Backing to reputation generation. The Cambridge Dictionary \footnote{\url{https://dictionary.cambridge.org/fr/dictionnaire/anglais/reputation}} defines reputation as \textit{"the opinion that people, in general, have about someone or something, or how much respect or admiration someone or something receives, based on past behavior or character"}. The pioneer work that tackles the task of reputation generation based on mining opinions expressed in natural languages was firstly proposed by Yan et al. (2017) \cite{yan2017fusing} in which reviews are fused into several opinion sets based on their semantic relations, then, a single reputation value is generated by aggregating the fused and grouped opinions' statistics (the sum of similarities, the sum of ratings, the number of reviews). In \cite{benlahbib2019unsupervised}, the authors applied K-means clustering algorithm to group similar reviews into the same cluster using Latent Semantic Analysis (LSA) before producing a reputation value using the statistics of each cluster. However, both approaches have relied on extracting semantic relations between reviews and have disregarded the fact that the majority of online customer and user reviews are opinionated. Benlahbib \& Nfaoui (2019) \cite{benlahbib2019hybrid} proposed a fourfold approach to improve the reputation system proposed in \cite{yan2017fusing}. First, Na\"{\i}ve Bayes and Linear Support Vector Machines classifiers were applied to separate reviews into positives and negatives by predicting their sentiment polarity. Second, positive and negative reviews were fused into different sets based on their semantic similarity (Latent Semantic Analysis and cosine similarity). Third, a custom reputation value is computed separately for both positive opinion sets and negative opinion sets. Finally, a single reputation value is calculated using the weighted arithmetic mean. The same authors designed and built a new reputation system \cite{9098950} that considers review time, review helpfulness votes, and review sentiment orientation for the purpose of generating reputation toward online entities (products, movies, hotels, and services). The system also provides a new form of reputation visualization by depicting the numerical reputation value, the distribution of sentiment over the reviews, the top-k positive reviews, and the top-k negative reviews. Recently, Benlahbib \& Nfaoui \cite{Benl2112:MTVRep} incorporated fine-grained opinion mining and semantic analysis to generate and visualize reputation toward movies and TV shows (MTVRep). The authors trained the Multinomial Na\"{\i}ve Bayes classifier on the SST-5 dataset \cite{manning-etal-2014-stanford} to group reviews into five emotion classes: strongly negative, weakly negative, neutral, weakly positive, and strongly positive, then, Embeddings from Language Models (ELMo) \cite{peters-etal-2018-deep} and cosine similarity were applied to extract the semantic similarity between reviews and to compute a custom score for each emotion class. Finally, the weighted arithmetic mean is used to compute the movie or TV show reputation value. \par

\section{Document-level Sentiment Analysis}

Ahlgren (2016) \cite{7836762} defines sentiment analysis as \textit{"the process of identifying and detecting subjective information using natural language processing, text analysis, and computational linguistics"}. Generally, sentiment analysis can be divided into three levels: sentence-level opinion mining, document-level opinion mining, and fine-grained opinion mining. \par
According to \cite{Sun:2017:RNL:3174385.3174741}, document-level opinion mining is \textit{"a task of extracting the overall sentiment polarities of given documents, such as movie reviews, product reviews, tweets, and blogs"}. \par
Many approaches have been used to handle the task of document-level sentiment analysis:
\begin{itemize}
	\item Supervised approaches: These approaches require annotated corpus to train machine learning models. The first work for supervised document-level opinion mining was proposed by Pang et al. (2002) \cite{Pang:2002:TUS:1118693.1118704}. Three machine learning classifiers (Support Vector Machines (SVMs) \cite{Cortes1995}, Na\"{\i}ve Bayes classifier \cite{Maron:1961:AIE:321075.321084} and Maximum Entropy classifier \cite{Berger:1996:MEA:234285.234289}) were trained with movie reviews labeled by sentiment (positive/negative). The authors trained the three models on various kinds of features (unigrams, bigrams, parts of speech, and position) and found that the sentiment classification task performs well when adopting unigrams as features. Kennedy \& Inkpen (2006) \cite{doi:10.1111/j.1467-8640.2006.00277.x} trained Support Vector Machine classifiers on unigrams and bigrams by incorporating three types of context valence shifters: \textit{"intensifiers"}, \textit{"negations"} and \textit{"diminishers"}. The trained model achieved an accuracy of 0.859 on movie review data\footnote{\url{http://www.cs.cornell.edu/people/pabo/movie-review-data/}} \cite{pang-lee-2004-sentimental}. Koppel \& Schler (2006) \cite{koppel2006the} defined the sentiment classification task as a three-category problem (positive, negative, and neutral) and used different learning algorithms: SVM, J48 Decision Tree \cite{Quinlan:1993:CPM:152181}. Na\"{\i}ve Bayes, Linear Regression \cite{doi:10.1142/6986}, and Frank \& Hall (2001) \cite{Frank:2001:SAO:645328.649997} classification method. The results show significant improvement in the sentiment classification accuracy when using neutral examples over ignoring them. In \cite{wang-manning-2012-baselines}, the authors combined Na\"{\i}ve Bayes and Support Vector Machine by training SVM with Na\"{\i}ve Bayes log-count ratios as features. The proposed model has achieved promising results across several datasets. Jing et al. (2015) \cite{Jing2015HowSF} applied Na\"{\i}ve Bayes algorithm on 3046 customer reviews related to fifty-eight business-to-team (B2T) websites to study the survival conditions of B2T companies. Augustyniak et al. (2016) \cite{augustyniak2016comprehensive} presented a wide comparison and analysis of opinion mining task for several classifiers: Random Forests \cite{breiman2001random}, Linear SVC, Bernoulli Na\"{\i}ve Bayes, Mulinomial Na\"{\i}ve Bayes, Extra Tree Classifier \cite{geurts2006extremely}, Logistic Regression, and AdaBoost \cite{freund1996experiments, freund1999short}. They conducted experiments on Amazon review data \cite{McAuley:2013:HFH:2507157.2507163} and found that the Logistic Regression classifier outperforms the other classifiers in predicting sentiment polarity of product reviews.
    \item Unsupervised approaches: They attempt to determine the sentiment orientation of a text by applying a set of rules and heuristics obtained from language knowledge. Turney (2002) \cite{Turney:2002:TUT:1073083.1073153} was the first to propose an unsupervised sentiment analysis technique to classify reviews as \textit{"recommended"}  or \textit{"not recommended"}. The semantic orientation of a phrase is computed as the pointwise mutual information (PMI) \cite{Turney:2001:MWS:645328.650004} between the given phrase and the word \textit{"excellent"} minus the pointwise mutual information between the given phrase and the word \textit{"poor"}. The proposed algorithm achieved an accuracy of 84\% for automobile reviews, 80\% for bank reviews, 71\% for travel destination reviews, and 66\% for movie reviews. In \cite{taboada-etal-2011-lexicon}, the authors proposed a lexicon-based method to mine text by using a dictionary of sentiment words and their semantic orientations varied between -5 and +5. The authors also incorporated amplifiers, downtoners, and negation words to compute a sentiment score for each document. Vashishtha \& Susan (2020) \cite{vashishtha2019fuzzy} proposed a fuzzy rule-based approach to perform opinion mining of tweet. The authors use a novel unsupervised nine fuzzy rule-based system to predict the sentiment orientation of the post (positive, negative, or neutral). In \cite{fernandez2016unsupervised}, Fernández-Gavilanes et al. (2016) proposed a sentiment analysis approach to predict the polarity in online textual messages such as tweets and reviews using an unsupervised dependency parsing-based text classification method.
	\item Deep learning approaches: Over the past few years, deep learning models have greatly improved the state-of-the-art of opinion mining. Moraes et al. (2013) \cite{Moraes:2013:DSC:2388116.2388142} made a comparative study between Support Vector Machines (SVM) and Artificial Neural networks (ANN) for document-level opinion mining and found that ANN results are at least comparable or superior to SVMs. In \cite{Le:2014:DRS:3044805.3045025}, the authors proposed an unsupervised algorithm named paragraph vector (doc2vec), an extension to word2vec approach \cite{Mikolov:2013:DRW:2999792.2999959} in order to overcome the weakness of the bag-of-words (BoW) model. The proposed algorithm learns vector representations for variable-length texts such as sentences, paragraphs, and documents. Experimental results depict that doc2vec algorithm achieved new state-of-the-art results on several sentiment analysis tasks. Johnson \& Zhang (2015) \cite{johnson-zhang-2015-effective} trained a parallel Convolutional Neural Network (CNN) \cite{fukushima1980neocognitron} without using pre-trained word vectors: word2vec, doc2vec, and GloVe\footnote{\url{https://nlp.stanford.edu/projects/glove/}} \cite{pennington2014glove}. Instead, convolutions are directly applied to one-hot encoding vectors to leave the network solely with information about the word order. The proposed approach achieved an accuracy rate of 92.33\% on Large Movie Review Dataset\footnote{\url{https://ai.stanford.edu/~amaas/data/sentiment/}} outperforming both SVM \cite{Bishop:1995:NNP:525960} and NB-LM \cite{mesnil2014ensemble}. Baktha \& Tripathy (2017) \cite{8286763} investigated the performance of Long Short-Term Memory (LSTM) \cite{doi:10.1162/neco.1997.9.8.1735}, vanilla RNNs, and Gated Recurrent Units (GRU) \cite{cho-etal-2014-learning} on the Amazon health product reviews dataset and sentiment analysis benchmark datasets SST-1 and SST-2. The results depict that GRU achieved the highest sentiment classification accuracy. In \cite{xie2019unsupervised}, the authors combined unsupervised data augmentation (UDA) with Bidirectional Encoder Representations from Transformers (BERT) \cite{devlin2018bert} and compared it with fully supervised $BERT_{LARGE}$ on six text classification benchmark datasets. The authors reported that their proposed approach outperforms $BERT_{LARGE}$ on five text classification benchmark datasets including Large Movie Review Dataset. Facebook AI and University of Washington researchers \cite{liu2019roberta} improved BERT by proposing \textit{"Robustly Optimized BERT approach"} (RoBERTa) that was trained with more data and more number of pretraining steps and has dropped the next sentence prediction (NSP) approach used in BERT. Recently, Google researchers \cite{yang2019xlnet} proposed XLNet, a \textit{"generalized autoregressive pretraining method"} that outperforms Bidirectional Encoder Representations from Transformers (BERT) on twenty text classification tasks and achieves state-of-the-art results on eighteen text classification tasks including sentiment analysis. At the beginning of 2020, Lan et al. (2019) \cite{lan2019albert} proposed ALBERT: \textit{"A Lite BERT for Self-supervised Learning of Language Representations"}. The paper describes parameter reduction techniques to lower memory reduction and increase the training speed and accuracy of BERT models. In \cite{raffel2019exploring}, the authors introduced a novel \textit{"Text-to-Text Transfer Transformer"} (T5) neural network model pre-trained on a large text corpus that can convert any language problem into a text-to-text format. The T5 model achieved state-of-the-art results on the SST-2 binary classification dataset with an accuracy of 97.4\%. Recently, Clark et al. (2020) \cite{Clark2020ELECTRA:} presented \textit{ELECTRA}, a method for self-supervised language representation learning that uses a new pre-training task called replaced token detection (RTD). The experiment results showed that RTD is more efficient than masked language modeling (MLM) pre-training models such as BERT.

\end{itemize}

\section{Fine-grained Sentiment Analysis on the 5-class Stanford Sentiment Treebank (SST-5) dataset}

Differently from binary sentiment analysis that aims to determine whether a given text holds a positive or negative polarity, fine-grained sentiment analysis is a significantly more challenging task in which the polarity of a given text is classified into five discrete classes: strongly negative, weakly negative, neutral, weakly positive, and strongly positive. \par

\begin{figure}[H]
\centerline{\includegraphics[scale=0.9]{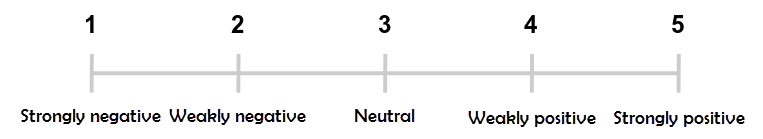}}
\caption{Typical class labels (or intensities) for fine-grained sentiment classification}
\end{figure}

Many approaches have been proposed to handle the task of fine-grained sentiment analysis for the 5-class Stanford Sentiment Treebank (SST-5) dataset. Xu et al. \cite{xu-etal-2018-emo2vec} proposed Emo2Vec which are word-level representations that encode emotional semantics into fixed-sized, real-valued vectors. Mu et al. \cite{mu2017all} presented a simple post-processing operation that renders word representations even stronger by eliminating the top principal components of all words. Socher et al. \cite{socher-etal-2013-recursive} introduced Recursive Neural Tensor Networks and the Stanford Sentiment Treebank. Wang et al. \cite{10.1145/3178876.3186015} proposed RNN-Capsule, a capsule model based on Recurrent Neural Network (RNN) for sentiment analysis. Yang \cite{yang-2018-convolutional} presented RNFs, a new class of convolution filters based on recurrent neural networks. McCann et al. \cite{mccann2017learned} introduced an approach for transferring knowledge from an encoder pre-trained on machine translation to a variety of downstream natural language processing (NLP) tasks. Munikar et al. \cite{munikar2019fine} used the pre-trained BERT \cite{devlin2018bert} model and fine-tuned it for the fine-grained sentiment classification task on the SST-5 dataset. \par
Table 1.1 summarizes the latest works on fine-grained opinion mining applied to the Stanford Sentiment Treebank dataset (SST-5). \par

\begin{table}[htb!]
\centering
\scalebox{1.2}{
\begin{tabular}{|c|c|c|}
\hline
Method & Authors \& Year & Accuracy \% \\ \hline
RoBERTa-large+Self-Explaining & Sun et al. (2020) \cite{sun2020selfexplaining} & 59.1 \\ \hline
BCN+Suffix BiLSTM-Tied+CoVe & Brahma (2018) \cite{brahma2018improved} & 56.2  \\ \hline
BERT large & Munikar et al. (2019) \cite{munikar2019fine} & 55.5 \\ \hline
BCN+ELMo & Peters et al. (2018) \cite{peters-etal-2018-deep} & 54.7  \\ \hline
BCN+Char+CoVe & McCann et al. (2017) \cite{mccann2017learned} & 53.7  \\ \hline
CNN-RNF-LSTM & Yang (2018) \cite{yang-2018-convolutional} & 53.4  \\ \hline
RNN-Capsule & Wang et al. (2018) \cite{10.1145/3178876.3186015} & 49.3  \\ \hline
SWEM-concat & Shen et al. (2018) \cite{shen2018baseline} & 46.1  \\ \hline
RNTN & Socher et al. (2013) \cite{socher-etal-2013-recursive} & 45.7  \\ \hline
GRU-RNN-WORD2VEC & Mu et al. (2017) \cite{mu2017all} & 45.02   \\ \hline
GloVe+Emo2Vec & Xu et al. (2018) \cite{xu-etal-2018-emo2vec} & 43.6  \\ \hline
Emo2Vec & Xu et al. (2018) \cite{xu-etal-2018-emo2vec} & 41.6  \\ \hline
\end{tabular}
}
\caption{State-of-the-art results for Sentiment Analysis on SST-5 Fine-grained classification}
\end{table}

\section{Bidirectional Encoder Representations from Transformers (BERT)}
Bidirectional Encoder Representations from Transformers (BERT) \cite{devlin2018bert} is a pre-trained model that aims to pre-train deep bidirectional representations from unlabeled text (Figure 1.3).

\begin{figure}[!htb]
\centerline{\includegraphics[scale=0.6]{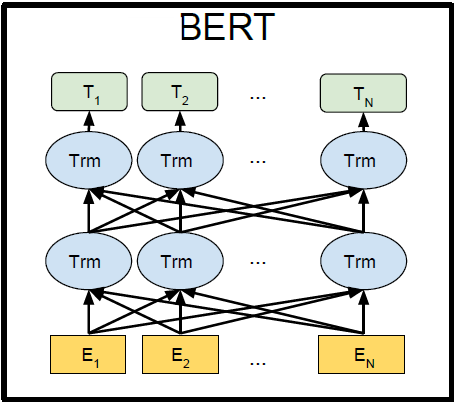}}
\caption{Bidirectional Encoder Representations from Transformers (BERT) pre-training architecture \cite{devlin2018bert}}
\end{figure}

BERT was pre-trained on two unsupervised tasks (Figure 1.4):
\begin{itemize}
\item \textbf{Masked LM (MLM)}: In order to train a deep bidirectional representation, BERT masks some percentage of the input tokens at random, and then predicts those masked tokens.
\item \textbf{Next Sentence Prediction (NSP))}: In order to train a model that understands sentence relationships, BERT predicts if a sentence B is the actual next sentence that follows a sentence A or not.
\end{itemize}
\begin{figure}[!htb]
\centerline{\includegraphics[scale=0.47]{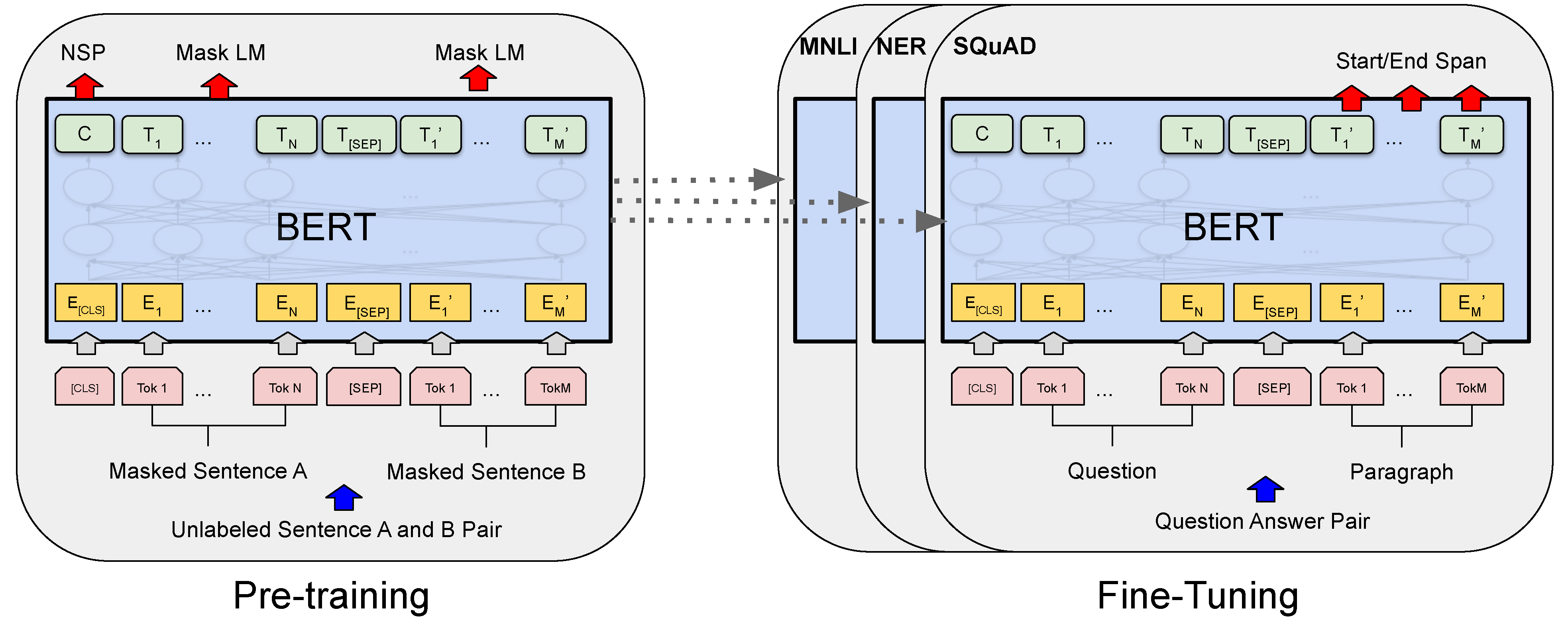}}
\caption{Overall pre-training and fine-tuning procedures for BERT \cite{devlin2018bert}}
\end{figure}

\section{Conclusion}

In this chapter, we described and examined previous research work done in the area of natural language processing (NLP) techniques for decision making in E-commerce, document-level sentiment analysis, and fine-grained sentiment analysis. We also provided the necessary background for understanding BERT model as we employed it to determine the polarity of customer and user reviews. \par
The next chapter will present a reputation system that incorporates sentiment analysis, semantic analysis, and opinion fusion to generate accurate reputation values toward online items.\par


%


\chapter{Sentiment Analysis and Opinion Fusion for Reputation Generation}

\section{Introduction}
Over the 21 century, the web has been expanding at an incredible rate. Thanks to social media and web communities, people express freely their opinions toward any entity they desire, with no chains or restrictions. Those opinions represent a valuable source of information that can contribute to track public perspectives and to generate reputation about a target entity for the reason that they carry the preferences and attitudes of humans. We can find many research papers and books for reputation systems \cite{Farmer:2010:BWR:1830476}, \cite{Yan:2013:TMM:2601608} and \cite{Josang:2007:STR:1225318.1225716}. However, there is particularly a lack of studies that handle the task of generating reputation based on mining opinions expressed in natural languages. Hence the importance of Yan et al. \cite{yan2017fusing} pioneering work that presents an approach to generate reputation toward Amazon products by fusing and mining opinions expressed in natural languages. \par
Yan et al. \cite{yan2017fusing} applied Latent Semantic Analysis (LSA) model and cosine similarity metric to group reviews into several principal opinion sets. The authors claimed that the opinions in each set hold a similar or same perspective. However, we believe that relying only on Latent Semantic Analysis model is not enough for extracting opinions with a similar perspective. In order to justify our point of view, we study the three reviews in Table 2.1. \par

\begin{table}[htb]
\centering
\caption{Latent Semantic Analysis model weakness in extracting opinions with a similar or same perspective}
{\renewcommand{\arraystretch}{2}
\scalebox{1.05}{
\small
\begin{tabular}{||p{3.5cm}||c||c||c||} \hline \hline
& The movie was good & The movie was not good at all & Amazing acting \\ \hline \hline
The movie was good & 1.000000 & 0.996730 & 0.221941 \\ \hline \hline
The movie was not good at all & 0.996730 & 1.000000 & 0.142422 \\ \hline \hline
Amazing acting & 0.221941 & 0.142422 & 1.000000 \\
\hline \hline\end{tabular}
}
}
\end{table}

Table 2.1 represents the results after applying LSA model, then computing document similarity using LSA components and cosine similarity metric. We notice that the similarity value between the first and the second review is very high, on the contrary, the similarity between the first and the third review is low, therefore, based on the opinion fusion and grouping algorithm proposed in \cite{yan2017fusing}, and by choosing an opinion fusion threshold $t_{0} \geq$ 0.23, we will get two opinion sets, the first set contains the first and the second reviews and the second set contains the third review, yet, it is obvious that the first and the third reviews stand for a positive point of view and the second review stands for a negative point of view, hence, we believe that adding a polarity classification step is an effective idea in order to classify reviews into positives and negatives (sentiment analysis) before grouping them based on their semantic relations. As a matter of fact, applying a sentiment classification step before the grouping phase will overcome the weakness of LSA model in extracting reviews with the same perspective and will guarantee that positive reviews will be grouped together and so on. \par
In this chapter, we have an interest in improving Yan et al. \cite{yan2017fusing} approach by generating accurate reputation values toward various online entities (movies, products, hotels, services \dots). The fact that the majority of reviews hold a positive or negative sentiment (point of view) toward a target entity led us to the idea of using a sentiment classification step to separate positive and negative reviews before grouping them into different sets based on semantic relations, then computing reputation toward the target entity using Weighted Arithmetic Mean. Indeed, the main steps of this contribution can be summarized as follows:
\begin{enumerate}
\item We apply a polarity classification step based on the two classifiers Na\"{\i}ve Bayes and Linear Support Vector Machine (LSVM) in order to determine positive and negative reviews.
\item We group \textbf{separately} positive and negative reviews into different opinion sets based on semantic relations.
\item We calculate a custom reputation value \textbf{separately} for positive and negative groups by considering some statistics of principal opinion sets.
\item We compute the final reputation value toward the target entity using Weighted Arithmetic Mean.
\end{enumerate}

\section{Problem definition}

This section covers the necessary background for understanding the remainder of this chapter, including the problem definition. \par
In this chapter, we face the problem of generating reputation for movies by combining opinion fusion and opinion mining. Given a set of reviews $R=\{r_{1},\ r_{2},\ \dots,\ r_{n}\}$ expressed for an entity $A$, the set of their attached ratings $V=\{v_{1},\ v_{2},\ \dots,\ v_{n}\}$ where $v_{i} \in [1,5]$ or $v_{i} \in [1,10]$ depending on the rating system, the set of their sentiment orientation predicted by Na\"{\i}ve Bayes model $nbp_=\{nbp(r_{1}),\ nbp(r_{2}),\ \dots,\ nbp(r_{n})\}$, the set of their sentiment orientation predicted by Linear Support Vector Machine model $svmp_=\{svmp(r_{1}),\ svmp(r_{2}),\ \dots,\ svmp(r_{p})\}$,  where $nbp(r_{i}), svmp(r_{i}) \in [positive,negative]$. The goal is to separate reviews into positives and negatives before grouping them into principal opinion sets based on their semantic relations (LSA model and cosine similarity metric), then, to compute a custom reputation value $Rep(A^{polarity})$ \textbf{separately} for positive and negative groups by considering some statistics of principal opinion sets such as $V_{k}^{polarity}$: the sum of ratings in principal opinion set k for a specific polarity, $S_{k}^{polarity}$: the sum of semantic similarity in principal opinion set k for a specific polarity, $N_{k}^{polarity}$: the number of similar opinions in principal opinion set k for a specific polarity and $K^{polarity}$: the number of opinion sets for a specific polarity (positive or negative). Finally, we compute the final reputation value $finalRep(A)$ using the Weighted Arithmetic Mean. Table 2.2 presents the descriptions of notations used in the rest of this chapter. \par

\begin{table}[!htb]
\caption{Symbol denotation}
\centerline{
\begin{tabular}{| l | p{14cm} |}
\hline
Symbol & Description                                                              \\ \hline

$A$      & The target entity                                                             \\
$R$      & The set of reviews expressed for the entity $A$                                                                  \\
$V$      & The set of ratings expressed for the entity $A$                                                                  \\

$V_{k}^{polarity}$ &  The sum of ratings in principal opinion set k for a specific polarity \\

$S_{k}^{polarity}$ & The sum of semantic similarity in principal opinion set k for a specific polarity\\

$N_{k}^{polarity}$ & The number of similar opinions in principal opinion set k for a specific polarity\\

$K^{polarity}$ & The number of opinion sets for a specific polarity (positive or negative) \\

$O^{polarity}$ & The number of (positive/negative) opinions\\

$O^{positive}$ & The number of positive opinions\\

$O^{negative}$ & The number of negative opinions\\

$A^{polarity}$ &  Groups of (positive/negative) opinions expressed for the entity $A$\\

$A^{positive}$ &  Groups of positive opinions expressed for the entity $A$\\

$A^{negative}$ &  Groups of negative opinions expressed for the entity $A$\\

$nbp$     & The set of sentiment orientation (positive or negative) predicted by Na\"{\i}ve Bayes model for reviews expressed toward the entity $A$                                             \\
$svmp$     & The set of sentiment orientation (positive or negative) predicted by Linear Support Vector Machine model for reviews expressed toward the entity $A$                                                      \\
$Rep(A^{polarity})$   & The custom reputation value for groups of (positive/negative) opinions.\\
$Rep(A^{positive})$     & The custom reputation value for groups of positive opinions.\\

$Rep(A^{negative})$     & The custom reputation value for groups of negative opinions.\\

$finalRep(A)$ &  The final reputation value toward entity $A$  \\ \hline                                                             
\end{tabular}
}
\end{table}

\section{Proposed approach}
\subsection{System overview}
Figure 2.1 describes the pipeline of our work.
\begin{enumerate}
	\item \textbf{Polarity classification phase.} We separate reviews into positives and negatives based on their sentiment polarity and their attached ratings by combining two classifiers: Na\"{\i}ve Bayes and Linear Support Vector Machine (LSVM). Both models are trained with 2000 movie reviews (1000 positive and 1000 negative processed reviews).
	\item \textbf{Fusion and grouping phase.} We apply the opinion fusion and grouping algorithm proposed in \cite{yan2017fusing} \textbf{\textit{separately}} for positive and negative reviews in order to group them into principal opinion sets based on their semantic relations (LSA model and cosine similarity metric).
	\item \textbf{Custom reputation generation phase.} We calculate a custom reputation value \textbf{\textit{separately}} for positive and negative groups by considering some statistics of principal opinion sets.
	\item \textbf{Final reputation generation phase.} We compute the final reputation value toward the target entity based on the custom reputation value for positive and negative groups using the Weighted Arithmetic Mean.\\
\end{enumerate}
\begin{figure}[!htb]
\centering
\includegraphics[width=1.0\textwidth]{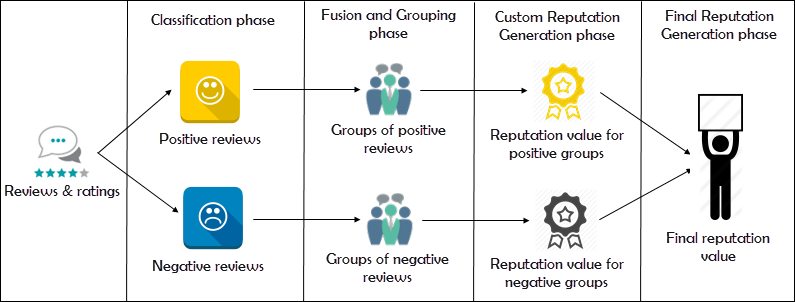}
\caption{Pipeline of our work}
\end{figure}

\subsection{Polarity classification phase}
We propose a combination of Na\"{\i}ve Bayes and LSVM classifiers to determine the polarity of the collected reviews (positives or negatives).  
The classification step is organized as follows:
\begin{enumerate}
\item First, we apply Na\"{\i}ve Bayes classifier in order to determine the sentiment orientation of the collected reviews, then we consider both the predictions of Na\"{\i}ve Bayes (positive or negative) and their ratings (numeric scale from 1 to 5 or 1 to 10) in the following way:
\begin{enumerate}
\item If the review polarity is predicted as positive and its attached rating is greater than 5, then, the review will be added to the \textit{"positiveReviews"} set.
\item If the review polarity is predicted as negative and its attached rating is less or equal to 5, then, the review will be added to the \textit{"negativeReviews"} set.
\item If the review polarity is predicted as positive and its attached rating is less or equal to 5, then, the review will be added to the \textit{"remainingReviews"} set.
\item If the review polarity is predicted as negative and its attached rating is greater than 5, then, the review will be added to the \textit{"remainingReviews"} set.
\end {enumerate}
\item Second, we apply SVM classifier with a linear kernel to the reviews contained in \textit{"remainingReviews"} set and we compare SVM polarity prediction with Na\"{\i}ve Bayes polarity prediction:
\begin{enumerate}
\item If the review polarity is predicted as positive by both SVM and Na\"{\i}ve Bayes classifiers, then, the review will be added to the \textit{"positiveReviews"} set.
\item If the review polarity is predicted as negative by both SVM and Na\"{\i}ve Bayes classifiers, then, the review will be added to the \textit{"negativeReviews"} set.
\item If the review polarity is predicted as positive (respectively negative) by SVM and negative (respectively positive) by Na\"{\i}ve Bayes classifier, then, the review will be added to the \textit{"remaining1Reviews"} set.
\end {enumerate}
\item Third and finally, the polarity of reviews contained in \textit{"remaining1Reviews"} set will be determined based on their attached ratings:
\begin{enumerate}
\item If the review attached rating is greater than 5, then, the review will be added to the \textit{"positiveReviews"} set.
\item If the review attached rating is less or equal to 5, then, the review will be added to the \textit{"negativeReviews"} set.
\end {enumerate}
\end {enumerate}

Shawe-Taylor and Sun \cite{SHAWETAYLOR20113609} presented Na\"{\i}ve Bayes, Support Vector Machine (SVM) and maximum entropy as the most frequently used models for opinion mining. Furthermore, Pang, Lee, and Vaithyanathan \cite{Pang:2002:TUS:1118693.1118704} were the first to use these 3 models to classify \textbf{movie reviews} as positive or negative. The authors reported that the use of unigrams as features improves the sentiment classification accuracy. Thus, we trained Na\"{\i}ve Bayes and SVM with unigrams (count) for the purpose of predicting reviews polarity. \par
We apply Na\"{\i}ve Bayes before SVM because, firstly, we found that Na\"{\i}ve Bayes achieves 82\% accuracy in predicting sentiment polarity of the collected reviews compared to 77\% accuracy for SVM (Table 2.5), and, secondly, Pang, Lee, and Vaithyanathan \cite{Pang:2002:TUS:1118693.1118704} applied Na\"{\i}ve Bayes and SVM to a movie review dataset (polarity dataset v2.0) \footnote{\url{www.cs.cornell.edu/people/pabo/movie-review-data/review_polarity.tar.gz}} and found that Na\"{\i}ve Bayes achieves higher accuracy in predicting movie reviews polarity compared to SVM. Therefore, we adopt the accurate classifier: Na\"{\i}ve Bayes as the primary classifier, then, we apply SVM classifier when there is a mismatch between Na\"{\i}ve Bayes predicted polarity for a review and its attached rating. \hyperref[sec:a1]{Algorithm 1} provides details about the sentiment classification step.  \par

\begin{algorithm}
\small
    \SetKwInOut{KwIDe}{Define}
    \SetKwInOut{KwIn}{Input}
    \SetKwInOut{KwOut}{Output}
    
    \KwIDe{
		$V=\{v_{1}, v_{2}, ..., v_{n\}}$: the set of ratings attached to each review;
		\\
		$R=\{r_{1}, r_{2}, ..., r_{n\}}$: the set of reviews expressed in natural language;
		\\

		%
		$NBPrediction=\{nbp(r_1) ,nbp(r_2) ,... ,nbp(r_n)\}$: the set of polarity of reviews $R$\\ predicted by Na\"{\i}ve Bayes classifier;
		\\
		$SVMPrediction=\{svmp(r_1), svmp(r_2), ..., svmp(r_p)\}$: the set of polarity of\\ $remainingReviews$ predicted by SVM classifier;
		\\
		$remainingReviews$: the set of indexes for remaining reviews after the first step;
		\\
		$remaining1Reviews$: the set of indexes for remaining reviews after the second step;
		\\
    }
    \KwIn{Reviews and their user ratings: $R$ and $V$.
    }
    \KwOut{The set of positive reviews and the set of negative reviews: $positiveReviews$ and $negativeReviews$.
    }
    Apply Na\"{\i}ve Bayes classifier to predict the polarity of reviews $R$ and store the results in $NBPrediction$.\\
    \For{$i\gets0$ \KwTo $n-1$ \KwBy $1$}{
    \uIf{$nbp(r_i)$ is 'positive' \textbf{and} $v_{i}$ $>$ 5}{
    add $i$ to $positiveReviews$ \;
    }
    \uElseIf{$nbp(r_i)$ is 'negative' \textbf{and} $v_{i}$ $\leq$ 5}{
    add $i$ to $negativeReviews$ \;
    }
    \Else{
    add $i$ to $remainingReviews$ \;
    }
    }
    Apply SVM classifier to predict the polarity of $remainingReviews$ and store the results in $SVMPrediction$.\\
    $lengthRemainingReviews \leftarrow length\;of\;remainingReviews$\\
    \For{$i\gets0$ \KwTo $lengthRemainingReviews$ \KwBy $1$}{
    \uIf{$nbp(r_{remainingReviews[i]})$ is 'positive' \textbf{and} $svmp(r_{remainingReviews[i]})$ is 'positive'}{
    add $remainingReviews[i]$ to $positiveReviews$ \;
    }
    \uElseIf{$nbp(r_{remainingReviews[i]})$ is 'negative' \textbf{and} $svmp(r_{remainingReviews[i]})$ is 'negative'}{
    add $remainingReviews[i]$ to $negativeReviews$ \;
    }
    \Else{
    add $remainingReviews[i]$ to $remaining1Reviews$ \;
    }
    }
\caption{Polarity Classification phase}\label{sec:a1}

\end{algorithm}
\begin{algorithm}
    $lengthRemaining1Reviews \leftarrow length\;of\;remaining1Reviews$\\
    
    \For{$i\gets0$ \KwTo $lengthRemaining1Reviews$ \KwBy $1$}{
    \uIf{$v_{remaining1Reviews[i]}$ $>$ 5}{
    add $remaining1Reviews[i]$ to $positiveReviews$ \;
    }
    \Else{
    add $remaining1Reviews[i]$ to $negativeReviews$ \;
    }
    }
\end{algorithm}

\subsection{Fusion and grouping phase}

In order to overcome the weakness of LSA model in extracting opinions with the same perspective, we apply the opinion fusion and grouping algorithm \cite{yan2017fusing} \textbf{\textit{separately}} for positive and negative reviews after classifying them based on their sentiment polarity, which guarantees that positive opinions will be grouped together and so on. Yan et al. \cite{yan2017fusing} describe the opinion fusion and grouping algorithm as follows: \textit{"By applying Algorithm 1, we fuse and group opinions into several principal opinion sets. The opinions in each set hold a similar or same perspective"} and \textit{"Once the processing based on Algorithm 1 has been completed, the opinions are grouped into a number of K fused principal opinion sets. Meanwhile, we also get the statistics of the principal opinions, i.e., the number of similar opinions in each set, the sum of their ratings and the sum of their similarity"}.

\subsection{Reputation generation}
Based on the result of the grouping phase, we propose to compute a custom reputation value \textbf{\textit{separately}} for positive and negative groups. We tailor the average similarity $S_{k}^{polarity} / N_{k}^{polarity}$ with the average rating value $V_{k}^{polarity} / N_{k}^{polarity}$ of principal opinion set k for a specific polarity. \par
We compute a custom reputation value separately for positive and negative groups by applying this formula:
\begin{equation}
Rep(A^{polarity}) = \frac{1}{K^{polarity}}. \sum_{k=1}^{K^{polarity}} \frac{V_{k}^{polarity} . S_{k}^{polarity}}{N_{k}^{polarity} . N_{k}^{polarity}}
\end{equation}

$K^{polarity}$   : The number of opinion sets for a specific polarity (positive or negative opinions).\\

$N_k^{polarity}$ : The number of similar opinions in principal opinion set k for a specific polarity.\\

$S_k^{polarity}$ : The sum of semantic similarity in principal opinion set k for a specific polarity.\\

$V_k^{polarity}$ : The sum of ratings in principal opinion set k for a specific polarity.\\

In general, the number of positive opinions and the number of negative opinions are not equal, which implies that $O^{positive}$ and $O^{negative}$ don\textit{'}t contribute equally to the final reputation value. Thus, we propose to use the Weighted Arithmetic Mean for computing the final reputation value instead of using the ordinary arithmetic mean:
\begin{equation}
finalRep(A) = \frac{\sum_{polarity}^{[positive,negative]} Rep(A^{polarity}) . O^{polarity}}{\sum_{polarity}^{[positive,negative]} O^{polarity}}
\end{equation}
That leads to :
\begin{equation}
finalRep(A) = \frac{Rep(A^{positive}) . O^{positive} + Rep(A^{negative}) . O^{negative}}{O^{positive} + O^{negative}}
\end{equation}

$O^{positive}$   : The number of positive opinions.\\

$O^{negative}$   : The number of negative opinions.\\

$A^{positive}$   : Groups of positive opinions.\\

$A^{negative}$   : Groups of negative opinions.\\

$Rep(A^{positive})$ : Custom reputation value for groups of positive opinions.\\

$Rep(A^{negative})$ : Custom reputation value for groups of negative opinions.

\section{Experimental results and discussion}
\subsection{Training phase}
As we mentioned earlier, we used Na\"{\i}ve Bayes and Linear Support Vector Machine classifiers, both are trained with a movie review dataset (polarity dataset v2.0) \footnote{\url{www.cs.cornell.edu/people/pabo/movie-review-data/review_polarity.tar.gz}} that contains 1000 positive and 1000 negative reviews. The sentences in the corpus are processed and downcased. \par
The chosen features were unigrams (count) for Na\"{\i}ve Bayes classifier and unigrams TF-IDF (Term frequency - Inverse document frequency) weighted word frequency features for SVM. \par

\subsection{Datasets collection and preprocess}
In order to evaluate the proposed reputation system, we need a dataset that contains both reviews and ratings. However, there is no standard dataset suitable for our evaluation, therefore, we extracted reviews from IMDb\footnote{\url{www.imdb.com}} (Internet Movie Database) website which represents the world\textquotesingle s most popular and authoritative source for movie, TV, and celebrity content. Figure 2.2 represents a sample of the dataset without review polarity\\
\begin{figure}[H]
\centering
\includegraphics[scale=0.7]{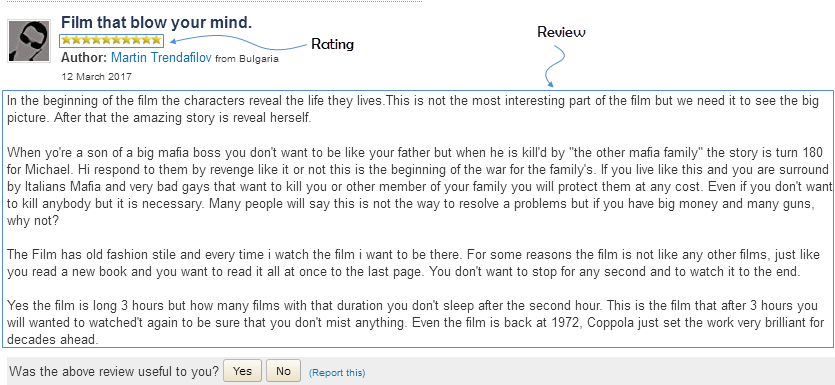}
\caption{Dataset Sample}
\end{figure}
We manually created 10 datasets for 10 different movies, each one contains 100 reviews [comment + rating + sentiment polarity (manually annotated)] randomly extracted, and we verified that the datasets are representatives based on IMDb users weighted average vote. The statistical information of datasets is shown in Table 2.3.

\begin{table}[htb]
\centering
\caption{Statistical information of Datasets}
\begin{tabular}{l|c} 
	\hline
	The total number of movie descriptions & 10 \\
	The total number of reviews and ratings (i.e., opinions) & 1000 \\
	The number of reviews per movie & 100\\
	\hline 
\end{tabular}
\end{table}

You may wonder why collecting 100 reviews for each entity instead of using all the reviews attached to it? Because not every single user who rates the entity (movie) shares its review toward it in the platform (IMDb, Amazon, etc...). Besides, the proposed approach exploits both reviews and ratings, so, if we consider all reviews attached to an entity (movie), we can encounter cases where a movie has a high IMDb weighted average vote, but only users that hold a negative point of view toward it, decide to share their opinions (review + rating), and vice versa. Therefore, those opinions are biased (not representative), which will lead to an unfair reputation value toward the target entity. Thus, we ensure that the ratings are representative during the manual gathering of reviews. Table 2.4 contains more details about datasets.\\

\begin{table}[htb]
\centering
\caption{Datasets details}
{\renewcommand{\arraystretch}{1.2}
\scalebox{1}{
\begin{tabular}{|c|c|c|} \hline
Movie & IMDb weighted average vote & Dataset average rating \\ \hline
2012 (Dataset 1) & 5.8 (301440 users) & 5.8 \\ \hline
A Beautiful Mind (Dataset 2) & 8.2 (672962 users) & 8.2 \\ \hline
Amadeus (Dataset 3) & 8.3 (295560 users) & 8.3 \\
\hline
Avatar (Dataset 4) & 7.8 (946090 users) & 7.8 \\
\hline
Clash of the Titans (Dataset 5) & 5.8 (240456 users) & 5.8 \\
\hline
Les Miserables (Dataset 6) & 7.6 (260252 users) & 7.6 \\
\hline
Star Wars Episode I (Dataset 7) & 6.5 (579916 users) & 6.5 \\
\hline
The Expendables (Dataset 8) & 6.5 (286373 users) & 6.5 \\
\hline
The Godfather (Dataset 9) & 9.2 (1257206 users) & 9.2 \\
\hline
The Matrix Revolutions (Dataset 10) & 6.7 (384761 users) & 6.7 \\
\hline
\end{tabular}
}
}
\end{table}

According to Table 2.4, all datasets have an average rating equal to IMDb weighted average vote. \par
After collecting the reviews, we applied a preprocessing algorithm to remove word segmentation and stop words. \par
In their work, Yan et al. \cite{yan2017fusing} developed a web spider to collect recent product descriptions, customer reviews, and review ratings from Amazon China (https://www.amazon.cn) and Amazon English (https://www.amazon.com). Each rating is associated with a review and subjectively given by customers. The number of reviews per product is between 80 and 100. \par

\subsection{Sentiment classification}
After training Na\"{\i}ve Bayes and Linear Support Vector Machine classifiers, we separate reviews into positive and negative based on their sentiment polarity by applying \hyperref[sec:a1]{Algorithm 1}. \par
We need to evaluate the effectiveness of \hyperref[sec:a1]{Algorithm 1} in classifying reviews. Thus, we compared it to Na\"{\i}ve Bayes and SVM classifiers. Figure 2.3 shows the accuracy of the proposed classification phase compared to Na\"{\i}ve Bayes and SVM for each dataset. 
\begin{figure}[H]
\centering
\includegraphics[scale=0.63]{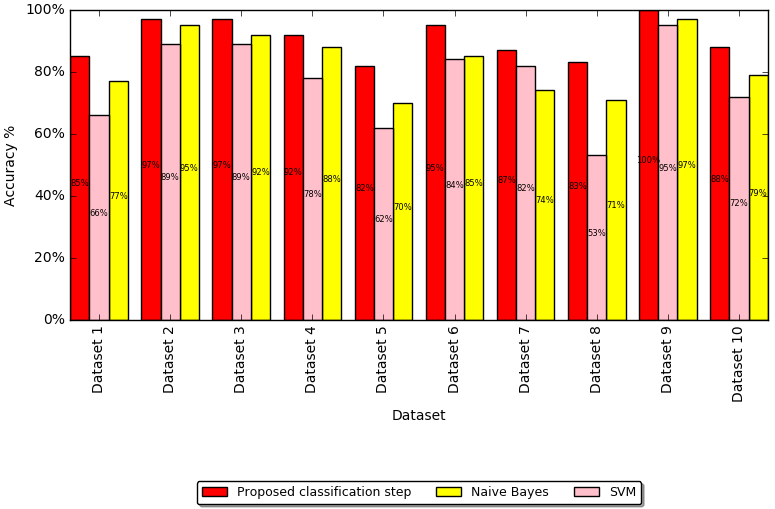}
\caption{Sentiment classification accuracy of Algorithm 1, Na\"{\i}ve Bayes and SVM}
\end{figure}

We can see from Figure 2.3 that the combination of Na\"{\i}ve Bayes and SVM classifiers (\hyperref[sec:a1]{Algorithm 1}) gives the highest sentiment classification accuracy for all datasets compared to the separate use of each classifier. Table 2.5 depicts precision, recall, f-score, and accuracy of Na\"{\i}ve Bayes, SVM and \hyperref[sec:a1]{Algorithm 1} applied to all 1000 reviews of datasets for sentiment analysis.

\begin{table}[htb]
\centering
\caption{Precision, Recall, F-score, and accuracy of Na\"{\i}ve Bayes, SVM and \hyperref[sec:a1]{Algorithm 1} for all reviews}
{\renewcommand{\arraystretch}{1}
\scalebox{1}{
\begin{tabular}{|| c || c || c || c || c || c ||} \hline \hline
	Approach
	& Polarity
	& Precision
	& Recall
	& F-score
	& Accuracy
	 \\ \hline \hline
	\multirow{2}{*}{Naive Bayes} & Positive & 0.88 & 0.91 & 0.89 & \multirow{2}{*}{0.82}\\
\cline{2-5}
& Negative & 0.54 & 0.46 & 0.5 &  \\ \hline \hline
	\multirow{2}{*}{SVM} & Positive & 0.94 & 0.76 & 0.84 & \multirow{2}{*}{0.77}\\
\cline{2-5}
& Negative & 0.43 & 0.79 & 0.56 &  \\ \hline \hline
	\multirow{2}{*}{\hyperref[sec:a1]{Algorithm 1}} & Positive & \textbf{0.95} & \textbf{0.92} & \textbf{0.94} & \multirow{2}{*}{\textbf{0.90}}\\
\cline{2-5}
& Negative & \textbf{0.71} & \textbf{0.81} & \textbf{0.76} &  \\ \hline \hline
\end{tabular}
}
}
\end{table}

We can see from Table 2.5 that our proposed classification method (\hyperref[sec:a1]{Algorithm 1}) outperforms Na\"{\i}ve Bayes and Linear Support Vector Machine (LSVM) in terms of precision, recall, f-score, and accuracy for both positive and negative reviews.

\subsection{Opinion fusion and grouping}
During this phase, the reviews can be grouped into several opinion sets after classifying them into positive and negative based on their sentiment polarity. At the end of this process, we acquire some statistics such as the sum of semantic similarity $S_k^{polarity}$, the sum of ratings $V_k^{polarity}$ and the number of similar opinions $N_k^{polarity}$ for each set. Table 2.6 provides example results of the grouping phase.
\begin{table}[htb]
\centering
\caption{Opinion fusion and grouping example results.}
\scalebox{1.3}{
\begin{tabular}{|| *4c || *4c ||} \hline \hline
\multicolumn{4}{||c}{Positive Reviews} & \multicolumn{4}{c||}{Negative Reviews} \\ \hline \hline
Set & Sim & Rat & Num & Set & Sim & Rat & Num \\ \hline \hline
$R_{1}$ & 66.86 & 569 & 67 & $R_{1}$ & 3.99 & 16 & 4 \\
$R_{2}$ & 3.99 & 33 & 4 & $R_{2}$ & 1 & 3 & 1 \\
$R_{3}$ & 13.98 & 113 & 14 & & & & \\
$R_{4}$ & 7.99 & 69 & 8 & & & & \\ 
$R_{5}$ & 1 & 8 & 1 & & & & \\ 
$R_{6}$ & 1 & 9 & 1 & & & & \\ \hline \hline
\end{tabular}
}
\end{table}

We denote:\\

Set: the principal review set by fusing and grouping reviews.\\

Sim: the sum of semantic similarity in a principal opinion set.\\

Rat: the sum of ratings in a principal opinion set.\\

Num: the number of similar reviews in a principal opinion set.\\
\\

From Table 2.6, we can see that reviews are grouped into several sets $R_{i}$ after fusing and grouping them (each set contains one review at least). \par
It is very important to provide customers with sufficient information to help them make an informed decision toward the target entity. Therefore, in addition to the numerical reputation value, we provide them with a pie chart (Figure 2.4) that depicts the distribution of sentiments over reviews (based on ratings).
\begin{figure}[H]
\centering
\includegraphics[scale=0.75]{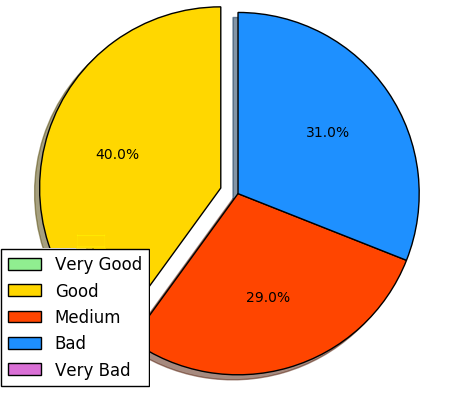}
\caption{Distribution of sentiments over reviews}
\end{figure}
As illustrated in Figure 2.4, 40\% of users are satisfied with the movie, 29\% of them think that the movie is \textit{"OK"} and the remaining 31\% of them dissatisfy with the target movie.

\subsection{Reputation generation}
In their work, Yan et al. \cite{yan2017fusing} have recruited ten volunteers to manually rate 1000 products after reading their descriptions and reviews. They averaged the volunteers' ratings (manually generated reputation values) toward each product, then, they computed the deviations between the manually generated reputation values and reputation values generated by their approach. However, we believe that even if they have recruited 1000000 volunteers, there is no guarantee that their judgment is right toward the target products, since they may not have enough knowledge or experience to give proper judgment toward them. For that, we have chosen the IMDb weighted average vote as ground truth for the reason that IMDb is a platform where both regular movie fans and expert reviewers rate and share their opinions. We can see in Table 2.4 that more than 1257206 people gave a rating toward the Godfather 1 movie, which represent an enormous amount. Besides, the vast majority of those people share their ratings after watching the target movie, which imply that they have knowledge and opinion toward it, Thus, in order to evaluate the effectiveness of our method in generating reputation, we compare the final reputation value generated by our method and Yan et al. \cite{yan2017fusing} with IMDb users weighted average vote applied by IMDb to represent a rating toward the target movie, which is a number ranging from 1 to 10 as shown in Figure 2.5.

\begin{figure}[H]
\centering
\includegraphics[scale=0.9]{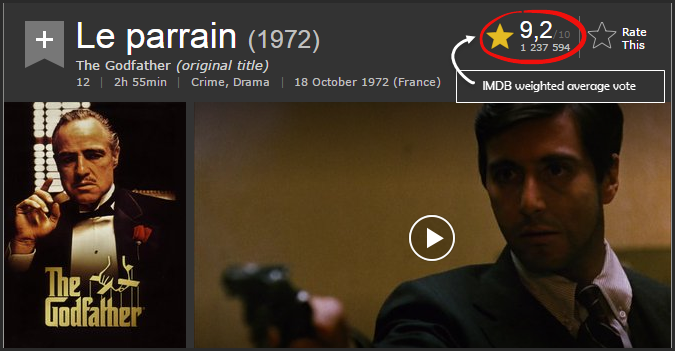}
\caption{IMDb users weighted average vote for The Godfather I movie}
\end{figure}

We compare our approach to Yan et al. work \cite{yan2017fusing} by applying them to the datasets that we collected before. We varied the value of the opinion fusion threshold $t_{0}$ from 0.05 to 0.95 with a step of 0.05, and we used the evaluation measure \textit{"Absolute Error"} where:

\begin{equation}
AER_{t_{0}}^{m_{i}} = |IMDbWAV^{m_{i}} - Rep_{t_{0}}^{m_{i}}|
\end{equation}

We denote:\\

$AER_{t_{0}}^{m_{i}}$   : Reputation Absolute Error for a specific $t_{0}$ value toward the target movie $m_{i}$.\\

$IMDbWAV^{m_{i}}$      : IMDb weighted average vote toward the target movie $m_{i}$.\\

$Rep_{t_{0}}^{m_{i}}$  : Reputation value computed by our method or Yan et al. \cite{yan2017fusing} for a specific ${t_{0}}$ value toward the target movie $m_{i}$.\\

Figures 2.6, 2.7, 2.8, 2.9, and 2.10 represent comparison results between the two methods

\begin{figure}[H]
\centering
\includegraphics[width=1.0\textwidth]{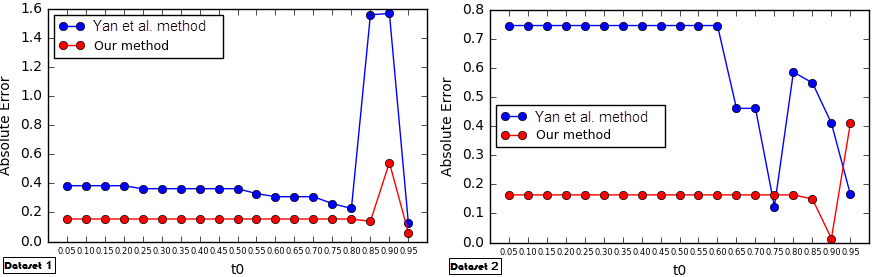}
\caption{Comparison between Yan et al. \cite{yan2017fusing} method and our contribution for dataset 1 and dataset 2}
\end{figure}
\begin{figure}[H]
\centering
\includegraphics[width=1.0\textwidth]{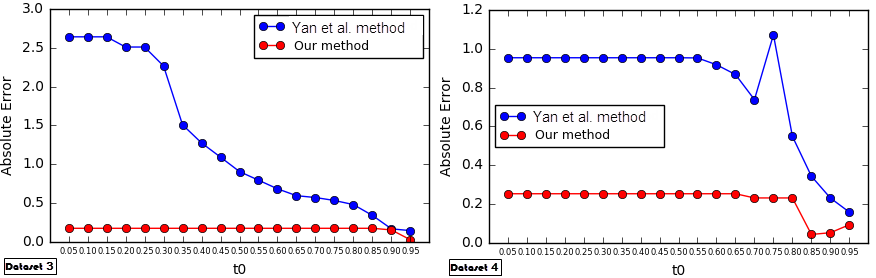}
\caption{Comparison between Yan et al. \cite{yan2017fusing} method and our contribution for dataset 3 and dataset 4}
\end{figure}
\begin{figure}[H]
\centering
\includegraphics[width=1.0\textwidth]{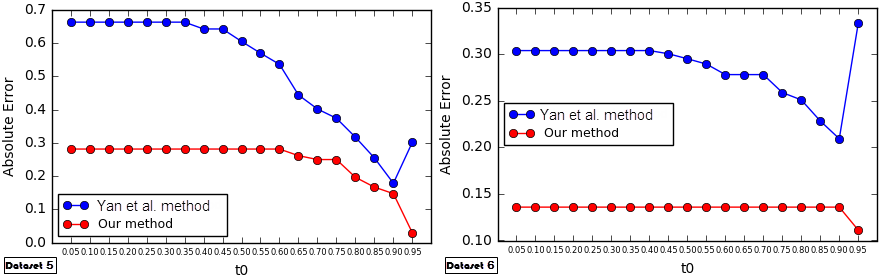}
\caption{Comparison between Yan et al. \cite{yan2017fusing} method and our contribution for dataset 5 and dataset 6}
\end{figure}
\begin{figure}[H]
\centering
\includegraphics[width=1.0\textwidth]{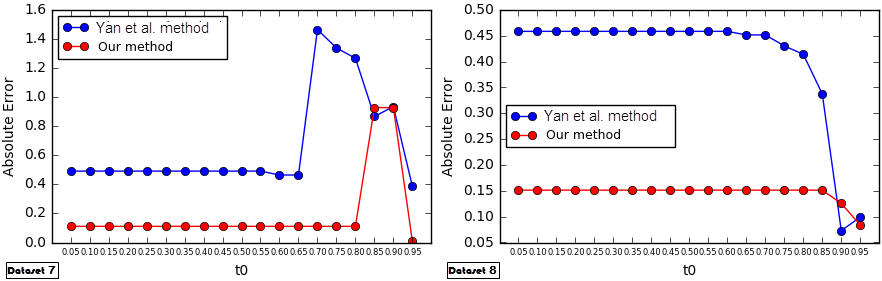}
\caption{Comparison between Yan et al. \cite{yan2017fusing} method and our contribution for dataset 7 and dataset 8}
\end{figure}
\begin{figure}[H]
\centering
\includegraphics[width=1.0\textwidth]{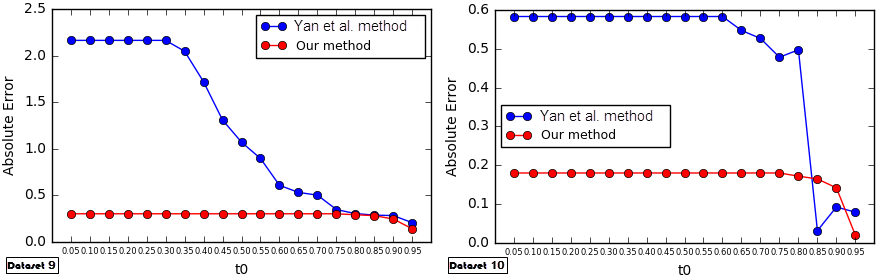}
\caption{Comparison between Yan et al. \cite{yan2017fusing} method and our contribution for dataset 9 and dataset 10}
\end{figure}
In Figure 2.6: dataset 1, Figure 2.7: dataset 3 and 4, Figure 2.8: dataset 5 and 6, and Figure 2.10: dataset 9, our method outperforms Yan et al. \cite{yan2017fusing} method for all values of $t_{0}$ varied from 0.05 to 0.95 with a step of 0.05, additionally, we can see that for all datasets, our method works better than Yan et al. approach \cite{yan2017fusing} when 0.05$\leq t_{0} \leq$0.70 , even in the worst cases, Figure 2.6: dataset 1 when $t_{0}=0.90$, dataset 2 when $t_{0}=0.95$, and Figure 2.9: dataset 7 when $t_{0}=\{0.85,0.90\}$, the Absolute Error (2.4) of our method doesn\textit{'}t exceed 0.97. Table 2.7 summarizes the results given by the two methods.

\begin{table}[htb]
\centering
\caption{Experiment results summarization}
\scalebox{0.87}{
\begin{tabular}{|c|c|c|c|c|c|} \hline
	& \multicolumn{1}{|p{2cm}|}{\centering IMDb weighted \\ average vote } 
	& \multicolumn{1}{|p{3cm}|}{\centering Average reputation \\(Yan et al. \cite{yan2017fusing} method)}
	&\multicolumn{1}{|p{3cm}|}{\centering $MAER^{m_{i}}$ \\ (Yan et al. \cite{yan2017fusing} method)} 
	& \multicolumn{1}{|p{3cm}|}{\centering Average reputation \\ (Our method)} 
	& \multicolumn{1}{|p{3cm}|}{\centering $MAER^{m_{i}}$ \\ (Our method)} 
	 \\ \hline
	Dataset 1 & 5.80 & 5.68401533142 & 0.458584523451 & 5.69411656318 & \textbf{0.169344478854}
	 \\ \hline
	Dataset 2 & 8.20 & 7.58329159051 & 0.616708409489 & 8.033252139 & \textbf{0.168193964067} \\ \hline
	Dataset 3 & 8.30 & 7.0236673294 & 1.2763326706 & 8.14088245291 & \textbf{0.161469513573} \\
	\hline
	Dataset 4 & 7.80 & 6.99168965196 & 0.808310348038 & 7.60117164592 & \textbf{0.218177388535} \\
	\hline
	Dataset 5 & 5.80 & 5.27820594837 & 0.521794051633 & 5.55387542084 & \textbf{0.246124579156}\\
	\hline
	Dataset 6 & 7.60 & 7.31385618805 & 0.286143811945 & 7.46540816442 & \textbf{0.134591835581}\\
	\hline
	Dataset 7 & 6.50 & 6.26579054984 & 0.662862169432 & 6.30898667249 & \textbf{0.191809871998} \\
	\hline
	Dataset 8 & 6.50 & 6.09915992375 & 0.408525260606 & 6.35358792992 & \textbf{0.146412070077} \\
	\hline
	Dataset 9 & 9.20 & 7.98707747356 & 1.21292252644 & 8.91509031525 & \textbf{0.284909684753} \\
	\hline
	Dataset 10 & 6.70 & 6.26880591164 & 0.486749395551 & 6.53429997883 & \textbf{0.167701915098} \\
	\hline
	\end{tabular}
}
\end{table}

From Table 2.7, we can see that our method provides the nearest reputation value to the ground truth (IMDb weighted average vote) for all datasets, also, the $MAER^{m_{i}}$ (2.5) of our method doesn\textit{'}t exceed 0.29. However, the $MAER^{m_{i}}$ reaches 1.27 in the third dataset and 1.21 in the ninth dataset for Yan et al. \cite{yan2017fusing} method. Moreover, the ninth dataset contains reviews of The Godfather I movie which is considered as one of the best movies of all time. Thus, this high error value could influence the popularity of the movie, which makes reputation generation a very delicate task.\\
We compute the $MAER^{m_{i}}$ by applying this formula:

\begin{equation}
MAER^{m_{i}} = \frac{1}{T} . \sum_{\substack{min\leq t_{0} \leq max\\t_{0} \, multiple \, of \, step}} AER_{t_{0}}^{m_{i}}
\end{equation}

We denote:\\

$MAER^{m_{i}}$ : Reputation Mean Absolute Error toward the target movie $m_{i}$ for all $t_{0}$ values.\\

$AER_{t_{0}}^{m_{i}}$   : Reputation Absolute Error for a specific $t_{0}$ value toward the target movie $m_{i}$.\\

$T$   : Total number of $t_{0}$ values, in our case, it varies from 0.05 to 0.95 with a step of 0.05, which means $t_{0}=\{0.05,0.10,0.15,0.20,0.25,0.30,0.35,0.40,0.45,0.50,0.55,0.60,0.65,0.70,0.75,0.80,0.85,\\0.90,0.95\}$ and $T=19$.\\

\subsection{Opinion fusion threshold $t_{0}$}
After comparing the two methods and showing the effectiveness of ours, we can see that different values of the opinion fusion threshold $t_{0}$ could lead to different grouping results, which cause different custom reputation values and different final reputation values. Therefore, choosing a suitable $t_{0}$ value is very important. Thus, we conducted experiments to determine the value of $t_{0}$ that provides the best results. We set the value of the opinion fusion threshold $t_{0}$ from 0.05 to 0.95 with a step of 0.05. Figure 2.11 shows the Absolute Error (AE) (2.4) of the final reputation value computed by our method.
\begin{figure}[H]
\centering
\includegraphics[width=1.0\textwidth]{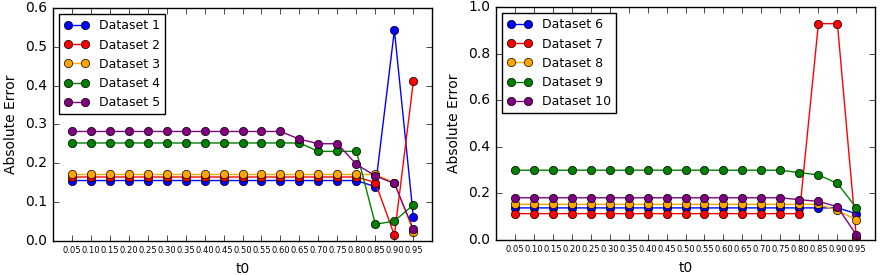}
\caption{Absolute Error (AE) of reputation value computed by our method}
\end{figure}
From Figure 2.11, we observed that for all datasets, the Absolute Error (2.4) value is stable when 0.05$\leq t_{0} \leq$0.6, and then, it begins to vary either by increasing or decreasing. But, it is still hard to determine the suitable $t_{0}$ value for generating an accurate reputation toward the target movie. That\textit{'}s why we conducted more experiments. We computed the $MAER_{t_{0}}$ (2.6) between IMDb weighted average votes (ground truth) and the reputation values generated by our method using all movies of datasets for each $t_0$ value between 0.05 and 0.95 with a step of 0.05.
\begin{table}[!htb]
\centering
\caption{$MAER_{t_{0}}$ between IMDb weighted average votes and the reputation values computed by our method using all the reviews of the 10 movies for $t_0$ value varied from 0.05 to 0.95 with a step of 0.05}
\scalebox{0.95}{
\begin{tabular}{|c|c|} 
	\hline
	Opinion fusion threshold $t_{0}$ value & $MAER_{t_{0}}$ \\
	\hline
	$t_{0}=0.05$ & 0.18978855 \\\hline
	$t_{0}=0.10$ & 0.18978855 \\\hline
	$t_{0}=0.15$ & 0.18978855 \\\hline
	$t_{0}=0.20$ & 0.18978855 \\\hline
	$t_{0}=0.25$ & 0.18978855 \\\hline
	$t_{0}=0.30$ & 0.18978855 \\\hline
	$t_{0}=0.35$ & 0.18978855 \\\hline
	$t_{0}=0.40$ & 0.18978855 \\\hline
	$t_{0}=0.45$ & 0.18978855 \\\hline
	$t_{0}=0.50$ & 0.18978855 \\\hline
	$t_{0}=0.55$ & 0.18978855 \\\hline
	$t_{0}=0.60$ & 0.18978855 \\\hline
	$t_{0}=0.65$ & 0.18780241 \\\hline
	$t_{0}=0.70$ & 0.18445943 \\\hline
	$t_{0}=0.75$ & 0.18445943 \\\hline
	$t_{0}=0.80$ & 0.17724106 \\\hline
	$t_{0}=0.85$ & 0.23243659 \\\hline
	$t_{0}=0.90$ & 0.24759925 \\\hline
	$t_{0}=0.95$ & \textbf{0.09713621} \\\hline
	\hline 
\end{tabular}
}
\end{table}

\begin{figure}[htb]
\centering
\includegraphics[scale=1]{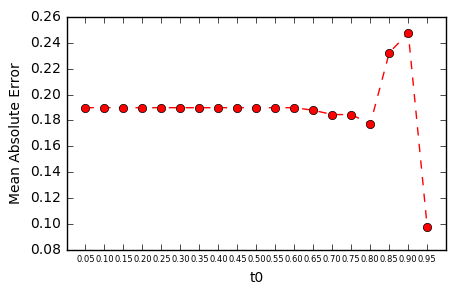}
\caption{$MAER_{t_{0}}$ between IMDb weighted average votes and the reputation values computed by our method using all the reviews of 10 movies for $t_0$ varied from 0.05 to 0.95 with a step of 0.05}
\end{figure}
Both Figure 2.12 and Table 2.8 show that the $MAER_{t_{0}}$ (2.6) is stable when $t{_0} \leq 0.60$. We also found that our reputation generation method performs better when $t_{0}=0.95$, since the Mean Absolute Error between IMDb weighted average votes and the values computed by formula (2.3) using all the reviews of the 10 movies reaches its minimum.\\
We compute the $MAER_{t_{0}}$ by applying this formula:
\begin{equation}
MAER_{t_{0}} = \frac{1}{E} . \sum_{\substack{1\leq i \leq E}} AER_{t_{0}}^{m_{i}}
\end{equation}
We denote:\\

$MAER_{t_{0}}$ : Mean Absolute Error between IMDb weighted average votes (ground truth) and the reputation values generated by our method for all movies of datasets for a specific $t_{0}$ value.\\

$E$ : number of target entities (In this case $E=10$).\\

\subsection{Classification phase impact on reputation generation}
As shown in Figure 2.3, Our proposed classification step gives the highest sentiment classification accuracy for all datasets compared to Na\"{\i}ve Bayes and SVM classifiers. It is important to analyze the impact of the classification step on reputation generation. For this purpose, we conducted more experiments in which we compared our method to two other approaches. The first approach uses Na\"{\i}ve Bayes for the classification phase, and the second one uses SVM. For each approach, we computed the $MAER^{m_{i}}$ between IMDb weighted average vote and the final reputation value ($t_{0}=0.95$) computed by each approach for all datasets using formula (2.5).

\begin{figure}[!htb]
\centering
\includegraphics[scale=1]{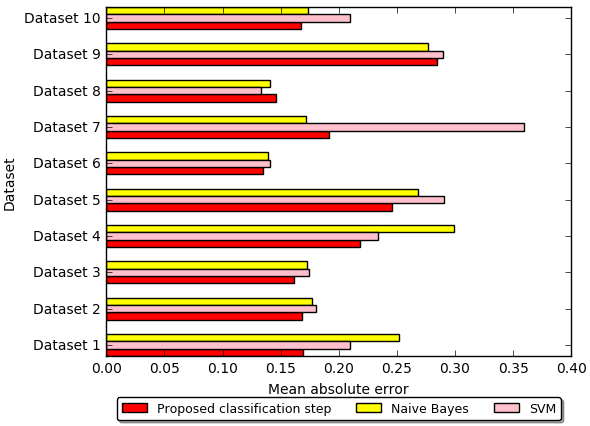}
\caption{Reputation Mean Absolute Error for the three approaches}
\end{figure}

We can see from Figure 2.13 that our proposed classification step leads to the lowest reputation Mean Absolute Error value (2.5) for datasets 1, 2, 3, 4, 5, 6, and 10, more than that, there is a very small difference between the $MAER^{m_{i}}$ of our approach and the $MAER^{m_{i}}$ computed during the use of Na\"{\i}ve Bayes for the classification phase in datasets 7, 8 and 9. Thus, we conclude that the use of an accurate sentiment classifier leads to an accurate reputation value toward the target entity.

\section{Summary and Limitations}

In this chapter, we propose a reputation generation system that combines Na\"{\i}ve Bayes and Linear Support Vector Machine (LSVM) classifiers for the purpose of separating online reviews into positives or negatives (sentiment analysis) before grouping them into different opinion sets based on their semantic relations, we also propose to compute the final reputation value using the Weighted Arithmetic Mean. As has been noted, our approach improves the work proposed in \cite{yan2017fusing} by generating accurate reputation values toward different movies. However, the main issues of our work are:
\begin{itemize}
\item Polarity classification step: the classifier depends on the domain of use (politics, economic, business, sport...) and the language (English, French, Chinese, Arabic...). Consequently, it is not practical to use different datasets for each domain and each language. Fortunately, there are many methods to handle those issues such as Machine Translation, Transfer Learning, etc... \cite{Sun:2017:RNL:3174385.3174741}.
\item Useless reviews: a lot of reviews are irrelevant to the target entity (ads, spam). Therefore, adding a filtering phase \cite{Sun:2017:RNL:3174385.3174741, Wang:2015:ARR:2844746.2844886} will reduce the processing time and will increase the accuracy of the reputation system (the system will deal only with relevant and useful reviews).
\end{itemize}
The next chapter will describe MTVRep, a movie and TV show reputation system that exploits fine-grained sentiment analysis and semantic analysis for the purpose of generating and visualizing reputation toward movies and TV shows.

\chapter{Fine-grained Opinion Mining and Semantic Analysis for Reputation Generation} 
\section{Introduction}
The exponential growth of Web 2.0 has dramatically impacted the evolution of e-commerce platforms \cite{10.1016/j.ipm.2016.12.002, bafna2013feature, zhuang2006movie, hou2019mining}. On the one hand, some recent statistics show that 72\% of customers will not take action until they read reviews, and only 6\% of consumers don't trust customer reviews at all, on the other hand, the number of user-generated reviews attached to an online entity could easily exceed thousands \cite{10.5555/1597148.1597269, 10.1007/s00500-017-2882-2}. Thus, a potential customer doesn't have time or effort to examine all the reviews manually in order to make an informed decision toward it \cite{pecar-2018-towards, 10.5555/1863190.1863201}. \par
Little research has been conducted in mining customer and user reviews with regard to feature-based summarization and reputation generation for the purpose of supporting customers' decision making process in E-commerce (buying, renting, booking \dots). Over the last two decades, few opinion summarizer systems have been proposed to produce an aspect-based summary for product reviews \cite{10.1145/1014052.1014073}, movie reviews \cite{zhuang2006movie}, hotel reviews \cite{10.1016/j.ipm.2016.12.002} and local service reviews \cite{34368}. Backing to the reputation generation task, to the best of our knowledge, there are very few reputation systems that have been proposed to compute a single reputation value toward different entities based on fusing and mining user and customer reviews expressed in natural language \cite{yan2017fusing, benlahbib2019unsupervised, benlahbib2019hybrid, 9068916}. Yan et al. \cite{yan2017fusing} applied opinion mining and fusion techniques on product reviews. Benlahbib \& Nfaoui \cite{benlahbib2019unsupervised} used K-Means clustering algorithm on movie reviews. The same authors \cite{benlahbib2019hybrid} incorporated semantic and sentiment analysis to generate reputation by mining user and customer reviews expressed in natural language (English). \par
An important issue that was neglected in the past research on reputation generation is identifying the sentiment strength during the phase of sentiment classification and opinion fusion. In fact, existing works have only focused on classifying reviews into positive or negative before generating a single reputation value, disregarding the sentiment strength. \par
In this chapter, we propose MTVRep, a movie and TV show reputation system that applies fine-grained opinion mining to separate reviews into five opinion groups: strongly negative, weakly negative, neutral, weakly positive, and strongly positive. Then, it computes a custom score for each group based on the acquired statistics of each group, i.e., the number of reviews in each group, the sum of their ratings, and the sum of their semantic similarity (ELMo and cosine metric). Finally, a numerical reputation value is produced toward the target movie or TV show using the weighted arithmetic mean. \par
In this manner, this study addressed the following research question: with the combination of fine-grained opinion mining and semantic analysis, can the proposed reputation system offer better results in terms of reputation generation than the previous reputation systems (consider only semantic relations)? \par

\section{Problem definition}

This section covers the necessary background for understanding the remainder of this chapter, including the problem definition. \par
In this chapter, we face the problem of generating reputation for movies and TV shows by combining fine-grained opinion mining and semantic analysis. Given a set of reviews $R=\{r_{1},\ r_{2},\ \dots,\ r_{n}\}$ expressed for an entity $E$, the set of their attached ratings $V=\{v_{1},\ v_{2},\ \dots,\ v_{n}\}$ where $v_{i} \in [1,5]$ or $v_{i} \in [1,10]$ depending on the rating system. The goal is to separate reviews into five opinion groups: strongly negative, weakly negative, neutral, weakly positive, and strongly positive by applying the Multinomial Na\"ive Bayes model where $G^{polarity}=\{r_{1}^{polarity},\ r_{2}^{polarity},\ \dots,\ r_{m}^{polarity}\}$ is the opinion group that contains reviews that hold the sentiment orientation $polarity$ and $V^{polarity}=\{v_{1}^{polarity},\ v_{2}^{polarity},\ \dots,\ v_{m}^{polarity}\}$ is the set of their numerical ratings, then, to compute a custom score $CS(G^{polarity})$ for each opinion group $G^{polarity}$ by considering some statistics of each opinion group such as $SS^{polarity}$: the sum of semantic similarity for $G^{polarity}$, $SV^{polarity}$: the sum of ratings for $G^{polarity}$, $N^{polarity}$: the number of reviews in $G^{polarity}$, and $maxR$: the highest value of user ratings (5 or 10) depending on the rating scale (1 to 5 or 1 to 10). Finally, we compute the reputation value $Rep(E)$ toward the target entity $E$ using the Weighted Arithmetic Mean. Table 3.1 presents the descriptions of notations used in the rest of this chapter.

\begin{table}[!htb]
\caption{Symbol denotation}
\centerline{
\begin{tabular}{| l | p{14cm} |}
\hline
Symbol & Description                                                              \\ \hline

$E$      & The target entity                                                             \\
$R$      & The set of reviews expressed for the entity $E$                                                                  \\
$V$      & The set of ratings expressed for the entity $E$                                                                  \\
$G^{polarity}$ & The opinion group that contains reviews that hold the sentiment orientation $polarity$  
                            \\

$V^{polarity}$ &  The set of ratings attached to $G^{polarity}$ reviews\\

$SS^{polarity}$ & The sum of semantic similarity for $G^{polarity}$
            \\

$SV^{polarity}$ & The sum of ratings for $G^{polarity}$   \\

$N^{polarity}$ & The number of reviews in $G^{polarity}$ \\

$maxR$ & The highest value of user ratings (5 or 10) depending on the rating scale (1 to 5 or 1 to 10) \\

$CS(G^{polarity})$ & The custom score for opinion group $G^{polarity}$ \\

$Rep(E)$ & The reputation value toward entity $E$  \\ \hline                                                             
\end{tabular}
}
\end{table}

\section{Proposed system}

\subsection{System overview}

The proposed approach consists mainly of four steps:
\begin{enumerate}
	\item We collect movie and TV show reviews from IMDb\footnote{\url{https://www.imdb.com/}} website using the web scraping tool ScrapeStorm\footnote{\url{https://www.scrapestorm.com/}}, then, we preprocess them.
	\item We train the Multinomial Na\"ive Bayes model on the 5-class Stanford Sentiment Treebank (SST-5) dataset in order to perform fine-grained sentiment analysis. The model classifies the collected reviews into five opinion groups: strongly negative, weakly negative, neutral, weakly positive, and strongly positive.
	\item For each opinion group, we acquire the sum of user ratings and the sum of reviews' semantic similarity. The semantic similarity between two reviews is computed as the cosine between their deep contextualized word embeddings (ELMo). These acquired statistics are used to compute a custom score for each opinion group.
	\item We compute the movie or TV show numerical reputation value based on the opinion groups' scores by applying the weighted arithmetic mean. 
\end{enumerate}

Figure 3.1 illustrates the work-flow of the reputation system (MTVRep).\\

\begin{figure}[!h]
\centering
 \includegraphics[scale=1]{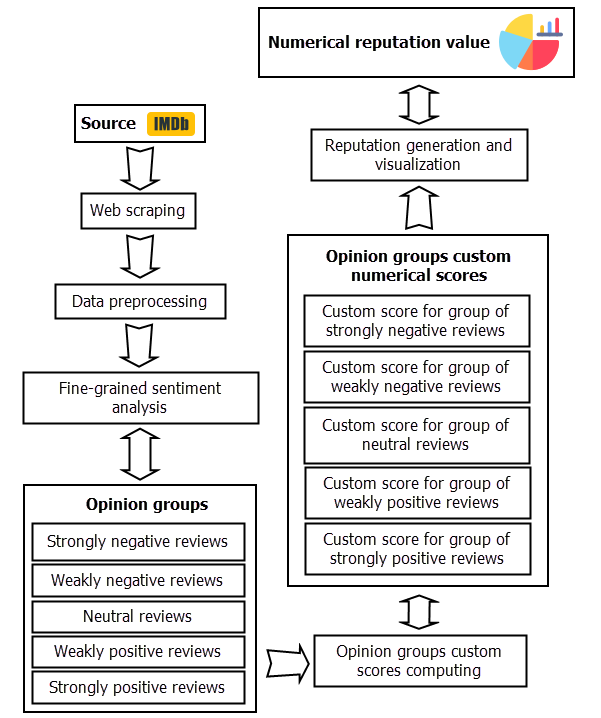}
  \caption{Reputation system pipeline}
\end{figure}

\subsection{Fine-grained sentiment analysis}

We classify the collected reviews into five opinion groups based on their sentiment intensities by applying the Multinomial Na\"ive Bayes model trained on the 5-class Stanford Sentiment Treebank (SST-5) dataset. The reasons behind using the Multinomial Na\"ive Bayes model are discussed in subsection 3.4.2.

\subsection{Opinion groups custom scores}

After separating movie and TV show reviews into five opinion groups: strongly negative, weakly negative, neutral, weakly positive, and strongly positive, we compute a custom score for each opinion group based on the sum of their ratings and the sum of their reviews' semantic similarity. The statistics of opinion groups are acquired by applying algorithm 2.

\newcommand\mycommfont[1]{\small\ttfamily\textcolor{blue}{#1}}
\SetCommentSty{mycommfont}
\begin{algorithm}[!htb]
    \SetKwFunction{isOddNumber}{isOddNumber}
    \SetKwInOut{KwIDe}{Define}
    \SetKwInOut{KwIn}{Input}
    \SetKwInOut{KwOut}{Output}
    
    \KwIDe{
     $G^{polarity}=\{r_{1}^{polarity},\ r_{2}^{polarity},\ \dots,\ r_{m}^{polarity}\}$: the opinion group that contains reviews that hold the sentiment orientation $polarity$.\\
     $V^{polarity}=\{v_{1}^{polarity},\ v_{2}^{polarity},\ \dots,\ v_{m}^{polarity}\}$: the set of ratings \\attached to $G^{polarity}$ reviews.\\
     $SS^{polarity}$: the sum of semantic similarity for $G^{polarity}$ reviews.\\
     $SV^{polarity}$: the sum of ratings for $G^{polarity}$ reviews.\\
     $N^{polarity}$: the number of reviews in $G^{polarity}$.\\
     $ELMo(r^{polarity}_{i})$: ELMo embeddings for review $i$ from $G^{polarity}$.\\
     $cos(ELMo(r^{polarity}_{i}), ELMo(r^{polarity}_{j}))$: the cosine similarity between\\ ELMo embeddings for reviews $i$ and $j$ from $G^{polarity}$.\\
     
    }
    \KwIn{Opinion groups, their lengths and their user ratings: $G^{polarity}$, $N^{polarity}$ and $V^{polarity}$.
    }
    \KwOut{Opinion groups' statistics: $SS^{polarity}$ and $SV^{polarity}$ 
    }
    \bigskip
    $polarity \leftarrow [strongly\;negative,\ weakly\;negative,\ neutral,\ weakly\;positive,\ strongly\;positive]$\\
    \bigskip
    \tcc{After applying the trained model on the collected movie and TV show reviews, we separate them into five opinion groups: strongly negative, weakly negative, neutral, weakly positive, and strongly positive. For each opinion group, we acquire the sum of their reviews' semantic similarity (cosine metric and ELMo embeddings) and the sum of their ratings}
    \bigskip
    \For{$i\ in\ polarity$}{
        $SS^{i} \leftarrow 0$\\
        $SV^{i} \leftarrow 0$\\
        \For{$j \leftarrow 1$ \KwTo $N^{i}$}{
            $SS^{i} \leftarrow SS^{i} + cos(ELMo(r^{i}_{1}), ELMo(r^{i}_{j}))$\\
            $SV^{i} \leftarrow SV^{i} + v_{j}^{i}$\\
        }
    }
    \caption{Opinion groups statistics acquisition}
\end{algorithm}

By applying algorithm 2, we retrieve for each group, the sum of their ratings and the sum of their semantic similarity. We propose formula (3.1) to compute a custom score for each opinion group.

\begin{equation}
    CS(G^{polarity}) = \frac{\frac{maxR \;\cdot\; SS^{polarity}}{N^{polarity}} + \frac{SV^{polarity}}{N^{polarity}}}{2}
\end{equation}

Formula (3.1) could also be written as follows:

\begin{equation}
    CS(G^{polarity}) = \frac{maxR \cdot SS^{polarity} + SV^{polarity}}{2 \cdot N^{polarity}}
\end{equation}

\bigskip
We denote: \\ \\
$maxR: $ The highest value of user ratings (5 or 10) depending on the range of ratings (1 to 5 or 1 to 10).\\
$SS^{polarity}: $ The sum of similarity for reviews contained in opinion group $G^{polarity}$.\\
$SV^{polarity}: $ The sum of user ratings in opinion group $G^{polarity}$.\\
$N^{polarity}: $ The number of reviews contained in opinion group $G^{polarity}$.\\

The custom score of each opinion group ranges between 1 and 5 or 1 and 10 depending on the range of user rating values. \par
Since the cosine metric returns values in the range of [0, 1], the average of the sum of semantic similarity for an opinion group is also between 0 and 1, therefore, we multiply the average of the sum of semantic similarity by 5 or 10 ($maxR$) to get a numerical value between 0 and 5 or 0 and 10, then, we add this value to the average of the sum of ratings and we divide them by 2.    

\subsection{Reputation generation}

We propose formula (3.3) (weighted arithmetic mean) to compute the movie or TV show reputation value.

\begin{equation}
    Rep(E) = \frac{\sum_{polarity} CS(G^{polarity}) \cdot N^{polarity}}{\sum_{polarity} N^{polarity}}
\end{equation}

\bigskip

$CS(G^{polarity}) $ is the custom score for opinion group $G^{polarity}$ computed by applying formula (3.1) or (3.2).\\

The movie or TV show reputation value has values in the range of [1, 5] or [1, 10] depending on the range of user ratings.

\section{Experimental evaluation}

\subsection{Dataset gathering}

We collect movie and TV show reviews and their numerical ratings from the IMDb web site using the web scraping tool ScrapeStorm. Figure 3.2 depicts the structure of IMDb user reviews.

\begin{figure}[!h]
\centering
  \includegraphics[width=0.8\textwidth]{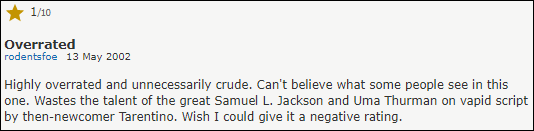}
  \caption{IMDb user reviews structure}
\end{figure}

The first ten datasets contain movie reviews and the remaining ten datasets contain TV show reviews. Table 3.2 shows the statistical information of the collected datasets.

\begin{table}[!h]
\centering
\begin{tabular}{l|l|l|l|}
\cline{2-4}
                                         & Movies & TV shows & Total \\ \hline
\multicolumn{1}{|l|}{Number of reviews}  & 1000   & 1000     & 2000  \\ \hline
\multicolumn{1}{|l|}{Number of entities} & 10     & 10       & 20    \\ \hline
\end{tabular}
\caption{Statistical information of the collected datasets}
\end{table}

After collecting the reviews, we replace the missing rating values with the average of the ratings, then, we lowercase them and remove punctuation marks and numbers.

\subsection{Training phase and fine-grained opinion mining}

We train the Multinomial Na\"ive Bayes model with the SST-5 dataset. The training set contains 1092 strongly negative reviews, 2218 weakly negative reviews, 1624 neutral reviews, 2322 weakly positive reviews, and 1288 strongly positive reviews. The test set contains 279 strongly negative reviews, 633 weakly negative reviews, 389 neutral reviews, 510 weakly positive reviews, and 399 strongly positive reviews. Figure 3.3 depicts the distribution of training and test samples over the five classes.  

\begin{figure}[!h]
\centering
  \includegraphics[width=1\textwidth]{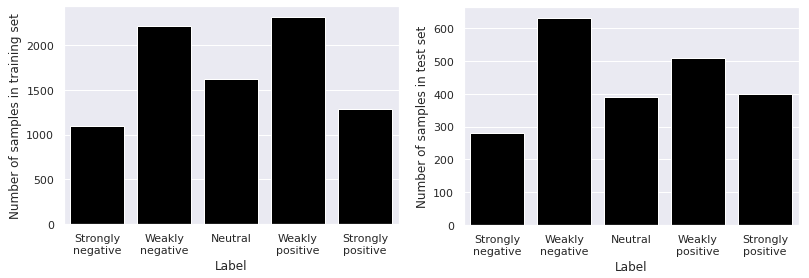}
  \caption{Number of samples in the SST-5 training and test set}
\end{figure}


Before feeding the data to the classifier for training, we preprocess them by removing punctuation marks, numbers, and whitespaces, then, we lowercase and lemmatize them. \par
After preprocessing the data, we must choose which classifier we will apply and which features we will use. Since deep learning models require substantial computing power (High-performance CPUs, GPUs, and RAM), we decided to work with one of the four models: Random Forest, Logistic Regression, Multinomial Na\"ive Bayes, and Linear Support Vector Machine (SVM). The last two classifiers (Na\"ive Bayes and SVM) have been recognized as the most popular supervised machine learning algorithms for polarity classification \cite{SHAWETAYLOR20113609}. For feature selection, we have tried many combinations: unigrams (count), bigrams (count), trigrams (count), TF-IDF unigrams, TF-IDF bigrams, and TF-IDF trigrams. We discarded some popular models such as word2vec and doc2vec because Wang et al. \cite{wang2017comparisons} have conducted experiments on Na\"ive Bayes, Logistic Regression, and Linear Support Vector Classifier (SVC) for short text classification using TF-IDF weighting, word2vec, and paragraph2vec (doc2vec), and they have reported that  TF-IDF/Counter features have the highest accuracy, while word2vec next, and doc2vec has the lowest accuracy. Table 3.3 summarizes the classification results of the four classifiers on the SST-5 dataset. 
    
\begin{table}[h]
\centering
\centerline{
\begin{adjustbox}{width=1\textwidth}
\begin{tabular}{l|c|c|c|c|c|c|c|}
\cline{2-8}
                                                                & \begin{tabular}[c]{@{}c@{}}Macro average\\  precision\end{tabular} & \begin{tabular}[c]{@{}c@{}}Macro average \\ recall\end{tabular} & \begin{tabular}[c]{@{}c@{}}Macro average \\ f1-score\end{tabular} & \begin{tabular}[c]{@{}c@{}}Weighted average \\ precision\end{tabular} & \begin{tabular}[c]{@{}c@{}}Weighted average \\ recall\end{tabular} & \begin{tabular}[c]{@{}c@{}}Weighted average \\ f1-score\end{tabular} & Accuracy      \\ \hline
\multicolumn{1}{|l|}{Random Forest (unigrams)}                  & 0.40                                                               & 0.31                                                            & 0.30                                                              & 0.40                                                                  & 0.36                                                               & 0.33                                                                 & 0.36          \\ \hline
\multicolumn{1}{|l|}{Random Forest (bigrams)}                   & 0.34                                                               & 0.29                                                            & 0.28                                                              & 0.34                                                                  & 0.32                                                               & 0.31                                                                 & 0.32          \\ \hline
\multicolumn{1}{|l|}{Random Forest (trigrams)}                  & 0.29                                                               & 0.23                                                            & 0.20                                                              & 0.31                                                                  & 0.23                                                               & 0.22                                                                 & 0.23          \\ \hline
\multicolumn{1}{|l|}{Random Forest (TF-IDF unigrams)}           & 0.40                                                               & 0.30                                                            & 0.28                                                              & 0.39                                                                  & 0.35                                                               & 0.31                                                                 & 0.35          \\ \hline
\multicolumn{1}{|l|}{Random Forest (TF-IDF bigrams)}            & 0.34                                                               & 0.29                                                            & 0.28                                                              & 0.34                                                                  & 0.32                                                               & 0.31                                                                 & 0.32          \\ \hline
\multicolumn{1}{|l|}{Random Forest (TF-IDF trigrams)}           & 0.28                                                               & 0.22                                                            & 0.20                                                              & 0.29                                                                  & 0.23                                                               & 0.21                                                                 & 0.23          \\ \hline
\multicolumn{1}{|l|}{Multinomial Naive Bayes (unigrams)}        & 0.43                                                               & \textbf{0.38}                                                   & \textbf{0.38}                                                     & 0.43                                                                  & \textbf{0.43}                                                      & \textbf{0.41}                                                        & \textbf{0.43} \\ \hline
\multicolumn{1}{|l|}{Multinomial Naive Bayes (bigrams)}         & 0.36                                                               & 0.30                                                            & 0.29                                                              & 0.36                                                                  & 0.35                                                               & 0.32                                                                 & 0.35          \\ \hline
\multicolumn{1}{|l|}{Multinomial Naive Bayes (trigrams)}        & 0.31                                                               & 0.26                                                            & 0.24                                                              & 0.31                                                                  & 0.29                                                               & 0.26                                                                 & 0.29          \\ \hline
\multicolumn{1}{|l|}{Multinomial Naive Bayes (TF-IDF unigrams)} & \textbf{0.48}                                                      & 0.34                                                            & 0.29                                                              & \textbf{0.46}                                                         & 0.41                                                               & 0.34                                                                 & 0.41          \\ \hline
\multicolumn{1}{|l|}{Multinomial Naive Bayes (TF-IDF bigrams)}  & 0.38                                                               & 0.29                                                            & 0.24                                                              & 0.38                                                                  & 0.35                                                               & 0.29                                                                 & 0.35          \\ \hline
\multicolumn{1}{|l|}{Multinomial Naive Bayes (TF-IDF trigrams)} & 0.29                                                               & 0.24                                                            & 0.19                                                              & 0.30                                                                  & 0.29                                                               & 0.23                                                                 & 0.29          \\ \hline
\multicolumn{1}{|l|}{Logistic Regression (unigrams)}            & 0.42                                                               & 0.37                                                            & 0.37                                                              & 0.42                                                                  & 0.41                                                               & 0.39                                                                 & 0.41          \\ \hline
\multicolumn{1}{|l|}{Logistic Regression (bigrams)}             & 0.38                                                               & 0.28                                                            & 0.23                                                              & 0.37                                                                  & 0.34                                                               & 0.27                                                                 & 0.34          \\ \hline
\multicolumn{1}{|l|}{Logistic Regression (trigrams)}            & 0.36                                                               & 0.23                                                            & 0.18                                                              & 0.35                                                                  & 0.28                                                               & 0.22                                                                 & 0.28          \\ \hline
\multicolumn{1}{|l|}{Logistic Regression (TF-IDF unigrams)}     & 0.42                                                               & 0.35                                                            & 0.34                                                              & 0.41                                                                  & 0.40                                                               & 0.37                                                                 & 0.40          \\ \hline
\multicolumn{1}{|l|}{Logistic Regression (TF-IDF bigrams)}      & 0.43                                                               & 0.28                                                            & 0.23                                                              & 0.41                                                                  & 0.35                                                               & 0.27                                                                 & 0.35          \\ \hline
\multicolumn{1}{|l|}{Logistic Regression (TF-IDF trigrams)}     & 0.30                                                               & 0.23                                                            & 0.17                                                              & 0.32                                                                  & 0.29                                                               & 0.21                                                                 & 0.29          \\ \hline
\multicolumn{1}{|l|}{Linear SVM (unigrams)}                     & 0.38                                                               & 0.37                                                            & 0.37                                                              & 0.39                                                                  & 0.40                                                               & 0.39                                                                 & 0.40          \\ \hline
\multicolumn{1}{|l|}{Linear SVM (bigrams)}                      & 0.33                                                               & 0.31                                                            & 0.31                                                              & 0.34                                                                  & 0.34                                                               & 0.33                                                                 & 0.34          \\ \hline
\multicolumn{1}{|l|}{Linear SVM (trigrams)}                     & 0.31                                                               & 0.25                                                            & 0.22                                                              & 0.32                                                                  & 0.29                                                               & 0.25                                                                 & 0.29          \\ \hline
\multicolumn{1}{|l|}{Linear SVM (TF-IDF unigrams)}              & 0.38                                                               & 0.38                                                            & 0.38                                                              & 0.39                                                                  & 0.41                                                               & 0.39                                                                 & 0.41          \\ \hline
\multicolumn{1}{|l|}{Linear SVM (TF-IDF bigrams)}               & 0.33                                                               & 0.31                                                            & 0.31                                                              & 0.34                                                                  & 0.34                                                               & 0.33                                                                 & 0.34          \\ \hline
\multicolumn{1}{|l|}{Linear SVM (TF-IDF trigrams)}              & 0.31                                                               & 0.27                                                            & 0.25                                                              & 0.31                                                                  & 0.30                                                               & 0.27                                                                 & 0.30          \\ \hline
\end{tabular}
\end{adjustbox}
}
\caption{Sentiment analysis classification results}
\end{table}

From table 3.3, we can see that the Multinomial Na\"ive Bayes classifier achieves the best classification results when it's trained with unigrams (count). Logistic Regression and linear SVM classifiers also gave good results when they are trained with unigrams (count) or TF-IDF unigrams. The worst results are provided by Random Forest since it achieves a 0.36 accuracy at its best. Figure 3.4 depicts the confusion matrix of the Multinomial Naive Bayes (unigrams) for the SST-5 test set.\\

\begin{figure}[!h]
\centering
  \includegraphics{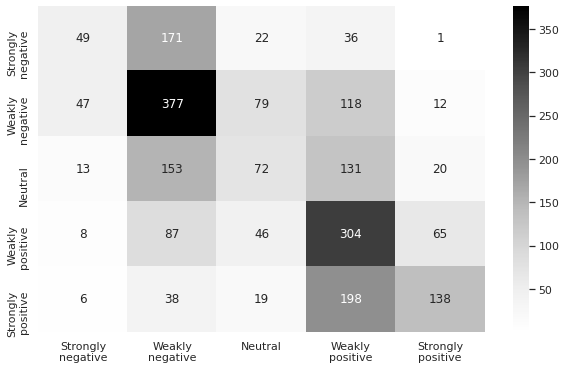}
  \caption{Confusion matrix of the Multinomial Naive Bayes (unigrams) for the SST-5 test set}
\end{figure}

We mention that $BERT_{base}$ achieves a 0.45 accuracy and 0.40 macro average f1-score, GRU-RNN-WORD2VEC achieves a 0.45 accuracy and Recursive Neural Tensor Network achieves a 0.46 accuracy. Besides, deep learning algorithms take a long time to train (Table 3.4) due to the large number of parameters. Based on that, we have chosen to apply the Multinomial Na\"ive Bayes classifier since it achieves an accuracy of 0.43 and it doesn't require substantial computing power to be trained. Table 3.4 depicts the training time of Bidirectional Gated Recurrent Unit (Bi-GRU), Bidirectional Long Short-Term Memory (Bi-LSTM), Recurrent Neural Network (RNN), and Multinomial Na\"ive Bayes (MNB) on the SST-5 dataset.\\

\begin{table}[htb!]
\centering
\caption{Training time of Bidirectional Gated Recurrent Unit (Bi-GRU), Bidirectional Long Short-Term Memory (Bi-LSTM), Recurrent Neural Network (RNN) and Multinomial Na\"ive Bayes (MNB) on the SST-5 dataset}
\begin{tabular}{l c c c}
\hline
Model & Epochs & Batch size & Training time (seconds) \\ \hline
Bi-GRU & 50 & 64 & 210.10 \\ 
Bi-LSTM & 50 & 64 & 180.25 \\
RNN & 50 & 64 & 85.26 \\ 
MNB & -- & -- & \textbf{3.77} \\ \hline
\end{tabular}
\end{table}

One of the benefits of fine-grained opinion mining is that it provides a better understanding of the distribution of reviews over the five emotion classes, therefore, visualizing these five classes will help users and customers make up their minds about the target item (buying, renting).

\subsection{Reputation evaluation}

MTVRep offers a comprehensive reputation visualization form (Figure 3.5) by depicting the numerical reputation value and the distribution of reviews over the five emotion classes, Table 3.5 shows comparison results between MTVRep and previous studies in terms of reputation visualization.

\begin{table}[!htb]
\caption{Comparison results: reputation visualization}
\centerline{
    \scalebox{1}{
    \begin{tabular}{l c c } 
	\hline

	Work & Distribution of reviews polarity & Numerical reputation value \\

	\hline

	Yan et al. (2017) \cite{yan2017fusing} & \xmark & \cmark  \\

	Benlahbib \& Nfaoui (2019) \cite{benlahbib2019unsupervised} & \xmark & \cmark  \\

    Benlahbib et al. (2019) \cite{9068916} & \xmark & \cmark   \\

	Benlahbib \& Nfaoui (2020) \cite{benlahbib2019hybrid} & \cmark & \cmark  \\

	MTVRep & \cmark & \cmark \\ \hline 
    \end{tabular}
            }
            }
\end{table}

\begin{figure}[!htb]
\centering
  \includegraphics[width=1\textwidth]{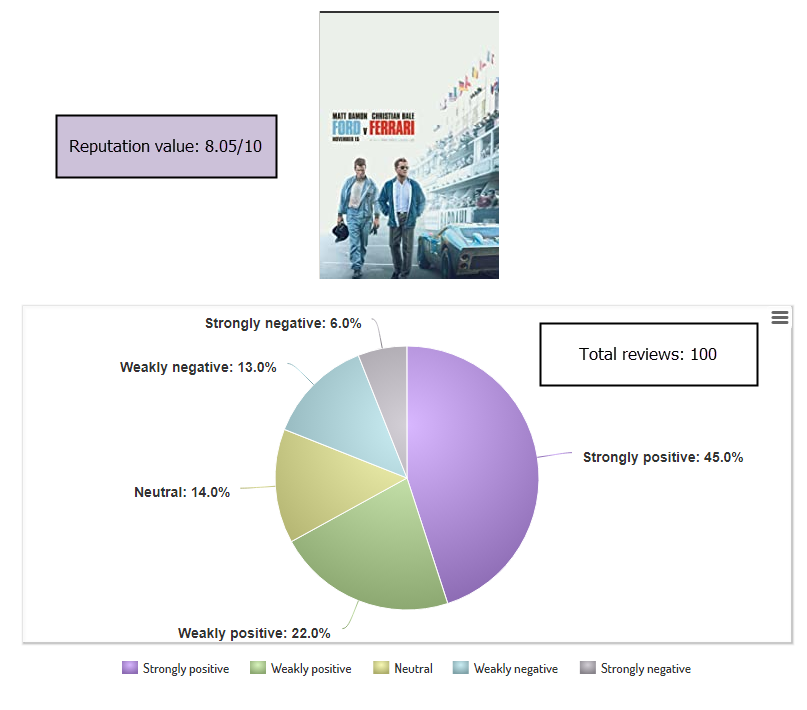}
  \caption{Reputation visualization}
\end{figure}

An important issue that was neglected in the past research on reputation generation is identifying the sentiment strength during the phase of opinion mining. Actually, existing studies have only focused on classifying reviews as positive or negative, disregarding sentiment intensity. Therefore, we propose MTVRep, a movie and TV show reputation system that combines fine-grained sentiment analysis and semantic analysis for the purpose of generating and visualizing reputation toward movies and TV shows. Table 3.6 depicts the features exploited by previous studies and MTVRep during reputation generation and visualization.

\begin{table}[!htb]
\caption{Comparison results: exploited features}
\centerline{
    \scalebox{1}{
    \begin{tabular}{l c c c} 
	\hline

	Work & Semantic & Sentiment (binary) & Sentiment (fine-grained) \\

	\hline

	Yan et al. (2017) \cite{yan2017fusing} & \cmark & \xmark & \xmark \\

	Benlahbib \& Nfaoui (2019) \cite{benlahbib2019unsupervised} & \cmark & \xmark & \xmark \\

     Benlahbib et al. (2019) \cite{9068916} & \cmark & \cmark & \xmark  \\

	Benlahbib \& Nfaoui (2020) \cite{benlahbib2019hybrid} & \cmark & \cmark & \xmark \\

	MTVRep & \cmark & \xmark & \cmark \\ \hline 
    \end{tabular}
            }
            }
\end{table}

In order to evaluate the performance of MTVRep in generating accurate reputation values toward various movies and TV shows, we compared it to Yan et al. \cite{yan2017fusing} reputation system. We set the opinion fusion threshold $t_{0}$ to 0.15 since the authors mentioned that their reputation system performs in its best when $t_{0}=0.15$. We applied the two reputation systems on the twenty collected datasets. The chosen evaluation measure is the squared error between the movie or TV show IMDb Weighted Average Ratings and the numerical reputation value computed by one of the two reputation systems. \par
The formula of the squared error is: $SE = (x_i - y_i)^2$ where $x_i$ is the reputation value returned by one of the two systems and $y_i$ is the IMDb Weighted Average Ratings toward the target movie or TV show. Figure 3.6 depicts the IMDb Weighted Average Ratings for the \textbf{Forrest Gump} movie.\\

\begin{figure}[!htb]
\centering
  \includegraphics[width=1\textwidth]{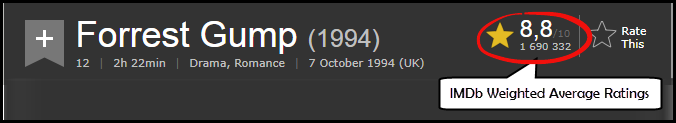}
  \caption{IMDb Weighted Average Ratings}
\end{figure}

According to IMDb\footnote{\url{https://help.imdb.com/article/imdb/track-movies-tv/weighted-average-ratings/GWT2DSBYVT2F25SK?ref_=helpsect_pro_2_8\#}}: \textit{"IMDb publishes weighted vote averages rather than raw data averages. Various filters are applied to the raw data in order to eliminate and reduce attempts at vote stuffing by people more interested in changing the current rating of a movie than giving their true opinion of it. The exact methods we use will not be disclosed. This should ensure that the policy remains effective. The result is a more accurate vote average"}. The motivation behind choosing the squared error instead of absolute error resides in the fact that reputation systems don't tolerate high error values. Consequently, the squared error will penalize large errors more. Figures 3.7 and 3.8 show the comparison results between the two reputation systems over the twenty datasets.

\begin{figure}[!htb]
\centering
  \includegraphics[width=1\textwidth]{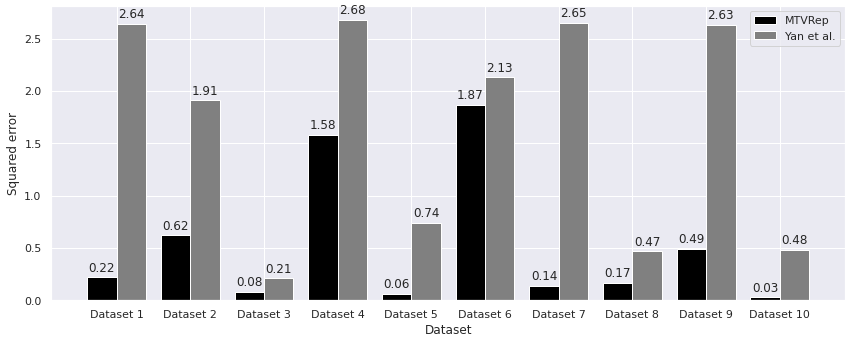}
  \caption{Squared error comparison results: dataset 1 to dataset 10}
\end{figure}

As illustrated in Figure 3.7, MTVRep produces the nearest reputation value to IMDb Weighted Average Ratings for the first ten datasets that contain movie reviews compared to Yan et al. reputation system \cite{yan2017fusing}. We observe that the squared error of reputation system \cite{yan2017fusing} exceeds 2.5 in dataset 1, dataset 4, dataset 7, and dataset 9. We also observe that the squared error of MTVRep doesn't surpass 0.1 in dataset 3, dataset 5, and dataset 10, which implies that the system generates accurate reputation values toward movies since the highest squared error achieved by MTVRep is 1.87 (dataset 6).

\begin{figure}[!htb]
\centering
  \includegraphics[width=1\textwidth]{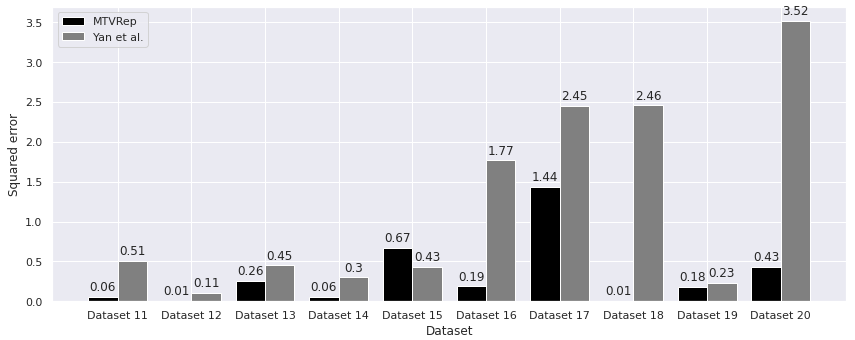}
  \caption{Squared error comparison results: dataset 11 to dataset 20}
\end{figure}

Figure 3.8 shows that except for dataset 15, MTVRep outperforms Yan et al. \cite{yan2017fusing} reputation system on all the remaining nine datasets that contain TV show reviews. We also observe that the squared error of reputation system \cite{yan2017fusing} exceeds 3.5 in dataset 20, on the other hand, MTVRep doesn't exceed 1.44 in its worst. \par
We conclude that the proposed reputation system MTVRep performs well in generating and visualizing reputation for movies and TV shows since it produces the nearest reputation value to IMDb Weighted Average Ratings for both movies and TV shows compared to Yan et al. \cite{yan2017fusing} reputation system.

\section{Conclusion, summary and future direction}

In this chapter, we have proposed MTVRep, a system that combines fine-grained opinion mining and semantic analysis for the purpose of generating and visualizing reputation toward movies and TV shows. The web scraping tool ScrapeStorm was used to collect 2000 movie and TV show reviews and their numerical ratings from IMDb, and the Multinomial Na\"ive Bayes classifier was trained on the SST-5 dataset to perform fine-grained opinion mining task. Experimental studies showed that MTVRep outperforms Yan et al. reputation system \cite{yan2017fusing} since it produces the nearest reputation values to the ground truth (IMDb Weighted Average Ratings) for both movies and TV shows. We believe that MTVRep could be integrated into any platform where users share their reviews and ratings freely toward movies and TV shows. \par
Future works will focus on (1) using more sophisticated models for opinion mining such as BERT and XLNet, (2) exploiting further features some of which are user credibility, review time, and review helpfulness, and (3) incorporating aspect-based opinion mining to enhance the reputation visualization form by showing more useful information toward the target movie or TV show (aspects). \par
The next chapter will present a reputation system that incorporates four review attributes: review helpfulness, review time, review sentiment polarity, and review rating in order to generate reputation toward various online items (products, movies, TV shows, hotels, restaurants, services). The system also provides a comprehensive reputation visualization form to help potential customers make an informed decision by depicting the numerical reputation value, opinion group categories, top-k positive reviews, and top-k negative reviews.



\chapter{Aggregating Customer Review Attributes for Online Reputation Generation} 
\section{Introduction}
The exponential growth of Web 2.0 has dramatically impacted the evolution of e-commerce platforms \cite{hou2019mining} \cite {10.1016/j.ipm.2016.12.002} \cite{bafna2013feature} \cite{Zhuang:2006:MRM:1183614.1183625}. Recent online shopping statistics showed that the number of users of some famous e-commerce websites such as Jingdong\footnote{\url{https://www.jd.com}}, Alibaba\footnote{\url{https://www.alibaba.com/}} and Amazon\footnote{\url{https://www.amazon.com}} has exceeded 1 billion \cite{yan2017fusing}. Thereby, customer reviews attached to a product can easily surpass thousands \cite{10.1016/j.ipm.2016.12.002, doi:10.1080/00207721.2015.1116640, Hu:2004:MSC:1014052.1014073}. In fact, while, a good number of reviews could indeed give a hint about the quality of an item, a potential customer may not have time or effort to read all reviews for the purpose of making a decision \cite{pecar-2018-towards}. Thus, the need for the right tools and technologies to help in such a task becomes a necessity for the buyer as for the seller.\par
Currently, little work has been performed to support customer decision making in E-commerce using natural language processing techniques. We identify principally two techniques. The first one is feature-based summarization that aims to identify the target entity (product, movie, hotel, restaurant, service) features and its corresponding opinions polarity (positive/negative), then, a feature-based summary of the reviews is generated \cite{Hu:2004:MSC:1014052.1014073, bafna2013feature, Zhuang:2006:MRM:1183614.1183625}. While the second technique is called reputation generation, whose main focus is to produce an estimation value in which an entity is held based on mining customer reviews expressed in natural languages \cite{yan2017fusing, benlahbib2019unsupervised, benlahbib2019hybrid, 9068916}.\par
Previous studies on reputation generation have primarily focused on using semantic and sentiment analysis \cite{yan2017fusing, benlahbib2019unsupervised, benlahbib2019hybrid, 9068916}, disregarding other useful information that could be extracted from user reviews, such as review helpfulness, which implies that reviews that receive higher votes from other users typically provide more valuable information, and review time, which implies that more recent reviews generally provide users with more up-to-date information.\par
An accurate and reliable reputation system should consider exploiting more online reviews' features such as review attached rating, review helpfulness, review time, and review sentiment orientation. For that reason, we propose a reputation system that incorporates all these attributes during the process of generating and visualizing reputation for various entities (movies, products, hotels, restaurants, and services). In this manner, this study addressed the following research question: with the consideration of review helpfulness, review time, review sentiment orientation probability and review attached rating, can the proposed reputation system offer better results in terms of reputation generation and visualization than the previous reputation systems (consider only semantic and sentiment relations)? \par
The contributions of this chapter are summarized as follows:
\begin{itemize}
	\item Firstly, we propose a novel system that incorporates review time, review helpfulness, review sentiment orientation and review attached rating for the purpose of generating a numerical reputation value toward various entities (movies, products, hotels, restaurants, services, etc). 
	\item Secondly, we propose a new comprehensive form to visualize reputation by showing numerical reputation value, opinion categories, top positive review, and top negative review in order to support customer decision making process in E-commerce (buying, renting, booking).
\end{itemize}

\section{Problem definition}

This section covers the necessary background for understanding the remainder of this chapter, including the problem definition. \par
In this chapter, we face the problem of generating reputation for movies, products, hotels, restaurants and services by aggregating review time, review helpfulness votes, review sentiment orientation and review attached rating. Given a set of reviews $R_j=\{r_{1j},\ r_{2j},\ \dots,\ r_{nj}\}$ expressed for an entity $E_j$, the set of their attached ratings $V_j=\{v_{1j},\ v_{2j},\ \dots,\ v_{nj}\}$ where $v_{ij} \in [1,5]$ or $v_{ij} \in [1,10]$ depending on the rating system, the set of their attached helpfulness votes $RH_j=\{rh_{1j},\ rh_{2j},\ \dots,\ rh_{nj}\}$ where $rh_{ij} \in \mathbb{N^*}$, the set of their posting time $RT_j=\{rt_{1j},\ rt_{2j},\ \dots,\ rt_{nj}\}$ and the set of their sentiment orientation probabilities predicted by fine-tuned $BERT_{base}$ model $BERT_{j}=\{bert(r_{1j}),\ bert(r_{2j}),\ \dots,\ bert(r_{nj})\}$ where $bert(r_{ij}) \in [0,1]^2$. The goal is to compute a review score for each review $RS_j=\{rs_{1j},\ rs_{2j},\ \dots,\ rs_{nj}\}$ based on its helpfulness votes, its posting time and its sentiment orientation, and finally, to compute a reputation value $Rep$ for an entity $j$ by averaging the product of reviews score and reviews attached rating. Table 4.1 presents the descriptions of notations used in the rest of this chapter. \par

\begin{table}[htb]
\caption{Symbol denotation}
\centerline{
\begin{tabular}{| l | p{12cm} |}
\hline
Symbol & Description                                                              \\ \hline
$R_j$      & The set of reviews expressed for the entity $j$                                                                  \\
$V_j$      & The set of ratings expressed for the entity $j$                                                                  \\
$RH_j$     & The set of reviews helpfulness votes expressed for the entity $j$                                               \\
$RT_j$     & The set of reviews posting time expressed for the entity $j$                                                     \\
$BERT_j$   & The set of sentiment orientation probabilities predicted by fine\_tuned $BERT_{Base}$ for reviews expressed for the entity $j$ \\
$RS_j$     & The set of reviews score expressed for the entity $j$                                                            \\
$E_j$      & The target entity $j$                                                                 \\
$O_j$      & The total number of reviews expressed for the target entity $j$                       \\
$Rep$    & The reputation value  \\ \hline                                                             
\end{tabular}
}
\end{table}

\section{Proposed approach}

\subsection{System overview}

Our approach consists mainly on four steps:

\begin{itemize}
    \item Firstly, we collect real data from websites that specialize in gathering customer reviews such as IMDb\footnote{\url{https://www.imdb.com/}}, TripAdvisor\footnote{\url{https://www.tripadvisor.com/}}, and Amazon\footnote{\url{https://www.amazon.com}} using web scraping tools, then, we preprocess them.
    \item  Secondly, we assign three numerical scores to each review: helpfulness score, time score, and sentiment orientation score.
    \item Thirdly, we compute a review score based on the pre-computed scores (helpfulness score, time score, and sentiment orientation score).
    \item Finally, we generate a numerical reputation value toward the target entity (product, movie, hotel, restaurant, service, etc). Then, we propose a new form to visualize reputation by depicting numerical reputation value, opinion categories, positive review with the highest score, and negative review with the highest score.
\end{itemize}

Figure 4.1 describes the pipeline of our work.

\begin{figure}[!h]
\centering
  \includegraphics[width=1\textwidth]{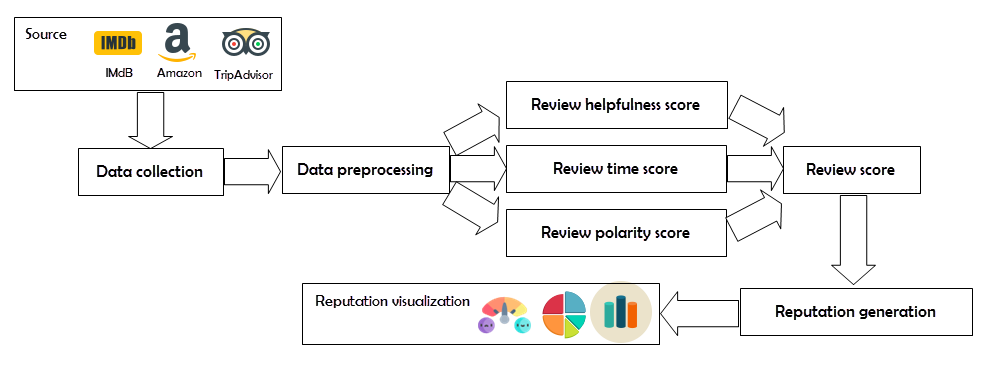}
  \caption{Proposed system pipeline}
\end{figure}

\subsection{Data collection and preprocessing}

Differently from previous studies on reputation generation, which mainly focus on extracting semantic and sentiment relations of reviews, our work incorporates other factors such as review helpfulness and review time. Hopefully, the majority of popular E-commerce websites such as TripAdvisor\footnote{\url{https://www.tripadvisor.com/}} and Amazon\footnote{\url{https://www.amazon.com}} gather online reviews with respect to the following structure: \textbf{textual review, review helpfulness votes, and review posting time}. Figure 4.2 describes the structure of the online reviews. \par

\begin{figure}[!h]
\centering
  \includegraphics[scale=0.9]{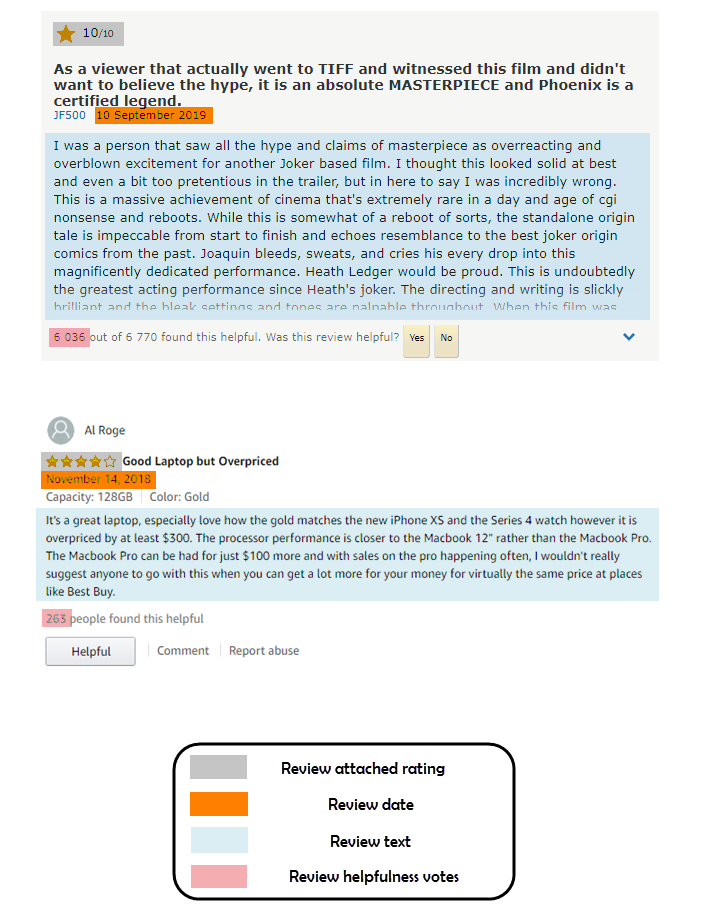}
  \caption{IMDb and Amazon reviews structure.}
\end{figure}

With the use of a web scraping tool, we have been able to collect raw data from some real data suppliers like Amazon, TripAdvisor, and IMDb.\par
After collecting all reviews, we applied some preprocessing techniques (lowercasing, tokenization, \dots). Technical details of the data collection and preprocessing phase are described in section 4.4  Experiment results. subsection 4.4.1. Data collection and preprocessing. \par

\subsection{Review helpfulness}

The number of helpfulness votes attached to a review indicates how informative it is, which implies that reviews that receive higher votes from other users typically provide more valuable information. Thus, we design formula (4.1) to compute review helpfulness score.
\begin{equation}
    H(r_{ij}) = 
        \begin{cases}
           \text{0.75} &\quad\text{if  } rh_{ij}=0 \text{  or  } \frac{\log_{10}(rh_{ij})}{\log_{10}(N_{j})} \leqslant 0.75\\
           \log_{N_j}(rh_{ij})\text{} &\quad\text{Otherwise}   \\
         \end{cases}
\end{equation}
\bigskip
We denote: \\ \\
$r_{ij}: $ Review number $i$ expressed for the entity $j$.\\
$H(r_{ij}): $ Helpfulness score of review $r_{ij}$.\\
$rh_{ij}: $ The number of helpfulness votes attached to review $r_{ij}$.\\
$N_{j}: $ The number of helpfulness votes attached to the most voted review toward the entity $j$.\\

The helpfulness score for a review ranges between 0.75 and 1 because we don't want to assign a low score to reviews with a small number of helpfulness votes. \par
We mention that $\frac{\log_{10}(rh_{ij})}{\log_{10}(N_{j})} \leqslant 0.75$ means that $\log_{N_j}(rh_{ij}) \leqslant 0.75$ due to the fact that:\\
$
    \log_N(n) = \frac{\log_{base}(n)}{\log_{base}(N)} = \frac{\log_{10}(n)}{\log_{10}(N)} 
$

\bigskip
By applying equation (4.1), the most voted review will receive a review helpfulness score of 1 since $\log_{N_j}(N_j) = 1$. Also, reviews with high helpfulness votes will receive a high review helpfulness score and reviews with low helpfulness votes will receive a low review helpfulness score since for $x \in [1, N]$ and $y \in [1, N]$: $x \leq y$ implies that $log_N(x) \leq log_N(y)$. Algorithm 3 computes the helpfulness score for review $r_{ij}$.

\begin{algorithm}[h]
\SetKwInOut{KwIDe}{Define}
\SetAlgoLined
\DontPrintSemicolon
\KwIDe{
     $RH_j=\{rh_{1j},\ rh_{2j},\ \dots,\ rh_{nj}\}$: The set of reviews' helpfulness votes expressed for the entity $j$.}
\KwIn{$RH_{j}$}    

    \SetKwFunction{FMain}{H}
    \SetKwProg{Fn}{Function}{:}{}
    \Fn{\FMain{$rh_{ij}$}}{
        
        \eIf{$ rh_{ij}=0 \text{  or  } \frac{\log_{10}(rh_{ij})}{\log_{10}(max(RH_j))} \leqslant 0.75 $}
            {$ H \longleftarrow 0.75$}
            {$ H \longleftarrow \log_{max(RH_j)}(rh_{ij})$}

        \textbf{return} $H$ 
}
\textbf{End Function}
\caption{Review helpfulness score}
\label{HelpfulnessScoreAlgo}
\end{algorithm}

\subsection{Review Time}

Could you tell what would happen if we take a very well-reviewed gaming laptop from 10 years ago and put it on an online store?  To answer this question, let us travel back in time to 20 years ago, where the gaming industry witnessed a great competition between gaming consoles, and where the enjoyment of a hardcore gaming experience was limited to that kind of tech. In that era, a gamer had to have a fat and heavy TV, with cables attached to a relatively big dedicated gaming console and its controllers in order to play a video game. All this bunch of materials and cables remain in one place in the house. Next, the industry shifted to computers, and then to mobile computers also known as laptops, which brought enough satisfaction to all the gaming consumers over the world. Although, laptops have been made much heavier than they should be, yet, it was very exciting to have the ability to enjoy your favorite games wherever you want just by packing your laptop on a backpack rather than having to be stuck in a room to play. Spatial freedom was a gift for the republic of players and so going mobile was their prior preference at that time. \par
\textbf{Time goes forward, and so, the consumer preferences and choices}. By today's standards, just going mobile is not good enough, gamers want lighter laptops, more performance, high-end graphic cards, high resolution/fps screens, mechanical keyboard, and the list is long \dots \\ Today, gamers all over the planet become more demanding, their preferences changed drastically and so the industry does while trying to keep up with the human desires. \par
Back to our question, it is obvious that nobody will care about a 20-year-old gaming laptop, even if it was a best seller with 1 million 5-star reviews at that time. Why is that?  simply because it becomes \textbf{obsolete} by the \textbf{modern user criteria}. Its 1 million review doesn't matter anymore. And so, \textbf{as all the things and beings}, \textbf{reviews} also have \textbf{an expiration date}. where they become irrelevant to the buyer.\par
Although, a product, laptop, movie or hotel may had very good reviews once, but time took off their power, their importance, and their effect over the judgment and decision of the consumer. At the end, \textbf{"you cannot beat time"}. \par
To conclude, we believe that more recent reviews generally provide users with more up-to-date information. Therefore, we design formula (4.2) to assign a time score to each review. 
\begin{equation}
    T(r_{ij}) = 
        \begin{cases}
           \text{0.8} &\quad\text{if  } y - rt_{ij} \geqslant 100\\
           1-(y-rt_{ij}) \times 0.002 &\quad\text{ Otherwise} \\ 
         \end{cases}
\end{equation}
\bigskip
We denote: \\ \\
$T(r_{ij}): $ Time score of review $r_{ij}$.\\
$rt_{ij}: $ Publication year of review $r_{ij}$.\\
$y: $ Current year.\\

The time score for a review ranges between 0.8 and 1, which implies that a higher time score is assigned to the most recent reviews.\\
Algorithm 4 computes the time score for review $r_{ij}$.

\begin{algorithm}[h]
\SetKwInOut{KwIDe}{Define}
\SetAlgoLined
\DontPrintSemicolon
\KwIDe{
     $RT_j=\{rt_{1j},\ rt_{2j},\ \dots,\ rt_{nj}\}$: The set of reviews' posting time expressed for the entity $j$.}
\KwIn{$RT_j$}    

    \SetKwFunction{FMain}{T}
    \SetKwProg{Fn}{Function}{:}{}
    \Fn{\FMain{$rt_{ij}$}}{
        
        \eIf{$ currentYear - rt_{ij} \geqslant 100 $}
            {$ T \longleftarrow 0.8$}
            {$ T \longleftarrow 1-(currentYear-rt_{ij}) \times 0.002 $}

        \textbf{return} $T$ 
}
\textbf{End Function}
\caption{Review time score}
\label{TimeScoreAlgo}
\end{algorithm}

With the help of a film critic, we have been able to determine suitable minimum values for each of the review helpfulness and review time scores. Indeed, we have performed multiple experiments on a various range of movies, by trying different minimum values for both scores as parameters, where in each one among these, we compare the generated reputation value to the film critic's own rating regarding a given movie. Which leads to 0.75 and 0.8 to be chosen successively as the fittest minimum values for review helpfulness and review time scores. Next, given the high accuracy achieved through our reputation system, the same last experiments have been done on other domains such as products, restaurants, and services, where we noticed very good results, particularly when using 0.75 and 0.8 as the minimum values for each of the scores.

\subsection{Review sentiment orientation}

We fine-tuned $BERT$ model to determine the sentiment orientation probability of a target review due to the fact that it has achieved state-of-the-art results in a wide variety of natural language processing tasks by learning contextual relations between words or sub-words in a text. In this paper, we have an interest in assigning a sentiment orientation score to each review. Since fine-tuned BERT returns an array of 2 values: the probability of being negative and the probability of being positive (Softmax activation function), we apply the $max$ function to the fine-tuned BERT output vector. The highest probability is kept as the sentiment orientation score of the target review.  
\begin{equation}
    S(r_{ij}) = max([P^{negative}_{r_{ij}},P^{positive}_{r_{ij}}])
\end{equation}
\bigskip
We denote: \\ \\
$S(r_{ij}): $ Sentiment orientation score for review $r_{ij}$.\\
$P^{negative}_{r_{ij}}: $ BERT model output prediction for review $r_{ij}$ being negative.\\
$P^{positive}_{r_{ij}}: $ BERT model output prediction for review $r_{ij}$ being positive.\\

The sentiment polarity of a target review $r_{ij}$ is predicted as negative if $P^{negative}_{r_{ij}} > P^{positive}_{r_{ij}}$ and predicted as positive if $P^{negative}_{r_{ij}} < P^{positive}_{r_{ij}}$.\\

\bigskip

Algorithm 5 computes the sentiment orientation score for review $r_{ij}$.

\begin{algorithm}[h]
\SetKwInOut{KwIDe}{Define}
\SetAlgoLined
\DontPrintSemicolon
\KwIDe{
     $R_j=\{r_{1j},\ r_{2j},\ \dots,\ r_{nj}\}$: The set of reviews expressed for the entity $j$.\\
	 $BERT_{j}=\{bert(r_{1j}),\ bert(r_{2j}),\ \dots,\ bert(r_{nj})\}$: The set of output vectors\\ of fine-tuned $BERT_{Base}$ (the sentiment orientation probability of reviews\\ expressed for the entity $j$.}
\KwIn{$R_j$}    

    \SetKwFunction{FMain}{S}
    \SetKwProg{Fn}{Function}{:}{}
    \Fn{\FMain{$r_{ij}$}}{

        $ S \longleftarrow max(bert(r_{ij}))$

        \textbf{return} $S$ 
}
\textbf{End Function}
\caption{Review sentiment orientation score}
\label{SentimentOrientationscoreAlgo}
\end{algorithm}

\subsection{Review score}

Based on the above scores, we design formula (4.4) to compute a numerical score for each review:
\begin{equation}
    RS(r_{ij}) = \frac{H(r_{ij}) + T(r_{ij}) + S(r_{ij})}{3}
\end{equation}
\bigskip
We denote: \\ \\
$RS(r_{ij}): $ Review score for review $r_{ij}$.\\
$H(n_{ij}): $ Helpfulness score of review $r_{ij}$.\\
$T(r_{ij}): $ Time score of review $r_{ij}$.\\
$S(r_{ij}): $ Sentiment orientation score for review $r_{ij}$.\\

Since review helpfulness score, review time score, and review sentiment orientation score range between 0 and 1, the generated review score is also between 0 and 1. Algorithm 6 computes the review score for all reviews.

\begin{algorithm}[!htb]
    \SetKwInOut{KwIDe}{Define}
    \SetKwInOut{KwIn}{Input}
    \SetKwInOut{KwOut}{Output}
    
    \KwIDe{
     $R_j=\{r_{1j},\ r_{2j},\ \dots,\ r_{nj}\}$: The set of reviews expressed for the entity $j$.\\
     $RH_j=\{rh_{1j},\ rh_{2j},\ \dots,\ rh_{nj}\}$: The set of reviews' helpfulness votes\\ expressed for the entity $j$.\\
	 $RT_j=\{rt_{1j},\ rt_{2j},\ \dots,\ rt_{nj}\}$: The set of reviews' posting time expressed\\ for the entity $j$.\\
	 $RS_j=\{rs_{1j},\ rs_{2j},\ \dots,\ rs_{nj}\}$: The set of reviews' score expressed for the\\ entity $j$.\\

    }
    \KwIn{$R_j$, $RH_j$, $RT_j$
    }
    \KwOut{$RS_j$}
    \For{$i\ in\ range(n)$}{
        $rs_{ij} \longleftarrow (H(rh_{ij}) + T(rt_{ij}) + S(r_{ij})) / 3$\\
        
    }
\caption{Review score}
\label{ReviewScoreAlgo}
\end{algorithm}

Table 4.2 represents example results of review score.

\begin{table}[!htb]
\caption{Example results of review score}
\centerline{
    \scalebox{0.7}{
\begin{tabular}{|l|c|c|c|c|c|}
\hline
Review   & \begin{tabular}[c]{@{}c@{}}Review \\ helpfulness score\end{tabular} & \begin{tabular}[c]{@{}c@{}}Review\\  time score\end{tabular} & \begin{tabular}[c]{@{}c@{}}Review \\ sentiment orientation\end{tabular} & \begin{tabular}[c]{@{}c@{}}Review\\  sentiment score\end{tabular} & \begin{tabular}[c]{@{}c@{}}Review \\ score\end{tabular} \\ \hline
Review 1 & 1                                                                   & 0.968                                                        & Positive                                                                & 0.99732805                                                        & 0.98844268                                              \\ \hline
Review 2 & 0.75                                                                & 0.982                                                        & Positive                                                                & 0.99679191                                                        & 0.9095973                                               \\ \hline
Review 3 & 0.87468842                                                          & 0.974                                                        & Negative                                                                & 0.99608659                                                        & 0.94825834                                              \\ \hline
Review 4 & 0.91100877                                                          & 0.964                                                        & Positive                                                                & 0.9970323                                                         & 0.95734702                                              \\ \hline
Review 5 & 0.77448754                                                          & 0.96                                                         & Negative                                                                & 0.99694509                                                        & 0.91047754                                              \\ \hline
\end{tabular}
}
}
\end{table}

\subsection{Reputation generation}

We propose formula (4.5) to compute a single reputation value toward the target entity using review score $RS(r_{ij})$ and review attached rating $v_{ij}$:
\begin{equation}
    Rep(E_{j}) = \frac{\sum_{i=1}^{O_{j}} RS(r_{ij}) . v_{ij}}{O_{j}}
\end{equation}
\bigskip
We denote: \\ \\
$E_{j}: $ Target entity $j$.\\
$Rep(E_{j}): $ Reputation value toward the target entity $j$.\\
$RS(r_{ij}): $ Review score for review $r_{ij}$.\\
$v_{ij}: $ Attached numerical rating to review $r_{ij}$.\\
$O_{j}: $ Total number of reviews expressed for the target entity $j$.\\

The reputation value varies from 1 to 5 or 1 to 10 depending on the range of rating values. Algorithm 7 computes the reputation value toward a target item.

\begin{algorithm}[ht]
    \SetKwInOut{KwIDe}{Define}
    \SetKwInOut{KwIn}{Input}
    \SetKwInOut{KwOut}{Output}
    \SetKwInOut{KwBeg}{Begin}
    
    \KwIDe{
     $V_j=\{v_{1j},\ v_{2j},\ \dots,\ v_{nj}\}$: The set of ratings expressed for the entity $j$.\\
	 $RS_j=\{rs_{1j},\ rs_{2j},\ \dots,\ rs_{nj}\}$: The set of reviews' score expressed for\\ the entity $j$.\\
	 $Rep$:\: reputation value toward the target entity $j$.\\

    }
    \KwIn{$V_j$, $RS_j$
    }
    \KwOut{$Rep$}
	$temp \longleftarrow 0$\\
    \For{$i\ in\ range(n)$}{
        $temp \longleftarrow temp + rs_{ij} \times v_{ij}$\\
        
    }
	$Rep \longleftarrow temp / n$
\caption{Reputation generation}
\label{ReputationGenerationAlgo}
\end{algorithm}

Assuming that an entity $E_j$ contains three reviews where $RH_j=\{100, 50, 1\}$, $RT_j=\{2020, 2010, 2000\}$, $BERT_{j}=\{0.998, 0.997, 0.996\}$ and $V_j=\{10, 10, 10\}$. By applying formula (4.1) and (4.2), we get the helpfulness and time scores: $H(r_{1j}) = 1$, $H(r_{2j}) = 0.849$, $H(r_{3j}) = 0.75$, $T(r_{1j}) = 1$, $T(r_{2j}) = 0.98$, and $T(r_{3j}) = 0.96$. After applying formula (4), we get the reviews scores: $RS_j=\{0.999, 0.942,  0.902\}$. In order to compute the reputation value toward $E_j$, we need to compute the product of $rs_{ij}$ and $v_{ij}$. We get $rs_{1j}.v_{1j} = 9.99$, $rs_{2j}.v_{2j} = 9.42$, and $rs_{3j}.v_{3j} = 9.02$. Since $Rep(E_{j}) = \frac{\sum_{i=1}^{O_{j}} rs_{ij} . v_{ij}}{O_{j}}$, we can conclude that the first review $r_{1j}$ has the highest impact (the highest product 9.99) on the reputation value of the entity $E_j$ since it is very helpful and recent. In the contrary, the third review $r_{3j}$ has the lowest impact (the lowest product 9.02). In fact, while it has the same attached rating as the first review, but, being both unhelpful and old made it by far less influential.\par
To conclude, recent and helpful reviews have more impact on the reputation value than old and unhelpful ones.

\subsection{Reputation visualization}

It is important to provide a potential customer or user with sufficient information for the purpose of assisting his decision. Thus, we propose a new way to visualize reputation by depicting the produced numerical reputation value toward the target entity, opinion categories, positive review with the highest review score (formula 4.4), and negative review with the highest review score (Figure 4.4).

\section{Experiment results}

\subsection{Data collection and preprocessing}

Five miscellaneous domains were addressed in our experiments, movie, TV show, product, hotel, and restaurant. We collected user reviews from IMDb\footnote{\url{https://www.imdb.com/}}, TripAdvisor\footnote{\url{https://www.tripadvisor.com/}}, and Amazon\footnote{\url{https://www.amazon.com}} using a web scraping tool called ScrapeStorm\footnote{\url{https://www.scrapestorm.com/}}. We extracted 400 reviews for 4 movies, 400 reviews for 4 TV shows, 200 reviews for 2 products, 100 reviews for 1 hotel, and 100 reviews for 1 restaurant. Each extracted review contains: raw text, review time, review helpfulness votes, and review attached rating (Figure 4.2). The statistical information of the dataset is shown in Table 4.3.

\begin{table}[htb]
\caption{Statistical information of dataset}
\centerline{
    \scalebox{0.8}{
        \begin{tabular}{| l | c | c | c |} 
    	\hline
    	Domain & Number of entities & Number of reviews & Number of reviews per entity
    	\\
    	\hline
    	Movie & 4 & 400 & 100\\
    	\hline
    	TV show & 4 & 400 & 100\\
    	\hline
    	Product & 2 & 200 & 100\\
    	\hline
    	Hotel & 1 & 100 & 100\\
    	\hline
    	Restaurant & 1 & 100 & 100\\
    	\hline
        \end{tabular}
            }
            }
\end{table}

After collecting the reviews, we:
\begin{enumerate}
    \item lowercase our text since we are using a  BERT lowercase model
    \item tokenize it: break words into WordPieces
    \item add special "CLS" and "SEP" tokens
    \item map our words to indexes using a vocab file that BERT provides
    \item append "index" and "segment" tokens to each input
\end{enumerate}

\subsection{Sentiment analysis}

We fine-tune $BERT_{Base}$ model to predict the sentiment orientation of the collected reviews. We build the model by creating a single new layer that will be trained with Large Movie Review Dataset v1.0 \cite{maas-EtAl:2011:ACL-HLT2011}\footnote{\url{https://ai.stanford.edu/~amaas/data/sentiment/}} which contains 25,000 positive and 25,000 negative processed movie reviews. We set the sequence length to 128, the batch size to 32, the learning rate to 0.00002, and the number of epochs to 3. Table 4.4 depicts the performance of fine-tuned BERT-base model on Large Movie Review Dataset v1.0. \\

\begin{table}[htb]
\caption{BERT-Base model classification result on Large Movie Review Dataset v1.0}
\centering
\begin{tabular}{l|c|c|c|c|}
\cline{2-5}
                                & Precision & Recall  & F1 score   & Accuracy \\ \hline
\multicolumn{1}{|l|}{BERT-Base} & 0.88048   & 0.89816 & 0.88923204 & 0.88812  \\ \hline
\end{tabular}
\end{table}

We compared BERT-Base model to Convolutional Neural Network (CNN), Long Short-Term Memory (LSTM), and Bidirectional Long Short-Term Memory (BiLSTM). GloVe embeddings were used to train LSTM, BiLSTM, and CNN. Table 4.5 depicts the performance of fine-tuned BERT-base model, Vanilla CNN, Vanilla LSTM, and Vanilla BiLSTM on Large Movie Review Dataset v1.0. \\ 

\begin{table}[htb]
\caption{Comparison results on Large Movie Review Dataset v1.0}
\centering
\begin{tabular}{|l||c|}
\hline
Approach & Accuracy       \\ \hline
Vanilla CNN              & 80.35          \\ \hline
Vanilla LSTM                   & 80.72          \\ \hline
Vanilla BiLSTM              & 81.73          \\ \hline
BERT base                      & \textbf{88.81} \\ \hline
\end{tabular}
\end{table}

We can see from Table 4.5 that BERT-Base model achieves the highest sentiment analysis accuracy compared to Vanilla CNN, Vanilla LSTM, and Vanilla BiLSTM. \par
We have mentioned in Chapter 1 some successful pre-trained models that achieve state-of-the-art results on Large Movie Review Dataset v1.0 such as XLNet-Large that achieves an accuracy of 96.21 and BERT-Large that achieves an accuracy of 95.79. However, these models require the combination of GPUs with plenty of computing power and a massive amount of memory. \par
We test BERT-Base model on our collected dataset. Figure 4.3 represents the accuracy of the model in predicting the sentiment orientation of the collected reviews.

\begin{figure}[!h]
\centering
  \includegraphics[scale=0.6]{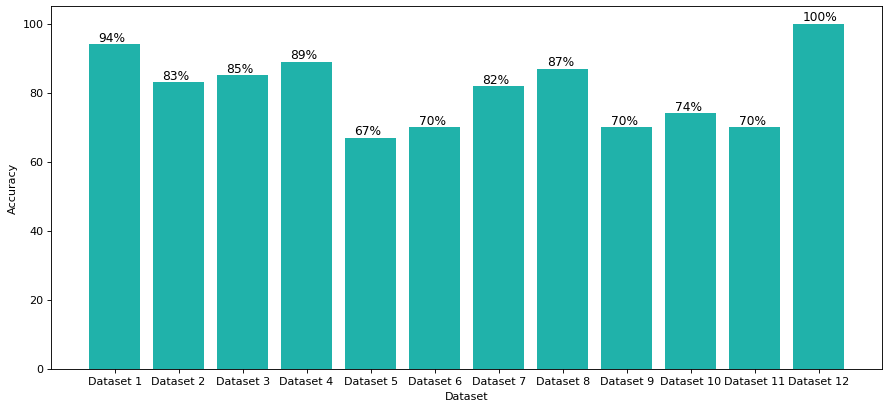}
  \caption{Accuracy of BERT-base model in predicting the sentiment orientation of the collected reviews.}
\end{figure}

We observe from Figure 4.3 that the model achieves good results in predicting the sentiment polarity of the extracted reviews. Even more impressively, the model performs well on datasets 9, 10, 11, and 12 that contain product, hotel, and restaurant reviews, even though it's trained with movie reviews.

\subsection{Reputation visualization}

We propose a new way to visualize reputation by depicting the produced numerical reputation value toward the target entity, opinion categories, top positive review, and top negative review. 

\begin{figure}[!h]
\centering
  \includegraphics[scale=0.75]{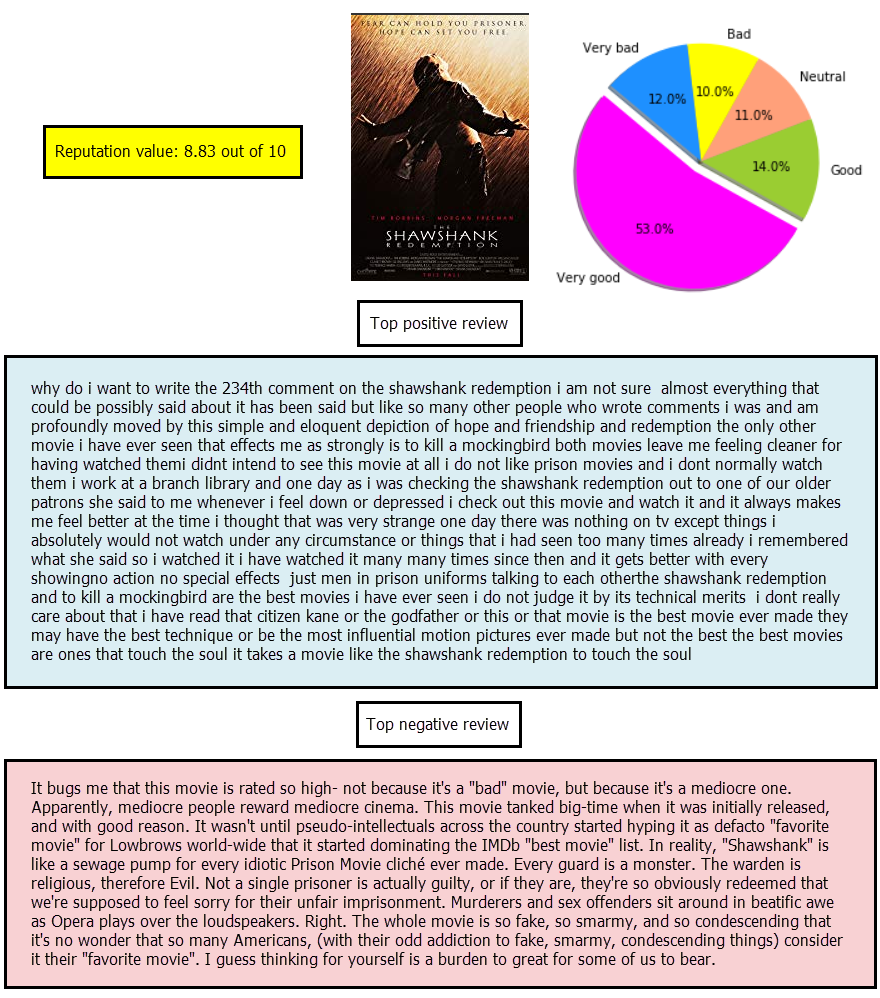}
  \caption{Reputation visualization.}
\end{figure}

As illustrated in Figure 4.4, our system provides users and potential customers with a comprehensive reputation visualization form that shows the numerical reputation value toward the target entity, opinion categories (very good, good, neutral, very bad, and bad) in a pie chart, top positive review (the positive review that holds the highest review score) and top negative review (the negative review that holds the highest review score). \par
Compared to previous studies on reputation generation \cite{yan2017fusing, benlahbib2019unsupervised, benlahbib2019hybrid}, our proposed system is the only one that presents all of this helpful information in order to support users and customers during the decision making process in e-commerce websites. Table 4.6 shows comparison results between our system and previous reputation systems in terms of reputation visualization. 

\begin{table}[htb]
\caption{Comparison results: reputation visualization}
\centerline{
    \scalebox{1}{
    \begin{tabular}{l c c c} 
	\hline

	Work & Opinion categories & Top positive review & Top negative review \\

	\hline

	Yan et al. (2017) \cite{yan2017fusing} & \cmark & \xmark & \xmark\\

	Benlahbib \& Nfaoui (2019) \cite{benlahbib2019unsupervised} & \xmark & \xmark & \xmark\\

	Benlahbib \& Nfaoui (2020) \cite{benlahbib2019hybrid} & \cmark & \xmark & \xmark \\

	This study & \cmark & \cmark & \cmark \\ \hline 
    \end{tabular}
            }
            }
\end{table}	

\subsection{System evaluation}

Previous studies on reputation generation based on mining user and customer reviews expressed in natural language have mainly focused on exploiting semantic and sentiment relations between reviews to generate reputation values toward various entities. However, customer and user reviews contain a lot of other useful information that could be exploited during the reputation generation phase like review posting time and review helpfulness votes. Unfortunately, up-to-date, no work has incorporated review time, review helpfulness votes, and review sentiment polarity to generate reputation. Therefore, we propose a new reputation system that combines review posting time, review helpfulness votes, and review sentiment orientation in order to generate accurate and reliable reputation values toward different entities. Table 4.7 depicts the difference between previous reputation systems \cite{benlahbib2019hybrid, benlahbib2019unsupervised, yan2017fusing} and our proposed reputation system.

\begin{table}[!htb]
\caption{Comparison results: review attributes exploited by recent reputation systems}
\centerline{
    \scalebox{1}{
    \begin{tabular}{l c c c c} 
	\hline

	Work & Semantic & Sentiment & Review helpfulness & Review time  \\

	\hline

	Yan et al. (2017) \cite{yan2017fusing} & \cmark & \xmark & \xmark & \xmark \\

	Benlahbib \& Nfaoui (2019) \cite{benlahbib2019unsupervised} & \cmark & \xmark & \xmark & \xmark \\

	Benlahbib \& Nfaoui (2020) \cite{benlahbib2019hybrid} & \cmark & \cmark & \xmark & \xmark \\

	This study & \cmark & \cmark & \cmark & \cmark  \\ \hline 
    \end{tabular}
            }
            }
\end{table}	

Since there are no standard evaluation metrics to assess the effectiveness and robustness of reputation systems, we conduct a user and expert survey as adopted in many research papers \cite{10.1016/j.eswa.2012.05.070}. We have invited 32 users and 3 experts to rate four reputation generation systems: System 1 (our reputation system), system 2 \cite{yan2017fusing}, system 3 \cite{benlahbib2019unsupervised} and system 4 \cite{benlahbib2019hybrid}. Each user and expert assigns a satisfaction score to each reputation system. The score ranges between 1 and 10.\par
The 32 users are from different backgrounds: 6 computer science PhD students, 2 math PhD students, an electrical engineer, an undergraduate student in mathematics, 2 computer science engineers, a physics teacher, 4 mathematics teachers, a research engineer in computer science, an electronic engineering student, an information systems engineer, a third-year student at the National School of Commerce and Management, a quality control technician, a sixth-year medical student, a housewife, 7 second-year medical students, and a software engineer.\par
Table 4.8 presents the average satisfaction score for each reputation system given by thirty-two users.

\begin{table}[!htbp]
\caption{User satisfaction comparison}
\centerline{
    \scalebox{0.85}{
    \begin{tabular}{l c c c c} 
	\hline
	Systems & Ours & Benlahbib \& Nfaoui \cite{benlahbib2019hybrid} &  Benlahbib \& Nfaoui \cite{benlahbib2019unsupervised} & Yan et al. \cite{yan2017fusing} \\ \hline
	User 1 & \textbf{9} & 8.5 & 6 & 8 \\
	User 2 & \textbf{9.5} & 8 & 7.5 & 6.9 \\
	User 3 & \textbf{9} & 8.5 & 8 & 8 \\
	User 4 & \textbf{9.5} & 8 & 4.5 & 8 \\
	User 5 & \textbf{8} & 6.5 & 6 & 6.5 \\
	User 6 & \textbf{9.5} & 8 & 5.5 & 7 \\
	User 7 & \textbf{9} & 7 & 4 & 7 \\
	User 8 & \textbf{9} & 8 & 7 & 8 \\
	User 9 & \textbf{9} & 7 & 5 & 7 \\
	User 10 & \textbf{9} & 6 & 4 & 6 \\
	User 11 & \textbf{9} & 8 & 5.5 & 8 \\
	User 12 & \textbf{9} & 8 & 6 & 7 \\
	User 13 & \textbf{8} & 6.5 & 5 & 6.5 \\
	User 14 & \textbf{8} & 5 & 3 & 5 \\
	User 15 & \textbf{9.5} & 8.5 & 5.5 & 7 \\
	User 16 & \textbf{8} & 7 & 4 & 7 \\
	User 17 & \textbf{9.5} & 8 & 7 & 7.5 \\
	User 18 & \textbf{9} & 8.5 & 6 & 7.5 \\
	User 19 & 8 & \textbf{9} & 2 & 5 \\
	User 20 & \textbf{8} & 7 & 6.5 & 7 \\
	User 21 & \textbf{9} & \textbf{9} & 6 & 6 \\
	User 22 & \textbf{9} & 8 & 6 & 7.5 \\
	User 23 &  \textbf{10} & 8 & 5 & 6 \\
	User 24 & \textbf{9} & 7 & 5 & 6 \\
	User 25 & \textbf{10} & 8.5 & 6 & 7 \\
	User 26 & \textbf{10} & 9 & 8 & 8.5 \\
	User 27 & \textbf{10} & 8 & 5 & 7 \\
	User 28 & \textbf{9} & \textbf{9} & 7 & 8 \\
	User 29 & \textbf{9} & 8 & 6 & 7 \\
	User 30 & \textbf{9.75} & 8 & 5 & 6.5 \\
	User 31 & \textbf{9} & 8 & 6 & 7 \\ 
	User 32 & \textbf{10} & 9 & 7 & 8 \\ \hline 
	\textbf{Average} & \textbf{9.07} & 7.83 & 5.625 & 7.01 \\  \textbf{Standard Deviation} & \textbf{0.63} & 0.93 & 1.32 & 0.84 \\ \hline 
    \end{tabular}
    }
            }
\end{table}	

The formula of the average satisfaction score is: $\mu = \frac{1}{N} \sum^{N}_{i=1} x_i$ where $\{x_1, x_2, \dots, x_N\}$ are the observed values of the sample items and $N$ is the number of observations in the sample. The standard deviation is a measure of the amount of variation or dispersion of a set of values \cite{Bland1654}. The formula for the standard deviation is: $\sigma = \sqrt{\frac{1}{N} \sum^{N}_{i=1}(x_i - \mu)^2}$ where $\{x_1, x_2, \dots, x_N\}$ are the observed values of the sample items, $\mu$ is the mean value of these observations, and $N$ is the number of observations in the sample. \par
We can see from Table 4.8 that 31 users favor our reputation system over the three other systems in terms of helpfulness and effectiveness in generating reputation and visualization since it achieves the highest average satisfaction score and the lowest standard deviation of satisfaction scores. Moreover,  only one user (user 19) favors system 2 \cite{benlahbib2019unsupervised}. System 2 takes second place by achieving an average satisfaction score of 7.83. System 4 \cite{yan2017fusing} comes next with a 7.01 average satisfaction score, which sounds very reasonable since the main goal of system 2 was to improve system 4 by exploiting both sentiment and semantic analysis techniques. System 3 \cite{benlahbib2019unsupervised} takes the last place by achieving an average satisfaction score of 5.625. System 3 doesn't provide users and customers with sufficient information to support their decision, since only providing them with reputation value alone isn't enough to help them make a judgment about a target item, indeed, the customers need more helpful information that could support them during the decision-making process such as opinion categories, top positive review and top negative review. \par
We enrich our experiment results by inviting 3 experts to rate each reputation system with a satisfaction score. Expert 1 is a former owner of an e-commerce website whose main field of interest is natural language processing and machine learning, while expert 2 is an active e-commerce buyer and seller with more than 9 years of experience. As for expert 3, he is a third-year PhD student in economics sciences. Table 4.9 presents the average satisfaction score for each reputation system given by the three experts.

\begin{table}[!htbp]
\caption{Expert satisfaction comparison}
\centerline{
    \scalebox{1}{
    \begin{tabular}{l c c c c} 
	\hline
	Systems & Ours & Benlahbib \& Nfaoui \cite{benlahbib2019hybrid} & Benlahbib \& Nfaoui \cite{benlahbib2019unsupervised} & Yan et al. \cite{yan2017fusing} \\ \hline
	Expert 1 & \textbf{10} & 6 & 5 & 6 \\
	Expert 2 & \textbf{8.5} & 7.5 & 5 & 5.5 \\
	Expert 3 & \textbf{9} & 8 & 7 & 7 \\ \hline 
	\textbf{Average} & \textbf{9.17} & 7.17 & 5.67 & 6.17 \\ \hline 
    \end{tabular}
    }
            }
\end{table}	

Based on the average satisfaction score given by the three experts (Table 4.9), reputation system 1 takes first place with an average satisfaction score of 9.17, preceded by system 2, system 4, and system 3 comes in last place with 5.67 as average satisfaction score. Figure 4.5 combines the results of Table 4.8 and Table 4.9.

\begin{figure}[!h]
\centering
  \includegraphics[scale=1]{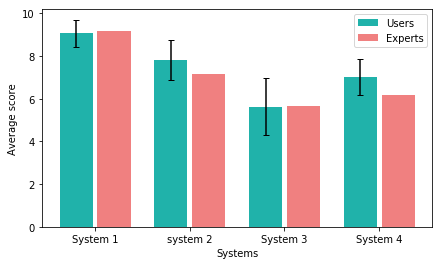}
  \caption{Users and experts average satisfaction score for each reputation system.}
\end{figure}

Figure 4.5 shows that both users and experts choose system 1 as the best in terms of reputation generation and visualization. System 2 holds second place, preceded by system 4. system 3 comes in fourth place. \par
We asked the three experts to share their opinions about system 1 strengths and weaknesses. Table 4.10 contains expert reviews toward system 1.

\begin{table}[!htbp]
\caption{Experts reviews}
\centerline{
    \scalebox{1}{
    \begin{tabular}{|p{17mm} | p{150mm}|}
	\hline
	Expert & Review \\ \hline
	Expert 1 & The present work proposes a new method for reputation generation and visualization by incorporating helpfulness votes and time review features to sentiment and semantic features. In addition, the system outputs the item reputation value, top positive review, top negative review, and a pie chart that shows the distribution of sentiment over the reviews. Based on all these properties, i give a 10 to system 1. I think that this work can be enhanced by mining the top reviews according to the system's user. For instance, if two reviews have been selected as top positive reviews (the same reviews score), which review should the system display? By analyzing the users' data and behavior, the system will output the most accurate review with regard to user preferences. \\ \hline
	Expert 2 & I choose system 1 as the best because it covers many important criteria neglected by other systems. On the one hand, the numerical value and opinion categories alone only reveal the general impact of the film (product) on users (good film, nice film, bad film) and ignore the in-depth details that are essential for any kind of reputation. On the other hand, System 1 takes into account all four attributes of the evaluation, which will have a positive effect on the accuracy and credibility of the reputation value. 
	I give it 8.5/10, because, in my opinion, the method still has some shortcomings since the best positive/negative reviews can contain spoilers for movies or bad personal experiences for a service, and this would create some confusion for the user. \\ \hline 
	Expert 3 & I underline the importance of this study in terms of its usefulness and interest. It is an issue that seeks to put in place a system that allows subjective analysis, and thus scrutinizes and makes the opinions of viewers more concrete. And so, after having an idea of the different reputation systems, I consider the first system to be the most efficient in terms of reputation generation since it takes into consideration more variables/determinants such as review helpfulness votes and review time.
  \\ \hline 
    \end{tabular}
    }
            }
\end{table}	

\subsection{Further discussion}

In summary, our reputation system exhibits the following advantages:
\begin{itemize}
    \item \textbf{Accuracy}: The system incorporates review helpfulness, review time, review sentiment orientation probability, and review attached rating in order to generate an accurate reputation value.
	\item \textbf{Holistic}: The system proposes a new form of reputation visualization that depicts numerical reputation value, opinion categories, top positive review, and top negative review. The system also can output the top-k positive reviews and the top-k negative reviews. This new form of reputation visualization provides customers with sufficient information toward the target item in order to make an informed decision (buying, renting, booking) toward it.
	\item \textbf{Generality}: The system can be applied in any website that allows web users to (1) post their reviews expressed in natural languages, (2) share their numerical or star ratings, and (3) vote for helpful reviews. Furthermore, the system can be applied to various domains (products, movies, services, hotels).
	\item \textbf{Usefulness}: The system is very useful in terms of supporting web customers during the decision-making process in E-commerce by instantly providing them with sufficient information toward the target item, saving them from spending both their time and effort on reading thousands of online reviews.
\end{itemize}

However, our reputation system suffers from:
\begin{itemize}
    \item \textbf{Safety}: Due to the openness of the Internet, many malicious users post fake reviews (false positive/false negative) aiming to impact the popularity and credibility of online products. Therefore, our system should incorporate a filtering phase in order to detect and remove fake and irrelevant reviews.
\end{itemize}

\section{Conclusion}

In this chapter, we have proposed a reputation system that generates reputation toward various items (products, movies, TV shows, hotels, restaurants, services) by mining customer and user reviews expressed in natural language. The system incorporates four review attributes: review helpfulness, review time, review sentiment polarity, and review rating. The system also provides a comprehensive reputation visualization form by depicting the numerical reputation value, opinion group categories, top-k positive reviews, and top-k negative reviews. To better evaluate the effectiveness of our reputation system, 32 users and 3 experts were invited to assign a score of one (least satisfaction) to ten (highest satisfaction) to four reputation generation systems. Our reputation system achieved the highest average satisfaction score given by both users and experts. The three experts were also invited to share their point of view toward the proposed system in terms of reputation generation and visualization. \par
We believe that the proposed system represents an interesting online reputation system, full of fascinating insights into customer's decision-making process in e-commerce websites.\\
Future studies will focus on:
\begin{itemize}
    \item exploiting further features including user credibility (prolific reviewers) and user's online behavior as suggested by expert 1 (Table 4.10).  
    \item detecting and removing fake and irrelevant reviews by applying a filtering phase, and therefore reducing the processing time, and increasing the efficiency of the system at once, since only relevant and useful reviews will be taken into account.
    \item incorporating aspect-based opinion mining during the phase of reputation generation and visualization. As a result, the reputation visualization will be enhanced. Indeed, the system will depict more useful information toward the target entity $E$ such as its features ($E_{featureX}$, $E_{featureY}$, $E_{featureZ}$ \dots), the number of positive reviews toward feature $E_{featureX}$, and the number of negative reviews toward feature $E_{featureY}$ \dots
\end{itemize}
The next chapter will describe AmazonRep, a reputation system that extends the system proposed in this chapter by exploiting review rating, review helpfulness votes, review time, review sentiment orientation, and user credibility to support Amazon's customer decision-making process. \par

\chapter{Reputation Generation and Visualization for Amazon's Products}
\section{Introduction}
Over the last few years, e-commerce has been significantly growing and expanding \cite{10.5555/1597148.1597269, 10.1016/j.eswa.2012.05.070, 10.5555/1863190.1863201, hou2019mining, 10.1016/j.ipm.2016.12.002}. According to \cite{10.1007/s00500-017-2882-2, doi:10.1080/00207721.2015.1116640, Amarouche_Benbrahim_Kassou_2018, 6509374, pecar-2018-towards}, the number of reviews attached to an online entity (product, movie, hotel, service) may exceed thousands, which makes it impossible for a potential customer to read them all in order to make a decision (buying, booking, renting) toward the target item \cite{coavoux-etal-2019-unsupervised}. Thus, the need for the right tools and technologies to help in such a task becomes a necessity for customers.\par
During the last five years, few systems have been proposed to generate and visualize reputation by mining user and customer reviews expressed in natural language \cite{yan2017fusing, benlahbib2019hybrid, benlahbib2019unsupervised, 9068916, 9098950, Benl2112:MTVRep}. However, none of them has combined review time, review polarity, review helpfulness votes, and user credibility for the purpose of generating and visualizing reputation.\par
In this chapter, we propose AmazonRep, a reputation system to support Amazon's customer decision making process by incorporating review rating, review helpfulness votes, review time, review sentiment orientation, and user credibility. The system also provides a comprehensive reputation visualization form by depicting the numerical reputation value, opinion group categories, top-k positive reviews, and top-k negative reviews in order to help potential customers make an informed decision on whether to purchase the product.\par

\section{Problem definition}

This section covers the necessary background for understanding the remainder of this chapter, including the problem definition. \par
In this chapter, we face the problem of generating reputation for Amazon's products by aggregating review time, review helpfulness votes, user credibility, review sentiment orientation, and review attached rating. Given a set of reviews $R_j=\{r_{1j},\ r_{2j},\ \dots,\ r_{nj}\}$ expressed for a product $E_j$, written by users $U=\{u_{1},\ u_{2},\ \dots,\ u_{n}\}$, the set of their numerical ratings $V_j=\{v_{1j},\ v_{2j},\ \dots,\ v_{nj}\}$ where $v_{ij} \in [1,5]$, the set of their attached helpfulness votes $RH_j=\{rh_{1j},\ rh_{2j},\ \dots,\ rh_{nj}\}$ where $rh_{ij} \in \mathbb{N^*}$, the set of their posting time $RT_j=\{rt_{1j},\ rt_{2j},\ \dots,\ rt_{nj}\}$, the set of the total number of helpful votes that users $U$ have received $UH=\{h_{1},\ h_{2},\ \dots,\ h_{n}\}$, and the set of the total number of reviews written by users $U$ in Amazon $N=\{n_{1},\ n_{2},\ \dots,\ n_{n}\}$. The goal is to compute a review score for each review $RS_j=\{rs_{1j},\ rs_{2j},\ \dots,\ rs_{nj}\}$ based on its helpfulness votes, its posting time, and its user credibility, and finally, to compute a reputation value $Rep$ for a product $j$ by averaging the product of reviews score and reviews attached rating. Table 5.1 presents the descriptions of notations used in the rest of this chapter. \par

\begin{table}[htb]
\caption{Symbol denotation}
\centerline{
\begin{tabular}{| l | p{12cm} |}
\hline
Symbol & Description                                                              \\ \hline
$R_j$      & The set of reviews expressed for the product $j$                                                                  \\
$V_j$      & The set of ratings expressed for the product $j$                                                                  \\

$U$      & The set of users that have expressed their opinions toward product $j$                                                                  \\

$RH_j$     & The set of reviews' helpfulness votes expressed for the product $j$                                               \\
$RT_j$     & The set of reviews' posting time expressed for the product $j$                                                     \\
$RS_j$     & The set of reviews score expressed for the product $j$                                                            \\

$UH$      & The set of the total number of helpful votes that users $U$ have received                                                                  \\

$N$      & the set of the total number of reviews written by users $U$                                                                  \\

$E_j$      & The target product $j$                                                                 \\
$O_j$      & The total number of reviews expressed for the target product $j$                       \\
$Rep$    & The reputation value  \\ \hline                                                             
\end{tabular}
}
\end{table}

\section{Proposed approach}

\subsection{System overview}

Our approach consists mainly on four steps:

\begin{itemize}
    \item Firstly, we collect 1300 product reviews from Amazon\footnote{\url{https://www.amazon.com}}, and, we preprocess them.
    \item  Secondly, we compute review helpfulness score, review time score, and user credibility score.
    \item Thirdly, we compute a review score based on the pre-computed scores (helpfulness score, time score, and user credibility score).
    \item Finally, we generate a numerical reputation value toward the target product. Then, we visualize reputation by depicting numerical reputation value, opinion categories, the top-k positive reviews with the highest scores, and the top-k negative reviews with the highest scores.
\end{itemize}

Figure 5.1 describes the pipeline of our reputation system.

\begin{figure}[!h]
\centering
  \includegraphics[scale=0.75]{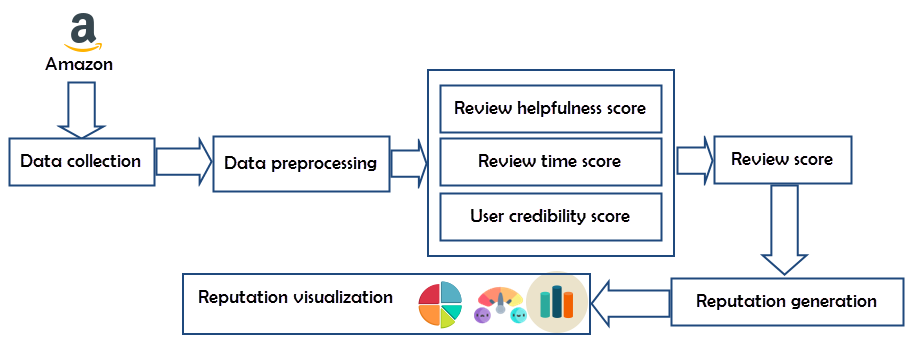}
  \caption{AmazonRep pipeline.}
\end{figure}

\subsection{Review structure}

We gather online reviews from Amazon\footnote{\url{https://www.amazon.com}} website with respect to the following structure: \textbf{textual review, review helpfulness votes, review posting time, number of user's reviews, and number of user's helpful votes}. Figure 5.2 describes the structure of Amazon's reviews.

\begin{figure}[!h]
\centering
  \includegraphics[scale=0.8]{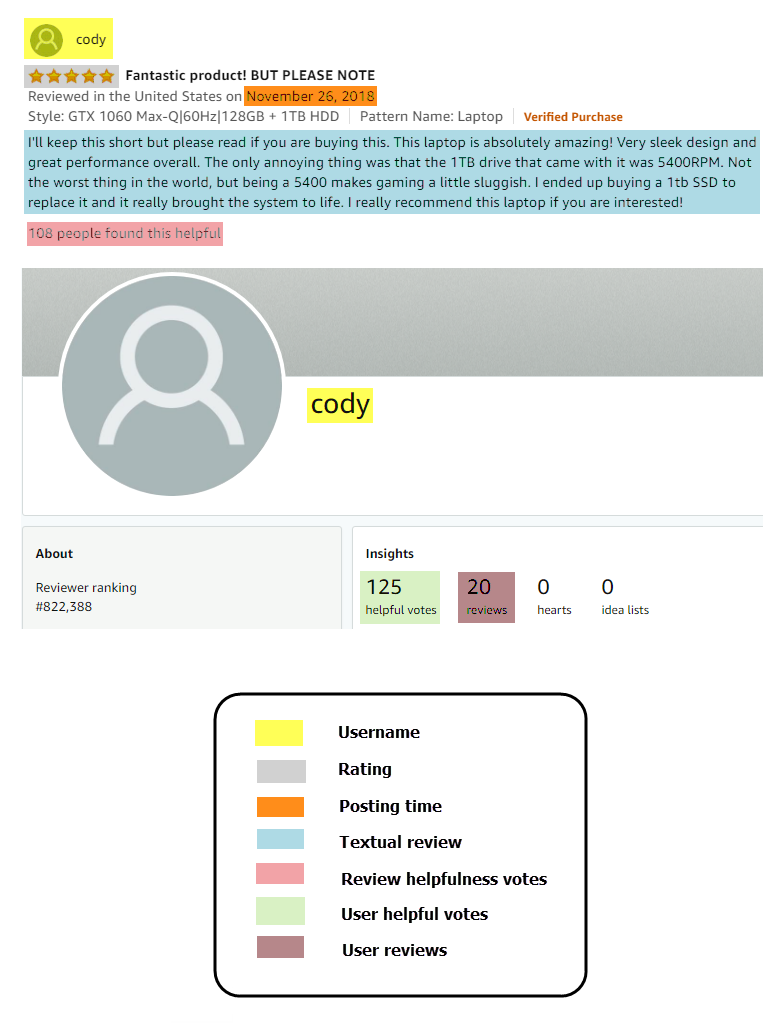}
  \caption{Amazon's review structure.}
\end{figure}

\subsection{Review helpfulness}

According to \cite{9098950}: The number of helpfulness votes attached to a review indicates how informative it is, which implies that reviews that receive higher votes from other users typically provide more valuable information. Thus, we design equation (5.1) to assign a helpfulness score to each review.
\begin{equation}
    H(r_{ij}) = 
        \begin{cases}
           \text{0.8} &\quad\text{if  } rh_{ij}=0 \text{  or  } \frac{\log_{10}(rh_{ij})}{\log_{10}(N_{j})} \leqslant 0.8\\
           \log_{N_j}(rh_{ij})\text{} &\quad\text{Otherwise}   \\
         \end{cases}
\end{equation}
\bigskip
We denote: \\ \\
$r_{ij}: $ Review number $i$ expressed for the entity $j$.\\
$H(r_{ij}): $ Helpfulness score of review $r_{ij}$.\\
$rh_{ij}: $ The number of helpfulness votes attached to review $r_{ij}$.\\
$N_{j}: $ The number of helpfulness votes attached to the most voted review toward the entity $j$.\\

The review helpfulness score is comprised between 0.8 and 1 because we don't want to assign a low score to reviews with a small number of helpfulness votes.\par
We mention that $\frac{\log_{10}(rh_{ij})}{\log_{10}(N_{j})} \leqslant 0.8$ means that $\log_{N_j}(rh_{ij}) \leqslant 0.8$ due to the fact that:\\
$
    \log_N(n) = \frac{\log_{base}(n)}{\log_{base}(N)} = \frac{\log_{10}(n)}{\log_{10}(N)} 
$

\bigskip

By applying equation (5.1), the most voted review will receive a review helpfulness score of 1 since $\log_{N_j}(N_j) = 1$. Also, reviews with high helpfulness votes will receive a high review helpfulness score and reviews with low helpfulness votes will receive a low review helpfulness score since for $x \in [1, N]$ and $y \in [1, N]$: $x \leq y$ implies that $log_N(x) \leq log_N(y)$.\\

\subsection{Review Time}

More recent reviews generally provide users with more up-to-date information. Therefore, we design formula (5.2) to assign a time score to each review. 
\begin{equation}
    T(r_{ij}) = 
        \begin{cases}
           \text{0.8} &\quad\text{if  } y - rt_{ij} \geqslant 100\\
           1-(y-rt_{ij}) \times 0.002 &\quad\text{ Otherwise} \\ 
         \end{cases}
\end{equation}
\bigskip
We denote: \\ \\
$T(r_{ij}): $ Time score of review $r_{ij}$.\\
$rt_{ij}: $ Publication year of review $r_{ij}$.\\
$y: $ Current year.\\

The time score for a review ranges between 0.8 and 1, which implies that a higher time score is assigned to the most recent reviews.\\

\subsection{User credibility}

We propose equation (5.3) to compute a credibility score for users (customers) based on the number of their helpfulness votes and the number of their reviews.

\begin{equation}
    C(u_{i}) = \frac{1} {{1 + e^{- \frac{h_{i}}{n_{i}}}}}
\end{equation}
\bigskip
We denote: \\ \\
$C(u_{i}): $ Credibility score of user $u_{i}$.\\
$u_{i}: $ The author of review $r_{ij}$.\\
$h_{i}: $ The number of helpful votes of user $u_{i}$.\\
$n_{i}: $ The number of reviews written by user $u_{i}$.\\

The user credibility score is computed by applying the sigmoid function to the ratio of $h_{i}$ to $n_{i}$ where $sigmoid(x) = \sigma(x) = \frac{1} {{1 + e^{- x}}}$. Figure 5.3 shows the plot of the sigmoid function.

\begin{figure}[!h]
\centering
  \includegraphics[scale=0.5]{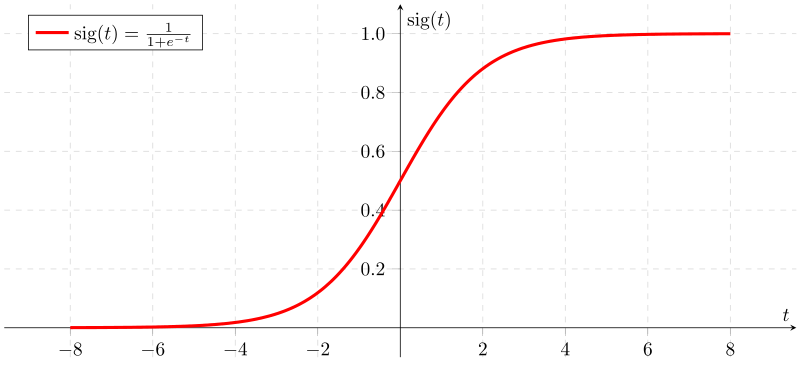}
  \caption{Plot of the sigmoid function.}
\end{figure}

From Figure 5.3, we can see that $0 \leq \sigma(x) \leq 1$ and $\sigma(0) = 0.5$.\\
Since $\frac{h_{i}}{n_{i}} \geq 0$, the user credibility score is ranging between 0.5 and 1. 

\subsection{Review score}

Based on the above scores, we propose formula (5.4) to assign a numerical score to each review:
\begin{equation}
    RS(r_{ij}) = \frac{\alpha_1 \times H(r_{ij}) + \alpha_2 \times T(r_{ij}) + \alpha_3 \times C(u_{i})}{\alpha_1 + \alpha_2 + \alpha_3}
\end{equation}
\bigskip
We denote: \\ \\
$RS(r_{ij}): $ Review score for review $r_{ij}$.\\
$H(r_{ij}): $ Helpfulness score of review $r_{ij}$.\\
$T(r_{ij}): $ Time score of review $r_{ij}$.\\
$C(u_{i}): $ Credibility score of user $u_{i}$.\\

$\alpha_1$, $\alpha_2$ and $\alpha_3$ are constants that represent the impact of each score (review helpfulness score, review time score, and user credibility score) on the reputation value.\par
After conducting several experiments, we have found that AmazonRep generates accurate reputation values toward products when $\alpha_1 = 0.4$, $\alpha_2 = 0.35$, and $\alpha_3 = 0.25$. Consequently, we set $\alpha_1 = 0.4$, $\alpha_2 = 0.35$, and $\alpha_3 = 0.25$. Table 5.2 represents an example results of review score.

\begin{table}[!htb]
\caption{Example results of review score}
\centerline{
    \scalebox{1}{
\begin{tabular}{|l|c|c|c|c|c|}
\hline
Review   & \begin{tabular}[c]{@{}c@{}}Review \\ helpfulness score\end{tabular} & \begin{tabular}[c]{@{}c@{}}Review\\  time score\end{tabular} & \begin{tabular}[c]{@{}c@{}}User \\ credibility score\end{tabular} & \begin{tabular}[c]{@{}c@{}}Review\\  sentiment orientation\end{tabular} & \begin{tabular}[c]{@{}c@{}}Review \\ score\end{tabular} \\ \hline
Review 1 & 1                                                                   & 0.994                                                        & 1                                                                & Negative                                                        &    0.9979                                           \\ \hline
Review 2 & 0.8                                                                & 0.994                                                        & 0.97861669                                                               & Negative                                                        & 0.91255417                                              \\ \hline
Review 3 & 0.8                                                          & 0.994                                                        & 0.99991620                                                                & Negative                                                        & 0.91787905                                              \\ \hline
Review 4 & 0.8                                                          & 0.996                                                        & 1                                                                & Positive                                                         & 0.9186                                              \\ \hline
Review 5 & 0.96668280                                                          & 1                                                         &    0.95257412                                                             & Negative                                                        & 0.97481665                                              \\ \hline
\end{tabular}
}
}
\end{table}

\subsection{Reputation generation}

We propose formula (5.5) to compute a single reputation value toward the target entity using review score $RS(r_{ij})$ and review attached rating $v_{ij}$:
\begin{equation}
    Rep(E_{j}) = \frac{\sum_{i=1}^{O_{j}} RS(r_{ij}) . v_{ij}}{O_{j}}
\end{equation}
\bigskip
We denote: \\ \\
$E_{j}: $ Target entity $j$.\\
$Rep(E_{j}): $ Reputation value toward the target entity $j$.\\
$RS(r_{ij}): $ Review score for review $r_{ij}$.\\
$v_{ij}: $ Attached numerical rating to review $r_{ij}$.\\
$O_{j}: $ Total number of reviews expressed for the target entity $j$.\\

The reputation value varies from 1 to 5 since Amazon website uses a five-star rating system.\\

\subsection{Reputation visualization}

Our system provides Amazon's customers with sufficient information toward the target product for the purpose of supporting them during the decision making process. The system depicts numerical reputation value, opinion categories, top positive review, and top negative review. Figure 5.4 shows an example of reputation visualization.

\section{Experiment results}

\subsection{Data collection}

We collected 1300 customer reviews for 13 products (1 book, 2 laptops, 3 smartphones, 2 washing machines, 2 refrigerators, 2 movies, and 1 video game) from Amazon\footnote{\url{https://www.amazon.com}}. The attributes related to the review (rating, posting time, helpfulness votes, and textual review) were collected using a web scraping tool called ScrapeStorm\footnote{\url{https://www.scrapestorm.com/}}. The attributes related to the user (number of user's reviews and number of user's helpful votes) were manually collected.\par
After collecting the reviews, we preprocess them by removing word segmentation and numbers. Also, we remove the commoner morphological and inflexional endings from words in English by applying Porter stemming algorithm \footnote{\url{https://tartarus.org/martin/PorterStemmer/}}.

\subsection{Sentiment analysis}

We train a Bidirectional Gated Recurrent Units (Bi-GRU) \cite{cho-etal-2014-learning} with Large Movie Review Dataset v1.0 \cite{maas-EtAl:2011:ACL-HLT2011}\footnote{\url{https://ai.stanford.edu/~amaas/data/sentiment/}} in order to determine the sentiment orientation (positive/negative) of the collected reviews. We split the dataset into 45000 reviews for training and 5000 reviews for validation. We set the number of epochs to 5, the batch size to 512. We apply Adam optimizer \cite{DBLP:journals/corr/KingmaB14} with a learning rate 0.001. Table 5.3 shows the Bi-GRU model classification result on the validation set.

\begin{table}[H]
\caption{Bi-GRU model classification result on Large Movie Review Dataset v1.0}
\centering
\begin{tabular}{l|c|c|c|c|}
\cline{2-5}
                                & Precision & Recall  & F1 score   & Accuracy \\ \hline
\multicolumn{1}{|l|}{Bi-GRU} & 0.89   & 0.89 & 0.89 & 0.888  \\ \hline
\end{tabular}
\end{table}

\subsection{Reputation generation and visualization}

Previous studies on reputation generation\cite{yan2017fusing, benlahbib2019hybrid, benlahbib2019unsupervised, 9068916, 9098950} have disregarded incorporating user credibility with other factors during the phase of reputation generation and visualization. Therefore, we propose AmazonRep that combines review time, review polarity, review helpfulness votes, and user credibility in order to generate and visualize reputation. Table 5.4 and Table 5.5 depict the comparison results between AmazonRep and five other reputation systems in terms of reputation generation and visualization.\\

\begin{table}[!ht]
\caption{Comparison results: reputation visualization}
\centerline{
    \scalebox{1}{
    \begin{tabular}{l c c c} 
	\hline

	Work & Opinion categories & Top positive review & Top negative review \\

	\hline

	Yan et al. (2017)  \cite{yan2017fusing} & \cmark & \xmark & \xmark\\

	Benlahbib \& Nfaoui (2019) \cite{benlahbib2019unsupervised} & \xmark & \xmark & \xmark\\

    Benlahbib et al. (2019) \cite{9068916} & \xmark & \xmark & \xmark \\

	Benlahbib \& Nfaoui (2020) \cite{benlahbib2019hybrid} & \cmark & \xmark & \xmark \\

	Benlahbib \& Nfaoui (2020) \cite{9098950} & \cmark & \cmark & \cmark \\

	This study & \cmark & \cmark & \cmark  \\ \hline 
    \end{tabular}
            }
            }
\end{table}	

\begin{table}[!htb]
\caption{Comparison results: review attributes exploited by recent reputation systems}
\centerline{
    \scalebox{0.9}{
    \begin{tabular}{l c c c c c} 
	\hline

	Work & Semantic & Sentiment & Review helpfulness & Review time & User credibility
	\\
	\hline

	Yan et al. (2017) \cite{yan2017fusing} & \cmark & \xmark & \xmark & \xmark & \xmark \\

	Benlahbib \& Nfaoui (2020) \cite{benlahbib2019unsupervised} & \cmark & \xmark & \xmark & \xmark & \xmark \\

	Benlahbib et al. (2019) \cite{9068916} & \cmark & \cmark & \xmark & \xmark & \xmark \\

	Benlahbib \& Nfaoui (2020) \cite{benlahbib2019hybrid} & \cmark & \cmark & \xmark & \xmark & \xmark \\

	Benlahbib \& Nfaoui (2020) \cite{9098950} & \cmark & \cmark & \cmark & \cmark & \xmark \\

	This study & \cmark & \cmark & \cmark & \cmark & \cmark  \\ \hline 
    \end{tabular}
            }
            }
\end{table}	

\begin{figure}[!h]
\centering
  \includegraphics[scale=0.9]{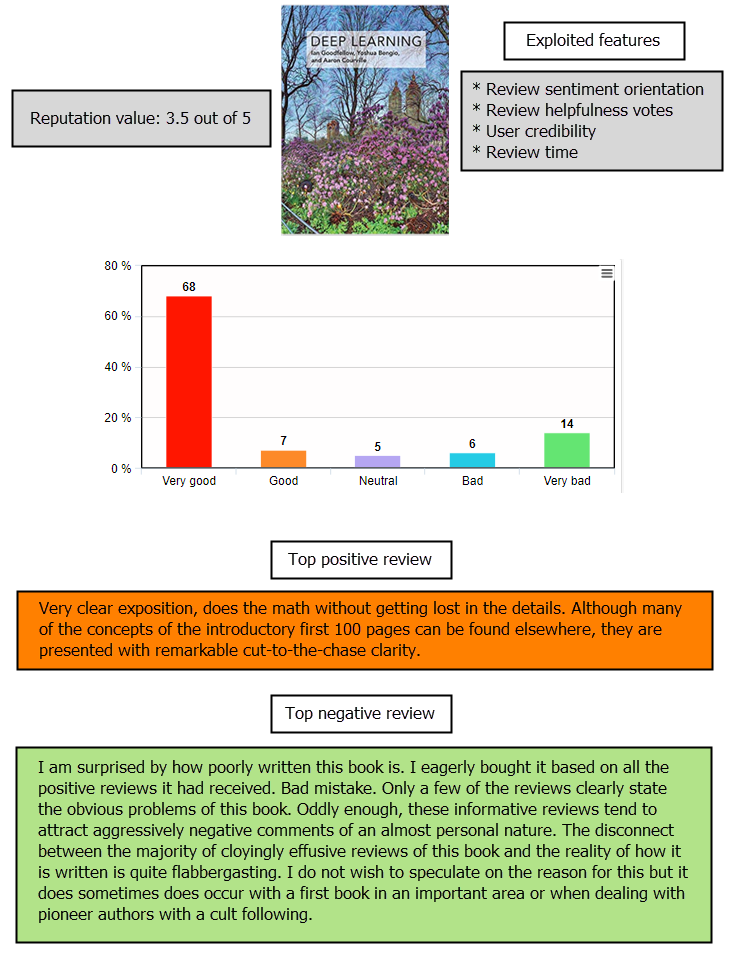}
  \caption{Reputation visualization.}
\end{figure}

\section{Conclusion}

In this chapter, we have designed and built AmazonRep, a reputation system to support Amazon's customer decision-making process based on mining product reviews and their attributes. The proposed system combines review sentiment orientation, review helpfulness votes, review time, and user credibility in order to generate accurate reputation values toward Amazon's products. The system is very useful since it instantly provides customers with sufficient information toward the target product (numerical reputation value, opinion categories, top positive review, and top negative review), saving them from spending both their time and effort on reading thousands of online reviews.\\
Future studies will focus on incorporating aspect-based opinion mining \cite{5708129, 10.1145/3177457.3177462} during the phase of reputation generation and visualization in order to improve the efficiency of the system.

\newpage
\chapter*{Conclusions}
\addcontentsline{toc}{chapter}{Conclusions }

In this dissertation, I introduced four reputation systems that can automatically provide E-commerce customers with valuable information to support them during their online decision-making process by mining online reviews expressed in natural language.\par
The first chapter describes and examines previous research work done in the area of natural language processing (NLP) techniques for decision making in E-commerce, document-level sentiment analysis, and fine-grained sentiment analysis. The chapter also covers the necessary background for understanding Bidirectional Encoder Representations from Transformers (BERT) model since we employed it to determine the sentiment orientation of customer and user reviews.\par
Chapter 2 presents a reputation system that incorporates sentiment analysis, semantic analysis, and opinion fusion to generate accurate reputation values toward online items.\par
The next chapter describes MTVRep, a movie and TV show reputation system that exploits fine-grained sentiment analysis and semantic analysis for the purpose of generating and visualizing reputation toward movies and TV shows.\par
Chapter 4 presents a reputation system that incorporates four review attributes: review helpfulness, review time, review sentiment polarity, and review rating in order to generate reputation toward various online items (products, movies, TV shows, hotels, restaurants, services). The system also provides a comprehensive reputation visualization form to help a potential customer make an informed decision by depicting the numerical reputation value, opinion group categories, top-k positive reviews, and top-k negative reviews.\par
Chapter 5 describes AmazonRep, a reputation system that extends the system proposed in chapter 4 by exploiting review rating, review helpfulness votes, review time, review sentiment orientation, and user credibility for the purpose of supporting Amazon's customer decision making process.\par
Future studies will focus on:
\begin{itemize}
    \item exploiting further features including user's online behavior.  
    \item detecting and removing fake and irrelevant reviews by applying a filtering phase, and therefore reducing the processing time, and increasing the efficiency of the system at once, since only relevant and useful reviews will be taken into account.
    \item incorporating aspect-based opinion mining during the phase of reputation generation and visualization. As a result, the reputation visualization will be enhanced. Indeed, the system will depict more useful information toward the target entity $E$ such as its features ($E_{featureX}$, $E_{featureY}$, $E_{featureZ}$ \dots), the number of positive reviews toward feature $E_{featureX}$, and the number of negative reviews toward feature $E_{featureY}$ \dots
\end{itemize}

\addcontentsline{toc}{chapter}{Bibliography}

\bibliographystyle{IEEEtran}

\bibliography{Bibliography}



\end{document}